\documentclass[a4paper,12pt]{elsarticle}

\usepackage{graphicx}

\usepackage{amsmath,amssymb}
\usepackage{amsthm}
\usepackage{mathtools}
\usepackage{enumitem}

\usepackage{booktabs}
\usepackage{adjustbox}

\usepackage{lmodern}

\usepackage{tikz}
\usetikzlibrary{shapes.geometric, positioning}

\PassOptionsToPackage{hyphens}{url}
\usepackage{xurl}
\usepackage{hyperref}

\newcommand{\prescAI}{Prescriptive AI}

\newcommand{\Mpred}{M_{\mathrm{pred}}}
\newcommand{\pih}{\pi_h}
\newcommand{\calS}{\mathcal{S}}
\newcommand{\calA}{\mathcal{A}}
\newcommand{\calN}{\mathcal{N}}
\newcommand{\calL}{\mathcal{L}}
\newcommand{\calE}{\mathcal{E}}
\newcommand{\calH}{\mathcal{H}}
\newcommand{\calU}{\mathcal{U}}
\newcommand{\calR}{\mathcal{R}}
\newcommand{\calM}{\mathcal{M}}
\newcommand{\ebias}{\varepsilon_{\mathrm{bias}}}
\DeclareMathOperator*{\argmax}{arg\,max}

\begin{document}

\begin{frontmatter}
\title{Prescriptive Artificial Intelligence:  A Formal Paradigm for Auditing Human Decisions Under Uncertainty}
\author[inst1]{Pedro Passos Farias\corref{cor1}}
\cortext[cor1]{Corresponding author.}
\ead{pedropassos@id.ufr.br}

\affiliation[inst1]{
  organization={Polytechnic School, Federal University of Rio de Janeiro(UFRJ)},
  addressline={Av. Gal. Milton Tavares de Souza},
  postcode={24210-346},
  city={Niterói},
  state={RJ},
  country={Brazil}
}

\begin{abstract}
\footnotesize
We formalize Prescriptive Artificial Intelligence as a distinct paradigm for human-AI decision collaboration in high-stakes, stochastic environments involving single-agent individual decision-making. Unlike predictive systems optimized for outcome accuracy, prescriptive systems audit human decisions under uncertainty, providing normative guidance while preserving human agency and accountability.

We introduce four domain-independent axioms characterizing prescriptive systems and prove fundamental separation results. Central is the \emph{Imitation Incompleteness} theorem: supervised learning from historical decisions cannot correct systematic biases in the absence of external normative signals. Under standard regularity conditions, the induced predictor converges almost surely to the biased action rather than the normatively optimal one. Performance in decision imitation is therefore bounded by a structural bias term \(\varepsilon_{\text{bias}}\) rather than the statistical rate \(\mathcal{O}(1/\sqrt{n})\)---a result extended to Markovian logs and finite-sample concentration bounds.

We complement this with a second impossibility result, \emph{Normative Non-Identifiability}: the utility function underlying rational choice cannot be recovered from behavioral data under systematic bias, and this failure is \emph{directional} under estimation, unlike the classical symmetric reward ambiguity of inverse reinforcement learning.

We demonstrate realizability through three independent instantiations spanning five decades: an interpretable fuzzy system for elite soccer auditing, revealing decision latency and risk states obscured by outcome and status quo biases; MYCIN, the historically validated rule-based clinical consultation system; and NEWS2, a nationally mandated clinical protocol validated on a prospective multi-center cohort. The framework establishes Prescriptive AI as a general, realizable class of decision-support systems for safety-critical domains where interpretability, contestability, and normative alignment are essential.
\end{abstract}


\end{frontmatter}
\clearpage

\section{Introduction}

Decision making under uncertainty in high-stakes environments constitutes a central challenge for human judgment. Settings such as clinical triage, operational risk management, and strategic resource allocation are characterized by irreversible actions, asymmetric consequences, and a critical dependence on timely and well-calibrated decisions. In such domains, decision errors are costly not only because of incorrect outcomes, but because they are often unjustified, delayed, or poorly aligned with underlying risk.

Despite these constraints, computational decision-support systems have been dominated by a fundamentally predictive paradigm. Models are typically optimized to replicate historical decision patterns, achieving statistical accuracy by mimicking prior human behavior. However, reliance on behavioral imitation is problematic: historical decisions often encode cognitive biases and institutional inertia. As a result, predictive systems create a structural mismatch between accuracy and decision quality, reaching a documented ``predictive ceiling''—for instance, state-of-the-art substitution models plateau at approximately $70\%$ accuracy by merely cloning human choices \citep{mohandas}. Moreover, standard evaluation practices in stochastic environments suffer from \emph{outcome bias}, judging decision quality by realized results rather than by the epistemic justification available at the time of choice \citep{BaronHershey1988}.

Importantly, this work does not aim to improve predictive accuracy, automate decision-making, or replace human expertise. Instead, it focuses on auditing the quality of human judgment under uncertainty by separating epistemic justification from stochastic outcomes. From this perspective, the limitations of predictive systems are not merely due to data insufficiency, but reflect a normative flaw inherent to supervised learning in agentic contexts. \emph{We argue that this limitation is fundamental: in the absence of external normative signals, decision imitation cannot, even asymptotically, correct systematic human bias.} By mapping contextual inputs to historical decisions, such systems inevitably reproduce embedded cognitive biases, including \emph{status quo bias} (reluctance to change strategy) and \emph{sunk cost fallacy} (persistence in failing courses of action). Consequently, predictive models tend to validate conservative human behavior rather than reveal when deviation is epistemically warranted.

To address this limitation, we distinguish between \emph{Predictive AI}, which forecasts events or replicates past decisions, and \emph{Prescriptive AI}, defined here as a class of systems whose primary function is to audit, justify, and support human decision-making under uncertainty. This perspective aligns with decision-theoretic and epistemic accounts of agency, in which rational choice is defined as a coherent state transition at time $t$, independent of the stochastic outcome realized at time $t+n$ \citep{vanBenthem2011}. While prescriptive analytics and interpretable decision-support frameworks have been advocated in theory \citep{Bertsimas2020, Lepenioti2020, Rudin2019}, empirical demonstrations in realistic, adversarial environments remain scarce. Black-box models are particularly ill-suited for such auditing tasks due to automation bias and opacity \citep{parasuraman2010}, motivating the need for systems that explicitly map observed states to epistemically justifiable actions.

In this work, we use elite-level soccer as a demanding natural laboratory to operationalize the prescriptive auditing paradigm. Substitution decisions share structural properties with many safety-critical contexts: they are time-sensitive, irreversible, and made under pervasive uncertainty. Existing approaches in this domain are limited either by insufficient temporal resolution—introducing \emph{exposure bias} by favoring cumulative playing time over efficiency \citep{Pappalardo2019, SchmidtLilloBustos2024}—or by a reliance on purely predictive modeling that fails to surface normative tactical risk \citep{wu, pamukkale}. To establish that the same axiomatic structure is not an artifact of the sporting domain, we complement this stress test with two independent historical instantiations from clinical medicine—MYCIN and NEWS2 (Section~\ref{sec:mycin_news2})—neither of which was engineered with the present framework in mind.

To enable systematic auditing of human judgment, we propose a hybrid statistical--symbolic framework that overlays intrinsically interpretable fuzzy reasoning onto robust statistical signals. Rather than automating decisions, the system functions as an auditing layer, surfacing decision-relevant risk—such as performance decay or defensive liability—before it manifests as observable failure. Unlike descriptive fuzzy models \citep{Marliere2017, zhou} or static ranking systems \citep{salabun}, the proposed approach operationalizes continuous, role-aware evaluation of evolving decision states.

This work makes five contributions. First, we formalize a prescriptive auditing framework for evaluating human decision-making under uncertainty, grounded in domain-independent axioms and separation results that distinguish auditing from prediction, together with an explicit model-theoretic semantics for the contestability axiom. Second, we strengthen the Imitation Incompleteness result to hold under standard M-estimator regularity conditions, extend it to Markovian (non-i.i.d.) decision logs, and give an explicit finite-sample concentration bound. Third, we establish a second, independent impossibility result—\emph{Normative Non-Identifiability}—showing that the normative utility underlying rational choice cannot be recovered from behavioral data under systematic bias, and that this failure is \emph{directional} rather than symmetric once a concrete estimator (MaxEnt IRL) is fit. Fourth, we introduce a role-aware cumulative mean metric that eliminates play-time exposure bias, enabling principled detection of intra-episode performance deterioration. Fifth, we demonstrate the computational realizability of the proposed prescriptive framework across three heterogeneous, independently engineered systems spanning five decades—a fuzzy-logic soccer substitution auditor, the MYCIN clinical consultation system, and the NEWS2 early-warning protocol—showing that it systematically reveals latent risk patterns overlooked by both human experts and black-box predictive models.

\section{Research Context and Review}

This section situates the present work within the broader literature on decision theory, prescriptive analytics, and human-centered decision support. Beyond surveying domain-specific applications in sports analytics, the review deliberately incorporates foundational results from decision theory and cognitive science that formalize how decisions should be evaluated under uncertainty and how humans systematically deviate from normative rationality. These theoretical contributions provide the normative and cognitive grounding for interpreting Prescriptive AI not merely as an optimization technology, but as an auditing mechanism designed to compensate for structural limitations in human real-time decision-making.
\clearpage
\subsection{Scope Summary of Reviewed Works}

\begin{table}[h]
\centering
\small
\begin{tabular}{p{5cm} p{10cm}}
\hline
\textbf{Category} & \textbf{Representative Works} \\
\hline
\textit{Theoretical Foundations} & \\
Prescriptive Analytics \& Optimization & \cite{Lepenioti2020}; \cite{Davenport2007}; \cite{Bertsimas2020}; \cite{Wissuchek2022}; \cite{Hullermeier2021} \\
Decision Support \& Agency & \cite{Power2002}; \cite{Panda2021}; \cite{Sun2020}; \cite{BaronHershey1988}; \cite{parasuraman2010} \\
Logic of Agency \& Information Dynamics & \cite{vanBenthem2007}; \cite{vanBenthem2011}; \cite{Halpern2017} \\
Normative Decision Theory \& Bounded Rationality & \cite{Savage1954}; \cite{Simon1955}; \cite{HogarthEinhorn1992}; \cite{GigerenzerGaissmaier2011}; \cite{parasuraman2000}; \cite{Endsley1995}; \cite{Kahneman1979}\\
Imitation Learning, Offline RL \& IRL & \cite{Pomerleau1991}; \cite{Ross2010}; \cite{Ross2011}; \cite{Levine2020}; \cite{NgRussell2000}; \cite{ZiebartEtAl2008}; \cite{Abbeel2004}; \cite{CaoEtAl2021}; \cite{SchlaginhaufenKamgarpour2023}; \cite{SkalseAbate2024}; \cite{ShahEtAl2019}\\
Counterfactual and Off-Policy Evaluation & \cite{Dudik2011}; \cite{Swaminathan2015}\\
Algorithmic Fairness & \cite{Hardt2016}; \cite{Kusner2017}; \cite{Zhang2018}\\
Machine Ethics \& Normative Precedents & \cite{Asimov1942}; \cite{AndersonAnderson2007}; \cite{AwadEtAl2018}\\
\hline
\textit{Domain Applications} & \\
Contextual Tactical Reasoning & \cite{zhou}; \cite{vallejo}; \cite{Marliere2017} \\
Fuzzy Individual Evaluation & \cite{zeng}; \cite{bazmara}; \cite{salabun}; \cite{onwuachu}; \cite{LACCEI2023} \\
Substitution Analysis \& ML & \cite{goes}; \cite{power}; \cite{pamukkale}; \cite{wu}; \cite{mohandas} \\
Performance Metrics & \cite{Pappalardo2019}; \cite{SchmidtLilloBustos2024} \\
Clinical Decision Support \& Early-Warning Systems & \cite{Shortliffe1975}; \cite{Yu1979}; \cite{RCP2017}; \cite{MartinRodriguez2019}\\
\hline
\textit{Methodological Pillars} & \\
Explainable AI (XAI) & \cite{Rudin2019}; \cite{Miller2019}; \cite{DoshiVelezKim2017}; \cite{Lipton2018}; \cite{Guidotti2018}; \cite{Holzinger2022} \\
\hline
\end{tabular}
\caption{Scope-oriented categorization of related work, spanning theoretical foundations, domain-specific applications, and methodological pillars.}
\end{table}

\subsection{Theoretical Framework: Prescriptive AI and Decision Agency}

To rigorously position the proposed system within the existing literature, it is necessary to distinguish between predictive and prescriptive paradigms of decision support. Prescriptive Analytics is commonly described as the highest stage of analytics maturity, addressing the question ``what should be done?'' through optimization, simulation, and rule-based reasoning \cite{Davenport2007,Lepenioti2020}. Recent systematic reviews confirm that while predictive models estimate future probabilities, prescriptive systems explicitly map observed states to recommended actions \cite{Wissuchek2022,Bertsimas2020}, evolving historically from static Decision Support Systems (DSS) toward more dynamic and intelligent decision agents \cite{Panda2021}.

Despite this evolution, a critical gap persists in how prescriptive systems are evaluated in stochastic environments. Traditional approaches, including standard reinforcement learning (RL) agents \cite{SuttonBarto2018}, often exhibit \textit{outcome bias}, in which the quality of a decision is judged primarily by its realized result rather than by the reasoning process that produced it \cite{BaronHershey1988}. In high-stakes domains involving human decision-makers, such retrospective evaluation is frequently insufficient. Moreover, the well-documented risk of \textit{automation bias}, whereby users over-rely on opaque algorithmic recommendations \cite{parasuraman2010}, further motivates the need for prescriptive systems that function as interpretable decision supporters  rather than black-box oracles.

At a conceptual level, these concerns resonate with work in the logical dynamics of information and agency, particularly van Benthem's account of decision-making as an epistemic state transition under uncertainty \citep{vanBenthem2007,vanBenthem2011}, and more broadly with axiomatic treatments of reasoning under uncertainty that unify probabilistic and epistemic-logical accounts of belief and action \citep{Halpern2017}. Within this tradition, rationality is not defined by the optimality of outcomes, but by the coherence of an action with respect to the informational state and constraints available at the moment the decision is taken. Decisions are thus evaluated ex ante, independently of the stochastic realization of downstream consequences.

This perspective provides a theoretical basis for separating decision quality at time $t$ from observed outcomes at time $t+n$, a distinction that has been increasingly explored across research communities concerned with decision agency, interpretability, and accountability. Rather than focusing exclusively on outcome prediction, these lines of work emphasize the importance of assessing whether decisions are contextually justified given the available evidence and risk structure, particularly in environments characterized by uncertainty, irreversibility, and delayed feedback.

From a normative standpoint, this ex-ante evaluation of decision quality is rooted in classical decision theory. Savage's axiomatization of rational choice formalizes decision optimality as a function of expected utility conditional on the informational state available at time $t$, independently of future realizations \cite{Savage1954}. In this sense, van Benthem defines the epistemic state underlying a decision, while Savage provides the normative evaluation over that state; a formal correspondence between the two, showing that a \prescAI{} recommendation implementing $\argmax_a\mathbb{E}_P[U(s,a)]$ realizes exactly the ex-ante decision rule that Savage's representation theorem attributes to a rational agent given $(U,P)$, is given in Proposition~4. Complementarily, the theory of bounded rationality introduced by Simon \cite{Simon1955} demonstrates that human decision-makers are structurally incapable of performing such continuous optimization under time pressure. Empirical models of belief updating further show that sequential evidence integration is systematically biased by order and salience effects \cite{HogarthEinhorn1992}, prospect-theoretic accounts document systematic, reference-dependent departures from expected-utility maximization under risk \cite{Kahneman1979}, and fast-and-frugal heuristics dominate real-time human judgment under uncertainty \cite{GigerenzerGaissmaier2011}. Together, these results reinforce the need for prescriptive systems that evaluate decision states normatively while preserving human agency, in line with established taxonomies of human--automation interaction \cite{parasuraman2000} and theories of situation awareness \cite{Endsley1995}.

\subsection{Imitation Learning, Offline Reinforcement Learning, and the Limits of Behavioral Supervision}
\label{sec:imitation-related}

A distinct but closely related literature addresses the statistical limits of learning policies directly from demonstrated behavior. \textit{Behavioral cloning} fits $P(a\mid s)\approx\pih(a\mid s)$ by supervised learning on logged state-action pairs and is known to suffer compounding distribution shift at deployment time, since small errors accumulate as the induced policy drifts into states underrepresented in the training log \citep{Pomerleau1991,Ross2010,Ross2011}. Crucially for our purposes, even absent distribution shift, behavioral cloning converges---by construction---to the demonstrating policy $\pih$ itself, not to a normatively optimal policy; Section~\ref{sec:imitation-incompleteness} formalizes this as a structural, sample-independent ceiling rather than a finite-sample artifact correctable with more data.

\textit{Offline reinforcement learning} \citep{Levine2020} instead fits a value or reward estimate to logged $(s,a,r)$ triples and optimizes a policy against it. This substitutes one problem for another: because the reward signal $r$ is itself derived from realized outcomes, offline RL reintroduces outcome bias by construction, and is further subject to extrapolation error and reward hacking, whereby the learned policy over-optimizes an imperfect proxy in ways that do not track the intended normative objective \citep{FujimotoEtAl2019,KumarEtAl2020,RewardHacking2025}. In low-scoring, high-variance, or otherwise sparse-outcome domains---precisely the setting of soccer substitution decisions or acute clinical deterioration---the outcome-derived reward is a particularly weak and delayed training signal, compounding both problems simultaneously. \textit{Counterfactual risk minimization} and doubly robust estimators \citep{Dudik2011,Swaminathan2015} relax some of these issues but require richer logged bandit feedback $(s,a,r)$ than a pure behavioral log provides, and remain fundamentally outcome-based, hence still subject to the individual-decision labeling impossibility formalized in Theorem~5. We treat all of these as legitimate techniques for implementing specific components of the GNPAF architecture (Section~\ref{sec:gnpaf}) when richer logs are available, rather than as substitutes for an externally specified normative criterion.

\subsection{Inverse Reinforcement Learning and Normative Recovery}
\label{sec:irl-related}

A closely related question is whether the normative criterion itself---rather than a policy---can be recovered from observed behavior. \emph{Inverse reinforcement learning} (IRL) \citep{NgRussell2000} and its maximum-entropy formulation \citep{ZiebartEtAl2008}, together with apprenticeship-learning variants \citep{Abbeel2004}, seek a reward function under which the observed policy is (approximately) optimal. This line of work assumes an approximately optimal demonstrator, an assumption that fails precisely in the setting motivating this paper: systematic, not merely noisy, human bias. \citet{NgRussell2000} themselves note that the recovered reward is not unique---a policy is consistent with an entire affine family of rewards---but this classical non-identifiability is \emph{symmetric}: no member of the family is privileged over any other by the data. Section~\ref{sec:non-identifiability} shows that once the demonstrator is systematically, rather than merely noisily, suboptimal, non-identifiability takes a qualitatively different and \emph{directional} form: the true normative utility is provably excluded from the class of rewards consistent with the data, and a concrete, standard estimator (MaxEnt IRL) can be shown to converge, almost surely, to a specific member of that class that rationalizes the bias itself. Related results in the IRL identifiability literature reinforce this picture from complementary angles: reward recovery is ill-posed even under an optimal demonstrator \citep{CaoEtAl2021}, worsens under additional safety constraints \citep{SchlaginhaufenKamgarpour2023}, is highly sensitive to misspecification of the assumed behavioral model \citep{SkalseAbate2024}, and modeling the demonstrator's bias explicitly, rather than assuming approximate optimality, has been shown to cost more in sample complexity than it recovers in accuracy \citep{ShahEtAl2019}. Jointly, these results support the position at the core of the present framework: the normative criterion $\calN$ should be specified externally by domain authority (Axiom~2, Component~C3 of GNPAF), not inferred from behavioral logs.

\subsection{Algorithmic Fairness and Machine Ethics}
\label{sec:fairness-ethics}

Algorithmic fairness methods \citep{Hardt2016,Kusner2017,Zhang2018} correct for bias at the level of a model's outputs---equalizing error rates or satisfying a counterfactual invariance criterion across protected groups---without addressing the fact that the training data itself may encode the very decision bias being corrected for. \prescAI{} generalizes this concern structurally: rather than post-hoc correcting a fitted model's outputs, it requires that the normative criterion $\calN$ never be fit to biased behavioral data in the first place (Axiom~3, Component~C3). A longer history of proposals for fixed, externally imposed normative constraints on machine behavior---most famously Asimov's Three Laws \citep{Asimov1942}---illustrates both the appeal and the difficulty of this approach: the well-documented failure mode of the Laws, namely that superficially precise natural-language rules admit conflicting or under-determined interpretations at the point of application \citep{AndersonAnderson2007}, is itself an instance of the specification problem formalized by Theorem~6 below---a normative criterion cannot be recovered from, or fully validated against, observed behavior alone, and must instead be made explicit, inspectable, and contestable (Axiom~4). Large-scale empirical work on moral preferences, most notably the Moral Machine experiment \citep{AwadEtAl2018}, which collected millions of pairwise trolley-problem judgments across 233 countries, found systematic and culturally varying disagreement with no consensual aggregate ordering---independent evidence that normative content cannot be recovered by aggregating preference data without losing precisely the disagreement that matters, reinforcing our position that $\calN$ should be treated as an externally supplied, contestable artifact rather than a statistically estimated quantity (see Section~\ref{sec:ustar-discussion}).

\subsection{Contextual Interpretation and Tactical Reasoning}

Several studies model contextual indicators to support tactical interpretation in sports. \cite{zhou} employ fuzzy contextual reasoning to produce action-oriented decisions, mapping continuous match descriptors into symbolic control outputs. Similarly, \cite{Marliere2017} applies fuzzy control systems to tactical arbitration under uncertainty. In contrast, \cite{vallejo} focus on the semantic recognition of tactical actions using fuzzy models, remaining descriptive rather than prescriptive. While these contributions advance contextual interpretation, they neither integrate individual performance evaluation nor address substitution decisions.

\subsection{Fuzzy-Based Individual Performance Evaluation}

A complementary line of research applies fuzzy logic to individual player evaluation. \cite{zeng} and \cite{bazmara} employ fuzzy inference systems to assess player suitability and generate rankings. Subsequent works, such as \cite{salabun} and \cite{onwuachu}, extend this paradigm through fuzzy and neuro-fuzzy models to synthesize performance-related variables. \cite{LACCEI2023} further demonstrate the suitability of fuzzy logic for transforming physical indicators into interpretable assessments. These approaches provide interpretability but generally lack temporal dynamics or prescriptive substitution reasoning.

\subsection{Substitution Analysis: From Prediction to Prescription}

Research on player substitutions has traditionally followed two paths: observational analysis and predictive modeling. Studies such as \cite{goes}, \cite{power}, \cite{pamukkale}, and \cite{wu} analyze substitutions using statistical, observational, or causal methods to understand timing and patterns. While valuable for post-match analysis, they do not directly support real-time decision-making.

More recently, \cite{mohandas} framed substitutions as a supervised prediction problem, applying machine learning models to estimate when a substitution is likely to occur. Despite achieving predictive accuracy of approximately 70\%, this line of work emphasizes imitation of historical decisions rather than prescriptive optimization. Unlike Reinforcement Learning approaches that maximize a reward signal \cite{SuttonBarto2018}, which can be noisy in low-scoring sports, our approach focuses on the normative evaluation of the \textit{need} for substitution, independent of the match's stochastic outcome.

\subsection{Performance Evaluation and Temporal Exposure Bias}

A central challenge in multi-agent environments is quantifying individual contributions. The PlayeRank framework \cite{Pappalardo2019} defines a multidimensional metric validated against professional scouts. Although effective, its cumulative formulation reintroduces exposure bias, preventing the detection of performance decay or momentum reversals. Recent work by \cite{SchmidtLilloBustos2024} demonstrates that player influence fluctuates meaningfully during a match due to tactical factors. Building on this, the present work adopts a role-aware cumulative mean over fixed temporal slices, addressing the exposure bias inherent in cumulative sums.

\subsection{Clinical Decision Support and Rule-Based Expert Systems}
\label{sec:clinical-related}

Rule-based expert systems for clinical consultation predate learned decision support entirely. MYCIN \citep{Shortliffe1975} produced antimicrobial-therapy recommendations from an externally specified corpus of clinician-authored rules with attached certainty factors, never fit to any single hospital's own prescribing records, and inspectable and revisable by the treating physician through dedicated WHY/HOW explanation and rule-acquisition subprograms. \citet{Yu1979} subsequently evaluated MYCIN's recommendations against nine practicing prescribers on ten real meningitis cases using a blinded panel of eight infectious-disease experts, finding MYCIN's regimens rated at least as acceptable as those of five faculty specialists---external, decades-old evidence that a non-imitative, externally specified rule base need not underperform the clinicians it audits. The National Early Warning Score 2 \citep{RCP2017}, a fixed physiological scoring protocol mandated across acute NHS trusts since 2019, offers a structurally analogous, more recent example, whose predictive validity for early in-hospital mortality has been confirmed in a prospective multi-center cohort \citep{MartinRodriguez2019}. We revisit both systems in Section~\ref{sec:mycin_news2} not as systems to be improved upon, but as independently engineered, independently validated historical instances against which the six structural components of GNPAF (Section~\ref{sec:gnpaf}) can be checked directly.

\subsection{Symbolic Reasoning and Explainable AI (XAI)}

The interpretation of imprecise information is critical for accountability. While ``black-box'' models dominate distinctive tasks, they are often unsuitable for high-stakes decision auditing \cite{Rudin2019}. Foundational works by \cite{DoshiVelezKim2017}, \cite{Lipton2018}, and \cite{Miller2019} emphasize the distinction between interpretability and post-hoc explanation. Surveys by \cite{Guidotti2018} and \cite{Holzinger2022} reinforce the need for transparency in human-centric AI. \cite{Rudin2019} specifically warns against explaining black boxes in high-stakes contexts, advocating for models where reasoning is transparent by design. This principle motivates our use of Fuzzy Logic not just as a controller, but as a semantic layer that ensures the system's recommendations are contestable and auditible by the human coach.

\subsection{Relation to Prescriptive Machine Learning}
\label{sec:hullermeier}

\citet{Hullermeier2021} introduces prescriptive machine learning as a paradigm distinct from predictive modeling, organized around five methodological challenges for \emph{learning} prescriptive models from data: (i) weak supervision from biased historical decisions; (ii) complex, multi-criteria prescriptions and performance criteria; (iii) representing uncertainty in the prescription itself; (iv) complexity and resource constraints on the decision-maker; and (v) ethics and fairness. The present work is complementary and operates at a different level of abstraction: where \citeauthor{Hullermeier2021} identifies these as challenges to be solved by better learning algorithms, we provide an axiomatic characterization of what any system---learned or not---must structurally satisfy to qualify as prescriptive at all. Concretely, Challenge~(i) is the target of Theorem~2 (Imitation Incompleteness), which shows the problem is not merely weak supervision but a structural impossibility: no imitation-based system can correct systematic bias without an externally supplied normative signal $U$, and Theorem~6 (Normative Non-Identifiability) further shows $U$ cannot itself be recovered from behavioral data once such bias is present. Challenge~(ii) is addressed structurally by GNPAF's Component~C2 (state-dependent admissible action sets) and Component~C3 (a multi-criteria normative mapping that need not collapse to a single scalar reward). Challenge~(iii) is addressed jointly by Axiom~1 and Theorem~5, which show that individual binary outcome-based labeling is ill-defined in stochastic domains, without foreclosing continuous-variable estimation or population-level calibration. Challenge~(iv) is formalized by Axiom~4 (Contestability), operationalizing bounded rationality \citep{Simon1955}; Theorem~7 shows black-box systems cannot satisfy it by construction. Challenge~(v) is addressed at the level of framework design, by requiring $\calN$ to be specified externally and explicitly (Component~C3) and preserving human agency through Axiom~4, which prevents the false objectivity of purely outcome-optimized systems. A substantive difference in scope remains: \citeauthor{Hullermeier2021} remains rooted in the machine-learning paradigm, whereas the axioms proposed here encompass any architecture satisfying Axioms~1--4, including symbolic rule systems such as MYCIN, fuzzy inference engines, and constraint-based reasoners, for which the very notion of a learned reward may be inapplicable.

\subsection{Identified Research Gap}

Despite substantial progress in both general analytics and sports science, the literature reveals three critical, intersecting gaps that this work addresses.

\textbf{1. Lack of Formal Definition for Prescriptive AI:}
First and foremost, there is a lack of a formalized, normative definition for ``Prescriptive AI.'' While the term appears in recent industrial reports and applied solutions (e.g., \cite{Sun2020}), the academic literature often conflates \textit{prescription} with \textit{automation} or simple \textit{prediction}. There is virtually no framework that defines Prescriptive AI not just as a technological stack, but as a normative agent responsible for auditing the decision logic itself \cite{Lepenioti2020, Wissuchek2022}.

\textbf{2. Methodological Limitations in Dynamic Environments:}
Standard approaches to generating actionable insights, such as Reinforcement Learning, struggle with \textit{outcome bias} \cite{BaronHershey1988}. In low-scoring, high-variance domains like soccer, a ``good'' decision can lead to a bad result (and vice-versa). Existing systems that rely on maximizing reward signals \cite{SuttonBarto2018} or mimicking historical substitutions \cite{mohandas} fail to decouple the \textit{quality of the decision state} from the stochastic \textit{outcome}, rendering them unreliable for objective auditing.

\textbf{3. Absence of Interpretable Decision Support Validated Across Domains:}
Finally, the domain-specific literature remains fragmented. Soccer analytics focuses heavily on descriptive metrics \cite{Pappalardo2019} or predictive modelling, neglecting the prescriptive auditing of tactical decisions; and where interpretable rule-based auditors do exist historically, as in clinical medicine \citep{Shortliffe1975,RCP2017}, they have not previously been connected to a common axiomatic account that would let their structural similarity to a soccer substitution auditor be recognized and verified. No prior work jointly integrates (i) a temporally resolved, exposure-aware performance metric, (ii) intrinsic interpretability that mitigates automation bias \cite{parasuraman2010}, and (iii) a fuzzy reasoning engine capable of handling qualitative context, while also (iv) showing that the same axioms are independently satisfied by unrelated, previously validated systems in an entirely different domain.

The present work addresses these gaps by unifying temporal evaluation with symbolic decision modeling, effectively proposing a formal structure for Prescriptive AI: a system that evaluates \textit{decision quality} via interpretable norms, independent of the stochastic match outcome, and by showing, via MYCIN and NEWS2 (Section~\ref{sec:mycin_news2}), that the same structure is not specific to sport.

\section{A Normative Taxonomy of Action-Oriented Decision Systems: Formalizing Prescriptive AI}

The ability to recommend or evaluate actions has been addressed across multiple research traditions and industrial domains under diverse labels, including decision optimization, action recommendation, prescriptive analytics, and AI-driven decision support. These approaches are often discussed as if they constituted a coherent class of systems. In reality, they occupy a heterogeneous and only loosely structured conceptual space, within which crucial distinctions regarding uncertainty, accountability, and human agency are frequently overlooked.

This section establishes a coherent conceptual hierarchy for action-oriented decision systems, characterizing the broad superset of Action-Oriented Decision Systems, positioning Prescriptive Analytics as a structured subset within that space, and defining Prescriptive AI as a normative, high-stakes subset of Prescriptive Analytics defined by its epistemological role in human decision-making rather than by algorithmic choices.
\clearpage
\begin{figure}[t]
    \centering
    \resizebox{\linewidth}{!}{%
    \begin{tikzpicture}[scale=0.8, every node/.style={align=center}]
        % 1. Superset: Action-Oriented Decision Systems
        \draw[fill=gray!8, draw=gray!70, thick] (0,1) ellipse (8.5cm and 7cm);
        
        \node[text=gray!80, font=\bfseries, text width=7cm] at (0, 6.5) 
        {Action-Oriented Decision Systems\\(The Heterogeneous Superset)};
        \node[font=\footnotesize, text=gray!60, text width=5cm] at (0, 5.0) 
        {Predictive Workflows\\(fixed trigger $\to$ action)};
        \node[font=\footnotesize, text=gray!60, text width=3cm] at (-6.3, 3.5) 
        {Naive\\Recommenders};
        \node[font=\footnotesize, text=gray!60, text width=3cm] at (6.3, 3.5) 
        {Threshold-Triggered\\Automation};
        % 2. Subset: Prescriptive Analytics
        \draw[fill=blue!5, draw=blue!50, thick] (0,-1) ellipse (6.5cm and 5cm);
        
        \node[text=blue!60, font=\bfseries, text width=5cm] at (0, 2.5) 
        {Prescriptive Analytics\\(Decision-Oriented)};
        \node[font=\footnotesize, text=blue!50, text width=2.6cm] at (-4.4, 1.0) 
        {Reinforcement\\Learning};
        \node[font=\footnotesize, text=blue!50, text width=2.6cm] at (4.4, 1.0) 
        {Recommender /\\Next-Best-Action};
        \node[font=\footnotesize, text=blue!50, text width=2.6cm] at (-4.0, -1.2) 
        {Black-Box\\Optimization};
        \node[font=\footnotesize, text=blue!50, text width=2.4cm] at (4.5, -1.2) 
        {Simulation-Based\\Planning};
        % 3. Normative Core: Prescriptive AI
        \draw[fill=blue!20, draw=blue!80, thick] (0,-2.5) circle (3cm);
        
        \node[text=blue!90, font=\bfseries] at (0,-2.0) 
        {Prescriptive AI};
        \node[font=\footnotesize, text=blue!90] at (0,-2.6) 
        {Auditable Decision Systems};
        \node[font=\footnotesize, text=blue!90] at (0,-3.1) 
        {Hybrid Interpretable DSS};
        \node[font=\footnotesize, text=blue!90] at (0,-3.6) 
        {Risk-Aware Reasoning};
    \end{tikzpicture}%
    }
    \caption{\textbf{A normative taxonomy of Action-Oriented Decision Systems} 
    The diagram illustrates the hierarchical relationship where Action-Oriented Decision Systems form the superset. 
    Prescriptive Analytics sits within as a decision-focused subset, while Prescriptive AI serves as the normative core, defined by auditability and interpretability requirements.}
    \label{fig:taxonomy_diagram}
\end{figure}
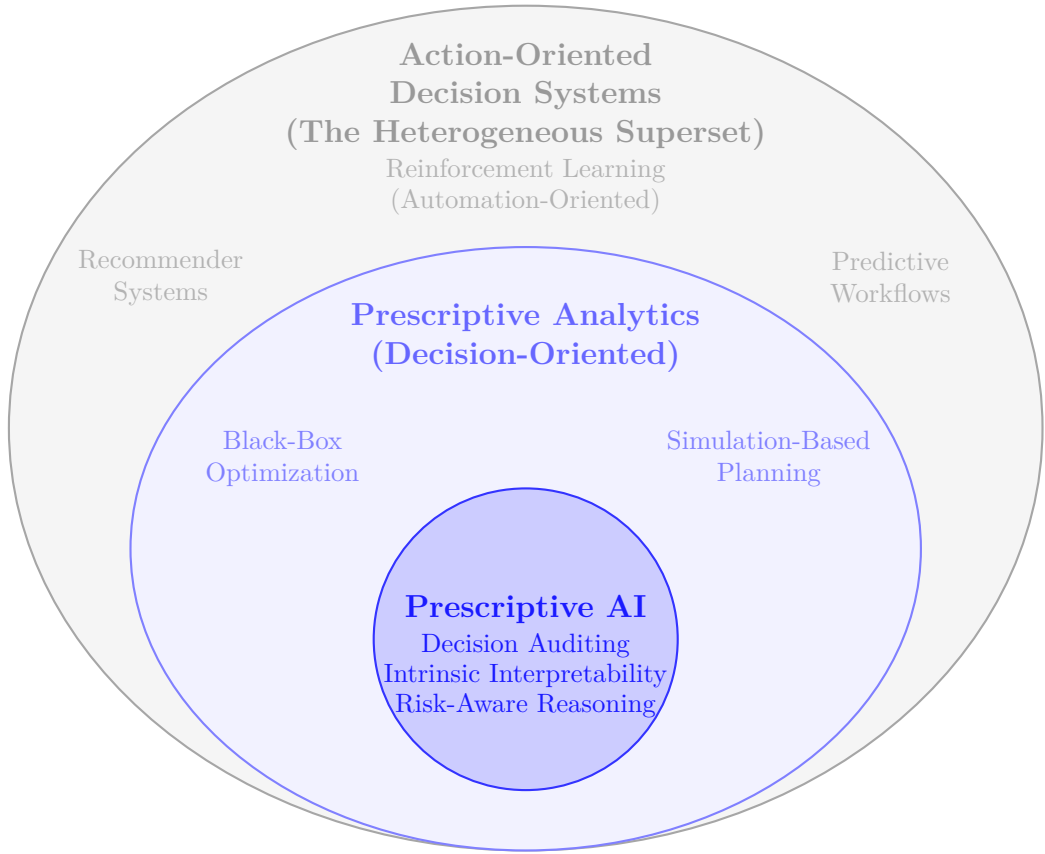

\subsubsection{Action-Oriented Decision Systems: A Heterogeneous Superset}

Action-Oriented Decision Systems form a broad superset of computational approaches whose shared property is the capacity to recommend, prioritize, or execute actions. This superset includes, among others:

\begin{itemize}
\item Optimization and mathematical programming systems;
\item Simulation-based and what-if analysis tools;
\item Rule-based and expert systems;
\item Reinforcement Learning (RL) agents;
\item Recommendation and next-best-action systems;
\item AI-augmented decision support tools embedding predictive models into workflows.
\end{itemize}

Despite their shared action-producing nature, these systems differ substantially in objectives, assumptions, and interaction with human decision-makers. Many are designed primarily for automation or outcome maximization, often assuming stable objectives, well-defined environments, and limited requirements for interpretability or accountability. Consequently, action-oriented systems do not constitute a single paradigm, but rather a conceptually heterogeneous collection of approaches.

\subsubsection{Prescriptive Analytics as a Decision-Oriented Subset}

Within this heterogeneous landscape, \textbf{Prescriptive Analytics} provides a more structured and widely recognized framework. Originating in business intelligence and operations research, prescriptive analytics focuses on determining what actions should be taken to achieve specified objectives, typically through optimization, simulation, and rule-based reasoning \cite{Davenport2007, Lepenioti2020, Bertsimas2020}.

Prescriptive Analytics narrows the scope of Action-Oriented Decision Systems by explicitly addressing decision recommendation. However, it remains largely agnostic with respect to how uncertainty, human judgment, or accountability should be handled. It encompasses both automated and decision-support systems and does not inherently distinguish between optimizing outcomes and supporting human deliberation.

Accordingly, Prescriptive Analytics is best understood as a decision-oriented subset of Action-Oriented Decision Systems: analytically grounded, but normatively under-specified.

\subsubsection{Formal Definition of Prescriptive AI}

Within Prescriptive Analytics, we define \textbf{Prescriptive Artificial Intelligence (Prescriptive AI)} as a specialized normative subset designed explicitly for high-stakes, human-in-the-loop decision-making.

\paragraph{Definition (Prescriptive System).}
Let $\calS$ denote the state space, $\calA$ a (finite) action space, $\mathcal{O}_{t+n}$ the space of possible future outcome realizations, $\calN$ a set of externally specified normative criteria, and $\calL$ a logical system used to express explicit justifications. A prescriptive system $\mathcal{S}$ is a recommendation function
\[
R:\calS\to\calA
\]
satisfying Axioms~1--4 below, agnostic to how $R$ is computed (symbolic, statistical, or hybrid), to the structure of $\calS$ or $\calA$ (discrete, continuous, or structured), and to whether the system learns, adapts, or remains static. \prescAI{} is a subclass of what is traditionally referred to as prescriptive analytics within the broader class of Action-Oriented Decision Systems, formally characterized by a normative decision operator whose role is to determine optimal actions and audit decision coherence under uncertainty, rather than to automate decisions or predict outcomes.

\paragraph{Norm-Invariant Recommendation and Auditing}
Prescriptive AI employs a single normative decision framework for
both recommendation and auditing. In prospective use, the system
operates online, producing action recommendations from the
epistemic state available at decision time. In retrospective
analysis, the same framework is applied offline to historical
records in a backtesting setting.

Auditing consists of reconstructing, from logged data, the
information that was available at the moment a decision was made
and recomputing the recommendation that would have been generated
under the same normative criteria. Future observations are used
exclusively to denoise or complete past measurements (e.g.,
correcting sensor errors or filling missing values), and never to
introduce outcome knowledge.

Decision coherence is then assessed by comparing the reconstructed
normative recommendation with the action actually taken. No
separate temporal operator or learning procedure is introduced:
auditing is simply norm-invariant backtesting. The same
prescriptive logic governs both real-time recommendation and
retrospective evaluation; only the data access mode differs.

Prescriptive AI systems are not characterized by the techniques
they employ, but by the normative role they play in the decision
process. In particular, they are distinguished by the following
defining properties:

\begin{itemize}
 \item A primary function of \textbf{decision recommendation and decision auditing} rather than decision automation;
\item Explicit reasoning under uncertainty and asymmetric risk;
 \item \textbf{Intrinsic interpretability} as a structural requirement, not a post-hoc feature \cite{Rudin2019};
 \item \textbf{Contestable recommendations} that preserve human agency;
 \item Evaluation criteria that \textbf{decouple state assessment from outcome realization}, explicitly avoiding outcome bias \cite{BaronHershey1988}.
\end{itemize}
In its lexical sense, a ``prescription'' is the action of laying down authoritative rules or directions, or more generally, something prescribed as a rule \cite{MerriamWebsterPrescription}. By this definition, a prescriptive statement is inherently normative: it asserts what ought to be done rather than merely predicting or suggesting an outcome. Consequently, any system that cannot be inspected, audited, or contested cannot properly prescribe.

Taken together, these properties establish a necessary boundary: a system that cannot be audited cannot prescribe; it can only suggest or automate.
These properties impose normative constraints that exclude many systems commonly labeled as prescriptive. Prescriptive AI therefore constitutes a proper subset of Prescriptive Analytics, defined by epistemological and accountability requirements rather than by optimization or learning paradigms.

\subsubsection{Decision-Making as Epistemic State Transition}
\label{sec:epistemic_foundations}

The normative necessity of Prescriptive AI follows from a dynamic conception of rationality. In the framework of logical dynamics developed by van Benthem, decision-making is not primarily evaluated by realized outcomes, but by the epistemic state transitions induced by actions under partial and evolving information \citep{vanBenthem2007, vanBenthem2011}, a perspective compatible with axiomatic treatments unifying probabilistic and epistemic reasoning about uncertainty and agency \citep{Halpern2017}. Rationality, under this view, is a property of how an agent updates beliefs, constraints, and commitments at the moment of action, rather than of the stochastic realization of downstream consequences.

This perspective provides a formal grounding for the core design principles of Prescriptive AI. If decision quality is defined at time $t$, prior to outcome realization at time $t+n$, then systems designed to support or evaluate human decisions must operate on the informational state available at the moment of commitment. Predictive accuracy and ex post outcome optimization are therefore insufficient as normative criteria. A prescriptive system must instead audit whether a decision constitutes a coherent, risk-aware, and justifiable epistemic update given the available evidence and constraints.

From this standpoint, Prescriptive AI can be understood as an operational instantiation of epistemic action auditing. Rather than attempting to predict or automate behavior, the system evaluates the internal consistency of the decision state itself, including the alignment between observed signals, inferred risk, and permissible actions. This aligns directly with the logical-dynamic view in which actions are epistemic interventions that transform an agent's informational state.

While the present work does not implement the full formal machinery of Dynamic Epistemic Logic, it adopts this foundational stance by treating decision evaluation as a state-based rather than outcome-based problem. The logical dynamics literature does not prescribe a specific algorithmic realization; instead, it characterizes the object of evaluation. Prescriptive AI derives its epistemic effectiveness precisely from this characterization: by auditing epistemic transitions rather than stochastic outcomes, it enables principled assessment of decision quality under uncertainty, irreversibility, and asymmetric risk.

\subsubsection{Prescriptive AI vs. Predictive and Outcome-Driven Systems}

Predictive AI systems, typically trained via supervised learning, map contextual inputs to historical human decisions or observed outcomes. While effective for forecasting and pattern discovery, such models inherently learn and reproduce historical behavior, including embedded cognitive biases such as status quo bias, sunk cost fallacy, and outcome bias.

Prescriptive AI systems, by contrast, operate as \textbf{decision auditors}. Rather than imitating past decisions, they evaluate the current state against explicit objectives, constraints, and domain knowledge. Their purpose is to assess whether a decision is justified given the information available at time $t$, not whether it coincidentally leads to a favorable outcome at time $t+n$.

This distinction is critical in stochastic, high-stakes environments. Evaluating decision systems based on realized outcomes introduces outcome bias. A high-risk decision state is not retroactively validated by a favorable random outcome. Prescriptive AI explicitly decouples state evaluation from outcome realization, treating disagreement with historical decisions or subsequent outcomes not as error, but as potential identification of latent risk.

To make this distinction precise, it is useful to separate a decision system's \emph{provenance}---how its recommendation function was constructed---from its \emph{functional form}---whether that function satisfies the axioms below. Let $\Sigma=(\Sigma_{\mathrm{fn}},\Sigma_{\mathrm{prov}})$ denote a decision system pairing a recommendation function $\Sigma_{\mathrm{fn}}:\calS\to\calA$ (or $\Delta(\calA)$) with its provenance $\Sigma_{\mathrm{prov}}$, and let $Y\in\{A,\mathcal{O}_{t+n}\}$ range over observable log targets. We define $\textsc{PRED}:=\{\Sigma:\exists\,Y \text{ s.t. } \Sigma_{\mathrm{fn}} \text{ is a functional of an estimator } \hat P(Y\mid s_t)\}$, i.e., a claim about \emph{provenance}; and $\textsc{PRESC}:=\{\Sigma:\Sigma_{\mathrm{fn}}=R \text{ satisfies Axioms~1--4}\}$, a claim about $\Sigma_{\mathrm{fn}}$ itself. These two classes are logically independent, as Theorem~4 below shows.

\clearpage

\begin{table}[h]
\hspace{-2cm}
\small
\renewcommand{\arraystretch}{1.35}
\begin{tabular}{p{3.2cm} p{4.2cm} p{4.2cm} p{4.2cm}}
\toprule
\textbf{Dimension} 
& \textbf{Predictive AI} 
& \textbf{Prescriptive Analytics} 
& \textbf{Prescriptive AI} \\
\midrule
Primary Objective 
& Forecast events or actions 
& Recommend actions to optimize objectives 
& recommend normatively optimal actions and to audit decision coherence \\
Learning Paradigm
& Supervised / statistical learning from historical data
& Optimization, simulation, rules, or learned models
& Rule-based, symbolic, or hybrid  \\
Optimization Target 
& Accuracy / loss minimization 
& Expected utility or objective maximization 
& Risk mitigation / decision recommendation \\
Relation to Bias 
& Learns and reproduces historical bias 
& Implicitly reflects modeled objectives 
& Audits, constrains, and exposes bias \\
Evaluation Criterion 
& Predictive accuracy 
& Outcome-based performance 
& State-based validity / risk detection \\
Outcome Dependence
& Strong (labels and metrics)
& Strong or implicit
& Explicitly decoupled from outcomes \\
Interpretability
& Optional / post-hoc
& Often secondary
& Structural requirement \\
Role of Human 
& Source of labels 
& Executor or overseer 
& Accountable decision-maker \\
\bottomrule
\end{tabular}
\caption{Conceptual comparison between predictive AI, prescriptive analytics, and prescriptive AI paradigms. Prescriptive AI is distinguished not by a specific learning paradigm, but by its normative role in auditing and supporting human decisions under uncertainty.}
\label{tab:predictive_vs_prescriptive}
\end{table}

\subsection{Axiomatic Foundations of Prescriptive AI}

We formalize Prescriptive Artificial Intelligence as a distinct
decision-making paradigm defined by normative axioms rather than by
architectures, learning procedures, or representational choices.
Axioms~2 and~4 are universally required, while Axioms~1 and~3 are essential in
stochastic domains and become vacuous, though not violated, in fully deterministic ones.

\paragraph{Notation.}
Let $\calS$ denote the state space and $\calA$ the action
space. Let $\mathcal{O}_{t+n}$ denote the space of possible future outcome
realizations. Let $\calN$ denote normative criteria, including
utility functions, constraints, or domain rules---formally a finite set $\{\nu_1,\dots,\nu_k\}$ of functions $\nu_j:\calS\times\calA\to\mathbb{R}$ or predicates over $\calS\times\calA$, with $f(s_t,\calN)$ an explicit, domain-specified aggregation of $\{\nu_j(s_t,\cdot)\}_{j=1}^k$. Let $L$ denote a
logical system used to express explicit justifications.

\subsubsection{Axioms}

\paragraph{Axiom 1 (Outcome Decoupling).}
\label{ax:decoupling}
For any decision state $s_t \in \calS$ and any $n\in\mathbb{N}$, the recommendation is a function of $s_t$ alone:
\[
\mathcal{O}_{t+n}\notin\mathrm{dom}(R),\qquad\text{written}\qquad R(s_t) \perp O_{t+n}.
\]

\textbf{Note (sense of $\perp$).} The symbol $\perp$ denotes \emph{informational}, not \emph{statistical}, independence: it asserts that $R(s_t)$ does not take $\mathcal{O}_{t+n}$ as an argument, not that $R(s_t)$ and $\mathcal{O}_{t+n}$ are statistically uncorrelated as random variables. The two notions come apart whenever $s_t$ is itself informative about the distribution of $\mathcal{O}_{t+n}$, which is the typical case: a recommendation issued under a high-risk epistemic state is generally strongly predictive of $\mathcal{O}_{t+n}$ precisely \emph{because} it is a correct function of $s_t$, and Axiom~1 does not prohibit this; it prohibits $R$ taking a not-yet-realized $\mathcal{O}_{t+n}$ as an input. This axiom is essential in stochastic environments where outcome realization does not reliably indicate decision quality.

\paragraph{Axiom 2 (Epistemic Justification).}
\label{ax:justification}
For every recommendation $a = R(s)$, there exists an explicit
justification $E$, expressed in a logical system $L$, such that
\[
E \vdash_{L} (s \rightarrow a),
\]
where $L$ supports human inspection of the reasoning linking state
evaluation to action selection, and $E$ is accessible to the
decision-maker at decision time.

\textbf{Note:} The logical system $L$ may be classical propositional 
logic, fuzzy logic, probabilistic reasoning, structured natural 
language, or even implicit domain knowledge encoded during system 
design (e.g., literature-informed model architectures, expert-validated 
parameter choices). In the fuzzy soccer implementation (Section~\ref{sec:fuzzy_system}), $L$ 
denotes the Mamdani inference engine, and $E$ consists of activated 
rules and their membership degrees; in MYCIN (Section~\ref{sec:mycin_news2}), $L$ is the production-rule certainty-factor calculus and $E$ the traced firing sequence retrievable via WHY/HOW.

\paragraph{Axiom 3 (State-Based Evaluation).}
\label{ax:statebased}
Decision quality is evaluated as a function of the current state and
normative criteria:
\[
\text{Quality}(a) = f(s_t, \calN),
\]
and not as a function of realized outcomes $o_{t+n}$.

\textbf{Note:} In stochastic environments, this axiom ensures that 
decisions are evaluated by their epistemic justification at time $t$, 
independent of which stochastic realization occurs at time $t+n$. 
In the soccer implementation (Section~\ref{sec:fuzzy_system}), we instantiate $f$ as 
the fuzzy priority function: $\text{Quality}(a, s_t, N) := P_{\text{final}}(a, s_t)$ 
(Equation~2, Section~\ref{sec:fuzzy_system}).

\paragraph{Axiom 4 (Contestability).}
\label{ax:contestability}
Any recommendation $a = R(s)$ must satisfy:
\begin{enumerate}
    \item \textbf{Inspectability}: the justification $E$ is accessible
    and comprehensible to a human agent;
    \item \textbf{Challengeability}: the agent can question the validity
    of $E \vdash_L (s \rightarrow a)$;
    \item \textbf{Overridability}: the agent retains authority to select
    an alternative action $a' \neq a$ with documented justification.
\end{enumerate}
These predicates---$\mathrm{Inspectable}$, $\mathrm{Challengeable}$, $\mathrm{Overridable}$---are treated as primitives of the axiomatization, on par with $\vdash_L$ in Axiom~2, exactly as ``knows'' or ``is accessible from'' are taken as primitives in axiomatic epistemic logic and given semantics by an accompanying Kripke-style model rather than reduced to set-theoretic primitives within the axiom itself \citep{vanBenthem2011}. We give these primitives a general model-theoretic semantics below.

\paragraph{Formal Semantics for Axiom 4.}
\label{sec:contestability-semantics}

\paragraph{Definition (Contestability Model).}
A \emph{contestability model} at state $s$ is a tuple
$\calM_s=(W,\mathrm{Acc},\mathrm{Der},\mathrm{Ctrl},V)$, where
$W\subseteq\calA\times L$ is a set of \emph{decision worlds}
$w=(a_w,E_w)$ with $E_w\vdash_L(s\to a_w)$;
$\mathrm{Acc},\mathrm{Der},\mathrm{Ctrl}\subseteq W\times W$ are
accessibility relations for the decision-maker $D$---retrieving the
justification, querying its derivation, and retaining authority to move
the system, respectively; and $V:W\to\{0,1\}$ flags whether the
retrieved rendering of $E_w$ falls within $D$'s comprehension class. Then
$\mathrm{Inspectable}(a,E)$ holds at $w$ iff some $\mathrm{Acc}$-reachable
$w'$ has $V(w')=1$; $\mathrm{Challengeable}(a,E)$ holds at $w$ iff some
$\mathrm{Der}$-reachable $w''$ renders every derivation step; and
$\mathrm{Overridable}(a)$ holds at $w$ iff some $\mathrm{Ctrl}$-reachable
$w'$ has $a_{w'}\neq a$. A system's recommendation at $s$ satisfies
Axiom~4 iff \emph{every} realizable
$\calM_s$ validates all three clauses at $w=(R(s),E)$.

\paragraph{Proposition 1 (Auditability as Vacuous Comprehension).}
\label{prop:auditability-vacuous}
If $V\equiv0$ on every world $\mathrm{Acc}$-reachable from $w$, then
$\mathrm{Inspectable}(a_w,E_w)$ fails at $w$ for every contestability
model, and $\mathrm{Challengeable}$ and $\mathrm{Overridable}$
cannot discharge Axiom~4 either.

\paragraph{Proof.} Fix $w=(a_w,E_w)$ and suppose $V(w')=0$ for every $w'$ with $w\,\mathrm{Acc}\,w'$. By the truth clause above, $\mathrm{Inspectable}(a_w,E_w)$ requires some $\mathrm{Acc}$-reachable $w'$ with $V(w')=1$; none exists, so it is false at $w$. $\mathrm{Challengeable}(a_w,E_w)$ requires a $\mathrm{Der}$-reachable world that renders every derivation step comprehensibly; a step-by-step rendering is itself subject to the same valuation $V$, so if no $\mathrm{Acc}$-reachable world has $V=1$, no comprehensible rendering can be produced along $\mathrm{Der}$ either, and $\mathrm{Challengeable}$ fails. Finally, $\mathrm{Overridable}(a_w)$ requires a $\mathrm{Ctrl}$-reachable $w'=(a',E')$ with $a'\neq a_w$ \emph{together with a documented justification $E'$}, per the axiom's own text; under $V\equiv 0$ on all $\mathrm{Acc}$-reachable worlds this documentation requirement is equally undischargeable, so $\mathrm{Overridable}$ fails as well. All three conjuncts fail simultaneously. $\hfill\square$

This proposition is used below to prove the Necessity of Auditability (Theorem~3) and Black-Box Exclusion (Corollary~4.1) as special cases of a single semantic argument, rather than as separate ad hoc claims. A system whose recommendations cannot be challenged or refused is an automation tool, not a prescriptive one.

\subsubsection{General Theoretical Results}

\paragraph{Theorem 1 (Outcome Independence).}
For any prescriptive system $S$, the recommendation $R(s_t)$ is
invariant to future outcome realizations: for any $\tilde R:\calS\times\mathcal{O}_{t+n}\to\calA$ with $\tilde R(s_t,o)=R(s_t)$ for every $s_t$ and every admissible $o$,
\[
\tilde R(s_t \mid o_1) = \tilde R(s_t \mid o_2)\qquad\forall\, o_1,o_2\in\mathcal{O}_{t+n}.
\]

\paragraph{Proof.} By Axiom~1, $R$ has domain $\calS$ exactly, with $\mathcal{O}_{t+n}\notin\mathrm{dom}(R)$. Fix $s_t$ and any two admissible outcomes $o_1,o_2$. By hypothesis $\tilde R(s_t,o_1)=R(s_t)=\tilde R(s_t,o_2)$, since both equal the same fixed element $R(s_t)\in\calA$ depending only on $s_t$. As $s_t,o_1,o_2$ were arbitrary, $\tilde R(s_t,\cdot)$ is constant on $\mathcal{O}_{t+n}$ for every $s_t$.
\hfill $\square$

\paragraph{Remark (Domain of Applicability).}
Outcome Independence is non-trivial only in stochastic or partially
observable domains, where future outcome realizations are not
deterministically entailed by the current epistemic state $s_t$. In
fully deterministic and fully observable settings, outcome
realizations collapse into state transitions and the distinction
between epistemic evaluation at time $t$ and outcome observation at
time $t+n$ becomes vacuous. Prescriptive AI is therefore primarily
concerned with decision-making under epistemic uncertainty.

\paragraph{Remark (Why an outcome-based RL policy does not satisfy this at the level that matters).}
A trained RL policy $\pi(s_t)=\argmax_a\hat Q(s_t,a)$ is, at \emph{execution} time, also a function of $s_t$ alone, and so trivially satisfies Theorem~1's conclusion in exactly the same syntactic sense. The substantive difference this theorem is used to draw is therefore not about the runtime signature of $R$ versus $\pi$, but about what is required to \emph{construct and validate} $R$ versus $\hat Q$: $\hat Q$ is fit by minimizing a loss that is itself a function of realized $\mathcal{O}_{t+n}$ (a TD target or Monte Carlo return), so $\mathcal{O}_{t+n}$ enters $\hat Q$'s \emph{construction} even though it does not appear in $\hat Q$'s \emph{input signature} at decision time. It is Axiom~3 and Theorem~5 below that rule out $\mathcal{O}_{t+n}$-dependence at the level of the \emph{evaluation criterion} used to build $R$---which is the level at which the comparison with RL is actually being made throughout this paper.

\paragraph{Remark (Connection to Epistemic Logic).}
The axiomatic structure of Prescriptive AI aligns with van Benthem's
view of rational agency as epistemic state transition
\cite{vanBenthem2011}. In the language of Dynamic Epistemic Logic (DEL), a prescriptive
recommendation $R(s_t) \rightarrow a$ can be interpreted as an
epistemic action that transforms the agent's information state $K_t$,
without requiring access to future outcome realizations $O_{t+n}$.
While we do not invoke the full DEL machinery in this work, Axioms~1--3
operationalize its core principle: decision quality is a property of
the epistemic transition at time $t$, rather than of stochastic
realizations at time $t+n$. This property distinguishes prescriptive systems from outcome-based
reinforcement learning, which explicitly optimizes expected returns
over realized outcomes.

\paragraph{Theorem 2 (Imitation Incompleteness).}
\label{thm:imitation_incompleteness}
\label{sec:imitation-incompleteness}
Let $\calA$ be a finite action space, $\Theta\subset\mathbb{R}^d$ compact, and let $\theta\mapsto \Mpred(a\mid s;\theta)$ be a parametric predictive model satisfying the following regularity conditions:
\begin{itemize}
\item[\textbf{(R1)}] $\calA$ is finite;
\item[\textbf{(R2)}] $\Theta\subset\mathbb{R}^d$ is compact;
\item[\textbf{(R3)}] for every $(s,a)$, $\theta\mapsto \Mpred(a\mid s;\theta)$ is continuous on $\Theta$ and bounded away from $0$, i.e., $\inf_{\theta\in\Theta}\Mpred(a\mid s;\theta)\ge c>0$;
\item[\textbf{(R4)}] $\theta^\star\in\Theta$ uniquely maximizes the population log-likelihood $Q(\theta):=\mathbb{E}_{s\sim p}\mathbb{E}_{a\sim\pih(\cdot\mid s)}[\log \Mpred(a\mid s;\theta)]$.
\end{itemize}
Let $\pih(a\mid s)$ denote a human decision policy and let $\calH=\{(s_i,a_i)\}_{i=1}^n$ be an i.i.d.\ sample from $\pih$. Assume access to a normative utility function $U:\calS\times\calA\to\mathbb{R}$ defining optimal actions $a^\star(s)=\argmax_{a}U(s,a)$, and suppose:
\begin{enumerate}
\item \textbf{(Realizability)} there exists $\theta^\star$ such that $\Mpred(\cdot\mid s;\theta^\star)=\pih(\cdot\mid s)$ for all $s$ (this makes (R4) equivalent to requiring $\theta^\star$ be the \emph{unique} parameter realizing $\pih$, since $Q(\theta)=Q(\theta^\star)-\mathbb{E}_{s\sim p}[D_{\mathrm{KL}}(\pih(\cdot\mid s)\,\|\,\Mpred(\cdot\mid s;\theta))]$ and the KL term vanishes iff $\Mpred(\cdot\mid s;\theta)=\pih(\cdot\mid s)$ a.e.);
\item \textbf{(Systematic bias)} there exists $s_b\in\calS$ such that $a_b:=\argmax_a\pih(a\mid s_b)\neq a^\star(s_b)$;
\item \textbf{(Separation)} the maximizer is unique with margin $\delta>0$: $\pih(a_b\mid s_b)\ge\max_{a\neq a_b}\pih(a\mid s_b)+\delta$.
\end{enumerate}
Then the maximum-likelihood estimator $\hat\theta_n=\argmax_\theta\frac1n\sum_i\log\Mpred(a_i\mid s_i;\theta)$ satisfies
\[
\lim_{n\to\infty} \Mpred(s_b)=a_b \neq a^\star(s_b)\qquad\text{almost surely},
\]
and therefore no imitation-based predictive system trained solely on $\calH$ can, in general, correct systematic bias present in $\pih$ without access to $U$ or counterfactual information.

\paragraph{Proof.} \emph{Step 1 (uniform convergence).} By (R1)--(R3), for each fixed $s$ the finitely many maps $\{\theta\mapsto\log\Mpred(a\mid s;\theta)\}_{a\in\calA}$ are continuous and bounded on the compact $\Theta$. By the i.i.d.\ strong law of large numbers together with equicontinuity on compact $\Theta$, a standard Wald-type chaining argument upgrades pointwise a.s.\ convergence of $Q_n(\theta):=\frac1n\sum_i\log\Mpred(a_i\mid s_i;\theta)$ to \emph{uniform} convergence: $\sup_{\theta\in\Theta}|Q_n(\theta)-Q(\theta)|\to0$ a.s.

\emph{Step 2 (argmax consistency).} By (R4), $\theta^\star$ uniquely maximizes $Q$ on compact $\Theta$, so $\sup_{\theta:\|\theta-\theta^\star\|\ge\varepsilon}Q(\theta)<Q(\theta^\star)$ for every $\varepsilon>0$. Combined with Step~1, this is precisely the hypothesis of argmax (Wald) consistency, giving $\hat\theta_n\to\theta^\star$ a.s.

\emph{Step 3 (transfer to the induced policy).} By (R1) and (R3), $\theta\mapsto\Mpred(\cdot\mid s;\theta)$ is continuous into the finite-dimensional simplex $\Delta(\calA)$; continuity plus Step~2 gives $\Mpred(\cdot\mid s;\hat\theta_n)\to\Mpred(\cdot\mid s;\theta^\star)=\pih(\cdot\mid s)$ a.s.\ for every $s$ (realizability), and since $\calA$ is finite this is automatically total-variation convergence.

\emph{Step 4 (argmax stability at $s_b$).} By the separation margin $\delta$ and Step~3, almost surely there is a random $N$ such that for $n\ge N$, $|\Mpred(a\mid s_b;\hat\theta_n)-\pih(a\mid s_b)|<\delta/2$ for all $a$ simultaneously; a direct triangle-inequality computation then shows $\Mpred(a_b\mid s_b;\hat\theta_n)>\Mpred(a\mid s_b;\hat\theta_n)$ for every $a\neq a_b$, hence $\argmax_a\Mpred(a\mid s_b;\hat\theta_n)=a_b$ for all sufficiently large $n$, a.s.

\emph{Step 5 (no recovery of $a^\star(s_b)$).} Since $a_b\neq a^\star(s_b)$ by hypothesis and $\calH$ consists only of samples from $\pih$ with no evaluation of $U$ anywhere in its construction, no measurable function of $\calH$---$\hat\theta_n$ in particular---can depend on $U$; the limit $a_b$ is therefore independent of $U$ and equals the biased action regardless of $n$.
\hfill $\square$

\paragraph{Remark (Scope: misspecification).} Under model misspecification (no $\theta^\star$ with $\Mpred(\cdot\mid s;\theta^\star)=\pih(\cdot\mid s)$), Step~3 fails and $\hat\theta_n$ instead converges to the information-projection $\theta^\dagger:=\argmax_\theta Q(\theta)$ of $\pih$ onto the model class; the exact limiting action at $s_b$ is then a distinct, model-dependent claim, though the qualitative conclusion---no recovery of $a^\star(s_b)$ from $\calH$ alone, since $\calH$ still carries no information about $U$---persists. This sensitivity to the assumed behavioral model is not unique to this setting: \citet{SkalseAbate2024} show, in the inverse-reinforcement-learning setting, that small misspecifications of the assumed decision model can produce large errors in the recovered reward, independently motivating not relying on a single behavioral-model specification when the goal is normative evaluation rather than imitation.

\paragraph{Corollary 2.1 (Bounded Normative Improvement).}
\label{cor:bounded-improvement}
Let $p(s)$ denote the state distribution induced by $\pih$ and define
\[
V(\pi)=\mathbb{E}_{s\sim p}\,\mathbb{E}_{a\sim\pi(\cdot\mid s)}[U(s,a)].
\]
Let $\ebias:=U(s_b,a^\star(s_b))-U(s_b,a_b)>0$. Then, as $n\to\infty$, $V(\Mpred)\to V(\pih)$, while the prescriptive system
$\mathcal S(s)=a^\star(s)$ satisfies
\[
V(\mathcal S)\ge V(\pih) + p(s_b)\,\ebias.
\]
The accuracy ceiling is structural: $\mathrm{Acc}(\Mpred)\le 1-\Pr_{s\sim p}[s\in\calS_{\mathrm{bias}}]$ with $\calS_{\mathrm{bias}}=\{s:\argmax_a\pih(a\mid s)\neq a^\star(s)\}$, independent of $n$, model capacity, or learning algorithm---a possible formal basis for both the empirically observed $\sim$70\% accuracy plateau of substitution-prediction models \citep{mohandas} and the empirical acceptability gap between imitating a single prescriber and an externally grounded rule base observed for MYCIN (Section~\ref{sec:mycin_news2}; \citealp{Yu1979}).

\paragraph{Proof.} From Theorem~2, $\Mpred(\cdot\mid s)\to\pih(\cdot\mid s)$ for all $s$, hence $V(\Mpred)\to V(\pih)$ by dominated convergence on finite $\calA$. Writing $V(\mathcal S)=V(\pih)+\mathbb{E}_{s\sim p}[U(s,a^\star(s))-\mathbb{E}_{a\sim\pih(\cdot\mid s)}[U(s,a)]]$, the bracketed term is non-negative for every $s$ by optimality of $a^\star$, and equals at least $\ebias$ at $s_b$ (where $\pih$ places its mode, and asymptotically increasing mass, on $a_b$); weighting by $p(s_b)$ gives the stated bound. \hfill $\square$

\paragraph{Corollary 2.2 (Imitation Incompleteness under Markovian Dependence).}
\label{cor:markov}
Suppose $\calH=(s_i,a_i)_{i\ge1}$ is instead a single stationary, irreducible, aperiodic Markov trajectory on a finite state space with $a_i\sim\pih(\cdot\mid s_i)$ and stationary marginal $p$, rather than i.i.d.\ samples. Then, under (R1)--(R3) and the separation hypothesis, $\hat\theta_n\to\theta^\star$ almost surely and Theorem~2's conclusion holds verbatim.

\paragraph{Proof sketch.} A finite-state, irreducible, aperiodic chain started in stationarity is ergodic; the successive excursions between visits to a fixed recurrent reference state are i.i.d.\ blocks by the strong Markov property. Applying the classical renewal-reward theorem to these blocks yields the same strong law $\frac1n\sum_i\log\Mpred(a_i\mid s_i;\theta)\to Q(\theta)$ a.s.\ for each fixed $\theta$ that i.i.d.\ sampling supplied in Theorem~2's Step~1; because $\calS,\calA$ are finite, the empirical state-action frequencies converge a.s.\ to $p(s)\pih(a\mid s)$, giving the same uniform convergence of $Q_n$ to $Q$. Steps~2--5 of Theorem~2's proof then apply verbatim, since they depend only on this uniform convergence and on the fact that $\calH$ carries no evaluation of $U$, both of which hold regardless of the sampling scheme. The extension does \emph{not} reach non-stationary or non-recurrent settings, since the regeneration argument requires the chain to revisit a fixed reference state infinitely often almost surely. \hfill $\square$

\paragraph{Corollary 2.3 (Finite-Sample Concentration).}
\label{cor:finite-sample}
Retain (R1)--(R4) and further assume (R5) $\theta\mapsto\log\Mpred(a\mid s;\theta)$ is $L$-Lipschitz for every $(s,a)$, and (R6) a quadratic identifiability margin $Q(\theta^\star)-Q(\theta)\ge\kappa\|\theta-\theta^\star\|^2$ near $\theta^\star$. Then there exist constants $C=C(L,B,D,\kappa)$ (with $B$ a bound on $|\log \Mpred|$ and $D=\mathrm{diam}(\Theta)$) such that, for any $\eta\in(0,1)$,
\[
n\ \ge\ C\cdot\delta^{-4}\left[d\log(1/\delta)+\log(1/\eta)\right]
\]
suffices to guarantee $\Pr\big(\argmax_a\Mpred(a\mid s_b;\hat\theta_n)\neq a_b\big)\le\eta$.

\paragraph{Proof sketch.} A pointwise Hoeffding bound on $Q_n(\theta)-Q(\theta)$, combined with an $\epsilon$-net covering argument over the compact $\Theta$ (using the $L$-Lipschitz property (R5) to interpolate between net points), yields a uniform concentration bound $\Pr(\sup_\theta|Q_n(\theta)-Q(\theta)|\ge t)\le 2(6LD/t)^d\exp(-nt^2/2B^2)$. The basic inequality $Q(\theta^\star)-Q(\hat\theta_n)\le 2\sup_\theta|Q_n(\theta)-Q(\theta)|$ (from optimality of $\hat\theta_n$ for $Q_n$) combined with the quadratic margin (R6) gives $\|\hat\theta_n-\theta^\star\|\le\sqrt{(2/\kappa)\sup_\theta|Q_n(\theta)-Q(\theta)|}$; Lipschitz transfer (as in Theorem~2, Step~3--4) converts this into a bound on $|\Mpred(a\mid s_b;\hat\theta_n)-\pih(a\mid s_b)|$, and the event of incorrect argmax at $s_b$ is contained in $\{\|\hat\theta_n-\theta^\star\|\ge\delta/(2L)\}$; substituting into the uniform concentration bound and solving for $n$ gives the stated sample-size requirement. \hfill $\square$

\paragraph{Theorem 3 (Necessity of Auditability).}
\label{thm:auditability}
Any non-auditable system cannot be prescriptive.

\paragraph{Proof.}
If a system is non-auditable, it fails $\mathrm{Inspectable}(a,E)$ for any $a=R(s)$, and by Proposition~1 (taking $V\equiv0$ on all $\mathrm{Acc}$-reachable worlds) $\mathrm{Challengeable}$ and $\mathrm{Overridable}$ are then also undischargeable, so Axiom~4 is violated, contradicting the definition of prescriptiveness. \hfill $\square$

\paragraph{Theorem 4 (Strict but Non-Disjoint Separation of Paradigms).}
\label{thm:separation}
Let $\textsc{PRED}$ denote predictive systems and $\textsc{PRESC}$
prescriptive systems, as defined in Section~3.2.3. Then:
\[
\textsc{PRED} \cap \textsc{PRESC} \neq \emptyset,\quad
\textsc{PRED} \nsubseteq \textsc{PRESC},\quad
\textsc{PRESC} \nsubseteq \textsc{PRED}.
\]

\paragraph{Proof.}
(i) \emph{Non-empty intersection.} Consider a Bayesian decision network that
(a) models $P(o \mid s, a)$ via probabilistic inference over an explicit, semantically labeled causal graph $G$ fit to historical $(s,a,o)$ triples---making it a member of $\textsc{PRED}$ with $Y=\mathcal{O}_{t+n}$;
(b) selects $a^* = \argmax_a \mathbb{E}[U(a) \mid s]$ via an explicit
decision rule with $E = \text{``}G \vdash_L (s \rightarrow a^*)\text{''}$ as justification---making it a member of $\textsc{PRESC}$, since it satisfies Axioms~1--4 (Axiom~4 verified via the Definition~1 contestability model, with $\mathrm{Acc}$ realized as access to $G$ and its posterior, $\mathrm{Der}$ as the ability to query parent nodes and conditional probabilities, $\mathrm{Ctrl}$ as the decision-maker's authority to override $a^*$, and $V(w)=1$ since $G$'s nodes are semantically labeled by construction). This exhibits $\Sigma\in\textsc{PRED}\cap\textsc{PRESC}$.

(ii) \emph{$\textsc{PRED} \nsubseteq \textsc{PRESC}$.} Let $\Sigma$ be a black-box predictive system (e.g., a deep neural network) fit by behavioral cloning, $\Sigma_{\mathrm{fn}}(s):=\argmax_a\hat P(a\mid s)$---this places $\Sigma\in\textsc{PRED}$ by construction, independently of $\Sigma_{\mathrm{fn}}$'s internal structure. Since $\Sigma_{\mathrm{fn}}$ exposes no expressible $E\vdash_L(s\to a)$ (its parameters admit no semantic decomposition into $L$-formulae), Axiom~2 fails; by Proposition~1 with $V\equiv0$ (no rendering of the network's internal computation is ever within a human decision-maker's comprehension class), Axiom~4 fails as well, so $\Sigma\notin\textsc{PRESC}$. This is an explicit witness in $\textsc{PRED}\setminus\textsc{PRESC}$, and the same argument applies to every $\Sigma\in\textsc{PRED}$ lacking a semantically decomposable justification---this is precisely Corollary~4.1 below, proved by this argument rather than merely stated alongside it.

(iii) \emph{$\textsc{PRESC} \nsubseteq \textsc{PRED}$.} Let $\Sigma$ be MYCIN's antimicrobial-therapy rule base (Section~\ref{sec:mycin_news2}), whose $\sim$200 production rules and certainty factors are authored ex ante from infectious-disease domain knowledge, with zero parameters fit to any historical sample of actions or outcomes \citep{Shortliffe1975}. $\Sigma_{\mathrm{fn}}$ satisfies Axioms~1--4 (verified in Section~\ref{sec:mycin_news2}), and since its provenance involves no estimator $\hat P(Y\mid s_t)$ for either $Y=A$ or $Y=\mathcal{O}_{t+n}$, $\Sigma\notin\textsc{PRED}$. Hence $\Sigma\in\textsc{PRESC}\setminus\textsc{PRED}$, an explicit witness. (This is a claim about the specific witness exhibited, not about rule-based systems in general: a rule base whose thresholds were tuned against historical data would instead fall in case~(i).)
\hfill $\square$

\paragraph{Proposition 2 (Computational Realizability).}
\label{prop:realizability}
Axioms~1--4 are jointly satisfiable by multiple, structurally distinct computational architectures.

\paragraph{Proof (by construction).}
We exhibit three structurally distinct implementations, each verified via a corresponding instantiation of the Definition~1 contestability model:
\textbf{(i) Fuzzy logic systems} (e.g., the soccer substitution auditor, Section~\ref{sec:fuzzy_system}; MYCIN, Section~\ref{sec:mycin_news2}), where activated or fired rules and their membership/certainty degrees constitute the justification $E$;
\textbf{(ii) Bayesian decision systems}, where expected utility is computed explicitly with a semantically labeled causal graph as justification (as in Theorem~4(i)); and
\textbf{(iii) constraint-based systems}, where recommendations arise
from an explicit constraint-satisfaction proof serving as $E$.
In each case Axiom~1 holds because the recommendation's domain is $\calS$ by construction, Axiom~3 holds because the decision criterion is computed from $(s_t,\calN)$ without reference to $\mathcal{O}_{t+n}$, and Axioms~2 and~4 hold by the cited instantiation. Prescriptiveness therefore characterizes a paradigm class rather than a unique architecture.
\hfill $\square$

\paragraph{Corollary 4.1 (Black-Box Exclusion).}
\label{cor:blackbox-exclusion}
Black-box decision systems are not prescriptive: any $\Sigma$ exposing no expressible $E\vdash_L(s\to a)$ violates Axiom~2 directly, and, by Proposition~1 applied with $V\equiv0$, violates Axiom~4 as well; by Definition~1 it is therefore not prescriptive. This is a claim about the specific axiomatic notion of prescriptiveness defined here, not a general dismissal of black-box models' practical utility: a black-box component may still be predictively useful, or may sit inside a hybrid pipeline whose outer decision layer independently satisfies the axioms.

\textbf{Theorem 5 (Impossibility of Outcome-Based Decision Labeling in Stochastic Environments).}
\label{thm:outcome_labeling}

Let $\mathcal{S}$ be a prescriptive decision system operating in a \textbf{stochastic environment}—a setting where outcomes are substantially influenced by factors unknown or uncontrollable at decision time.

At decision time $T$, the system observes an epistemic state $E_T$ consisting of all information available at time $T$. Let $O_{T+N}(a)$ denote the stochastic outcome realized at future time $T+N$ after action $a$ is taken. Define the \emph{episode space} $\calE:=\calS\times\calA\times\mathcal{O}_{t+n}$, whose elements $e=(s_t,a,o_{t+n})$ record a decision state, action, and realized outcome; a \emph{terminal} episode $\calE_T$ has $o_{t+n}$ observed, unlike $s_t$ alone (all that is available to $R$ under Axiom~1).

Assume:
\begin{enumerate}
    \item Exactly one action $a^* \in A$ is executed at time $T$.
    \item For any alternative action $a' \neq a^*$, $O_{T+N}(a')$ is counterfactual and unobservable.
    \item Multiple actions may lead to acceptable or unacceptable outcomes under different stochastic realizations.
    \item Prescriptive optimality is defined over $U(E_T, a)$, not over realized outcomes $O_{T+N}$.
\end{enumerate}

Then:
\begin{enumerate}[label=(\alph*)]
    \item There exists no binary labeling function $y: (E \times A) \to \{0,1\}$ assigning ``correct'' or ``incorrect'' decisions consistently with $U(E_T, a)$, where $E$ denotes the space of epistemic states.
    \item Consequently, outcome-based classification metrics (accuracy, precision, recall, F1-score) are ill-defined for prescriptive systems.
    \item Each prescriptive decision must instead be evaluated individually at time $T$ by its logical and normative consistency with epistemic state $E_T$, independently of realized outcome at $T+N$.
\end{enumerate}

\textit{Proof.} Suppose $y(E_T,a):=\mathbb{1}[a=\argmax_{a'}U(E_T,a')]$ is well-defined for every terminal episode $(\star)$. Consider two cases for how $y$ could be required to be measurable.

\emph{Case 1 (ex ante: $y$ is $\sigma(s_t)$-measurable).} Any labeling rule usable to evaluate $a=R(s_t)$ \emph{when it is made} must be a function of information available at $t$. But $(\star)$ makes $y$ a function of $\calE_T$, which includes $o_{t+n}$; whenever the environment is stochastic, $o_{t+n}$ is not a deterministic function of $s_t$, hence not $\sigma(s_t)$-measurable. No $\sigma(s_t)$-measurable $y$ can satisfy $(\star)$---this is precisely a violation of Axiom~3.

\emph{Case 2 (ex post: $y$ is only $\sigma(\calE_T)$-measurable).} Then $(\star)$ requires evaluating $U(\calE_T,a')$ for \emph{every} $a'\in\calA$, not only the $a$ actually taken. For $a'\neq a$, the counterfactual episode requires a draw $o'_{t+n}\sim P(\cdot\mid s_t,a')$ under an action never taken; since only one action is taken at $t$, this counterfactual draw is not a measurable function of any observable variable in the decision log. No function of available data can therefore evaluate the right-hand side of $(\star)$ for $a'\neq a$.

Both cases exhaust the possible measurability requirements on $y$, and in neither does a well-defined $y$ satisfying $(\star)$ exist. $\hfill\Box$

\paragraph{Remark (Scope of the Impossibility Result).}
This impossibility applies specifically to assigning binary correctness labels to \emph{individual} decision instances based on realized outcomes. It does \emph{not} prohibit: (1) predicting continuous variables and validating via regression metrics, since these require only the single realized $o_{t+n}$ under the action actually taken, never a counterfactual draw; (2) using aggregate outcome distributions for \emph{population-level} calibration without labeling individual decisions, since an aggregate statistic over many decision instances at which different actions were taken is a function of realized, non-counterfactual triples; or (3) fully deterministic environments, where outcomes are fully determined by observable state and choice, so there is no counterfactual to fail to observe.
\clearpage

\begin{figure}[h]
\centering
\begin{tikzpicture}[
  level 1/.style={sibling distance=10cm, level distance=2cm},
  level 2/.style={sibling distance=6cm, level distance=2.5cm},
  level 3/.style={sibling distance=5cm, level distance=2.5cm},
  every node/.style={draw, rectangle, rounded corners, align=center, 
                     minimum width=3cm, minimum height=1cm},
  decision/.style={fill=blue!10},
  prohibited/.style={fill=red!20, text=red!80!black, thick},
  permitted/.style={fill=green!20, text=green!50!black}
]

\node[decision] {ML System for\\Decision Support}
  child {node[decision] {Deterministic\\Environment?}
    child {node[permitted] {\textbf{Exception 3}\\Outcome = Ground Truth\\(e.g., Chess, Formal Proofs)\\$\checkmark$ \textsc{Permitted}}}
    child {node[decision] {Stochastic\\Environment}
      child {node[decision] {What is\\Validated?}
        child {node[permitted] {\textbf{Exception 1}\\Continuous Variable\\Estimation\\(MAE, R²)\\$\checkmark$ \textsc{Permitted}}}
        child {node[permitted] {\textbf{Exception 2}\\Population-Level\\Calibration\\$\checkmark$ \textsc{Permitted}}}
        child {node[prohibited] {\textbf{Theorem 5 Violation}\\Individual Decision\\Outcome Labeling\\$\times$ \textsc{Prohibited}}}
      }
    }
  };

\end{tikzpicture}
\caption{Decision tree for classifying ML-based decision systems under Theorem 5. 
The framework partitions all possible validation approaches into four mutually exclusive 
categories, identifying one prohibited class and three permitted exceptions.}
\label{fig:theorem5_taxonomy}
\end{figure}

\textbf{Corollary 5.1 (Invalidity of Predictive Baselines).}
\label{cor:invalid-baselines}

Let $M_{\text{pred}}$ be a predictive model evaluated via outcome-based classification metrics. Then comparing a prescriptive system $\mathcal{S}$ against $M_{\text{pred}}$ using these metrics constitutes a category error.

\textit{Proof.} From Theorem 5, outcome-based classification metrics are undefined for $\mathcal{S}$, as no labeling function $y: (E \times A) \to \{0,1\}$ exists. Therefore, $M_{\text{pred}}$ and $\mathcal{S}$ cannot be compared using a common metric space: any reported comparison either (a) silently substitutes a different, non-equivalent scoring rule, or (b) restricts attention to Exceptions~1--2 above, in which case it is valid but is not an individual-decision accuracy comparison. $\Box$

\paragraph{Remark (Practical Implications).} Theorem 5 does not prohibit empirical validation of prescriptive systems; it specifies that validation must be epistemic rather than outcome-based. Valid methods include: epistemic coherence assessment (does $a^*$ align with $U(E_T,\cdot)$?); decision-latency quantification; counterfactual case analysis; longitudinal pattern detection; and aggregate acceptability rating across a case series, as illustrated by the MYCIN blinded panel (Section~\ref{sec:mycin_news2}).

\subsection{Normative Non-Identifiability}
\label{sec:non-identifiability}

The Imitation Incompleteness theorem shows that a predictive system fit to $\pih$ converges to the biased action $a_b$ rather than the optimal $a^\star(s_b)$. A natural follow-up question is whether the normative utility $U^\star$ itself could instead be \emph{recovered} from behavioral data—e.g., via inverse reinforcement learning—so that the bias could be corrected after the fact. We show this is not possible in general, and that the resulting non-identifiability is qualitatively different from the classical non-identifiability of IRL.

\paragraph{Definition (Modal Rationalizability).}
$\pih:\calS\to\Delta(\calA)$ is \emph{modally rationalizable} by $U\in\calU=\{U:\calS\times\calA\to\mathbb{R}\}$ at $s$ if $\argmax_a U(s,a)=\argmax_a\pih(a\mid s)$: $U$ ranks the policy's own modal action highest. This is a weak notion, constraining only the arg max and not the full ordering—minimal in the sense that observing only the mode of $\pih(\cdot\mid s)$ gives no grounds to prefer the true $U^\star$ over any $\tilde U$ ranking the same action highest.

\paragraph{Theorem 6 (Normative Non-Identifiability).}
\label{thm:non-identifiability}
Suppose $\pih$ exhibits systematic bias at $s_b$ under $U^\star$: $a_b:=\argmax_a\pih(a\mid s_b)\neq a^\star(s_b):=\argmax_a U^\star(s_b,a)$. Let $\calU_{\mathrm{rat}}(s_b):=\{U\in\calU:\argmax_a U(s_b,a)=a_b\}$. Then:
\begin{enumerate}[label=(\alph*)]
\item $U^\star\notin\calU_{\mathrm{rat}}(s_b)$---the true utility does not rationalize the biased behavior, by definition of bias.
\item $|\calU_{\mathrm{rat}}(s_b)|=\infty$, and every $\tilde U,\tilde U'\in\calU_{\mathrm{rat}}(s_b)$ induce the identical likelihood of $\calH$, since $\calH$ is generated by $\pih$ alone and does not depend on $U$; no statistical test on $\calH$ can distinguish elements of $\calU_{\mathrm{rat}}(s_b)$.
\end{enumerate}
Any procedure inferring a utility from $\calH$ under modal rationality recovers at best an unidentifiable element of $\calU_{\mathrm{rat}}(s_b)$, and by (a), provably \emph{not} $U^\star$. External normative specification is thus \emph{epistemically necessary}.

\paragraph{Proof.} (a) By the bias hypothesis, $\argmax_a U^\star(s_b,a)=a^\star(s_b)\neq a_b$, so $U^\star$ fails the defining condition of $\calU_{\mathrm{rat}}(s_b)$. (b) For any $c>0$, define $\tilde U_c(s,a):=U^\star(s,a)$ for $s\neq s_b$, $\tilde U_c(s_b,a_b):=U^\star(s_b,a_b)+c$, and $\tilde U_c(s_b,a):=U^\star(s_b,a)-c$ for $a\neq a_b$. For every $c>0$, $\argmax_a\tilde U_c(s_b,a)=a_b$ by construction, so $\tilde U_c\in\calU_{\mathrm{rat}}(s_b)$; since $c\mapsto\tilde U_c$ is injective on $(0,\infty)$, this exhibits an (uncountable) injection into $\calU_{\mathrm{rat}}(s_b)$. The likelihood $L(\calH)=\prod_{i:s_i=s_b}\pih(a_i\mid s_b)$ is a function of $\pih$ alone and never contains $U$ as an argument, so it takes an identical value regardless of which $U\in\calU_{\mathrm{rat}}(s_b)$ is entertained; no likelihood-based test on $\calH$ can distinguish them. \hfill $\square$

\paragraph{Remark (Not merely classical IRL non-identifiability).} Part~(b) alone would be the classical observation that a policy is consistent with infinitely many reward functions \citep{NgRussell2000}—by itself symmetric, privileging no member of $\calU_{\mathrm{rat}}(s_b)$ over another. Part~(a) is the additional, non-classical content specific to the biased setting: the true normative standard $U^\star$ is not merely one indistinguishable member among others of the rationalizable class—it is \emph{excluded} from that class entirely.

\paragraph{Corollary 6.1 (IRL Bias Amplification).}
\label{cor:irl-amplification}
Let MaxEnt IRL \citep{ZiebartEtAl2008} fit a softmax policy $\pi_{\hat R_n}(a\mid s)\propto\exp(\hat R_n(s,a;\theta))$ by maximum likelihood on $\calH\sim\pih$, satisfying (R1)--(R4) of Theorem~2 under this parameterization. Then $\hat R_n\to\hat R_\infty$ with $\argmax_a\pi_{\hat R_\infty}(a\mid s_b)=a_b\neq a^\star(s_b)$ almost surely: MaxEnt IRL converges to a \emph{specific} reward ranking the biased action highest, not to an arbitrary member of $\calU_{\mathrm{rat}}(s_b)$.

\paragraph{Proof.} $\Mpred(a\mid s;\theta):=\pi_{\hat R_n}(a\mid s;\theta)$ is exactly an instance of the M-estimator of Theorem~2 under this parameterization, so Theorem~2's proof (Steps~1--4) applies verbatim: $\hat\theta_n\to\theta^\star$ a.s., hence $\pi_{\hat R_n}(\cdot\mid s_b;\hat\theta_n)\to\pih(\cdot\mid s_b)$ a.s., and $\argmax_a\pi_{\hat R_n}(a\mid s_b;\hat\theta_n)=a_b$ for all sufficiently large $n$, a.s. \hfill $\square$

\paragraph{Proposition 3 (Directional Concentration under Estimation).}
\label{prop:directional}
Let $\calR(\pih)=\{R:\pi_R=\pih\}$ be the classical non-identified reward class of \citet{NgRussell2000}—non-singleton whenever $\pih$ is optimal for some ground-truth reward $R_0$, since $\pi_{R_0+c}=\pi_{R_0}$ for any constant $c$ and the degenerate reward $R\equiv0$ induces every policy as weakly optimal; this ambiguity is \emph{symmetric}, with no element of $\{R_0+c:c\in\mathbb{R}\}\cup\{R\equiv0\}$ privileged over another by $\pih$ alone.

Under systematic bias at $s_b$, the MaxEnt IRL estimator $\hat R_n$ of Corollary~6.1 is \emph{not} symmetric: it converges in probability to a specific $\hat R_\infty$ with $\argmax_a\pi_{\hat R_\infty}(a\mid s_b)=a_b$ almost surely. So while \emph{identification} (Theorem~6) is symmetric—many rewards are consistent with $\pih$—\emph{estimation} by a concrete, standard algorithm is not: any consistent MaxEnt IRL estimator concentrates, with probability approaching one, on the biased action $a_b$, never on $a^\star(s_b)$ or an arbitrary element of $\calR(\pih)$. This is the precise sense in which IRL \emph{amplifies}, rather than merely fails to resolve, systematic bias.

\paragraph{Proof.} The symmetry of $\calR(\pih)$ follows from the two families of policy-invariant reward transformations noted above. Directionality of $\hat R_n$ follows from Corollary~6.1's proof, which shows the specific limit $\hat R_\infty$ satisfies $\argmax_a\pi_{\hat R_\infty}(a\mid s_b)=a_b$ deterministically (in probability), rather than landing on an arbitrary or unspecified element of $\calR(\pih)$. \hfill $\square$

\paragraph{Remark (Relation to Bayesian decision theory).} \citet{Savage1954} defines rational choice as $a^\star(s)=\argmax_a\mathbb{E}_{o\sim P(\cdot\mid s,a)}[U(s,a,o)]$. \prescAI{} separates the \emph{evaluating agent} (the system, applying $U$ correctly) from the \emph{acting agent} (the human, who may not); Theorem~2 shows convergence to $\pih$ is convergence to a biased epistemic transition function, and prescriptive auditing substitutes $U^\star$-grounded transitions for $\pih$-grounded ones. This correspondence is stated precisely in Proposition~4 below.

\paragraph{Theorem 7 (Interpretability Necessity).}
\label{thm:interpretability_necessity}
Let $\mathcal{S}$ satisfy Axiom 2 (Epistemic Justification) and Axiom 4 (Contestability). Then $\mathcal{S}$ must provide human-inspectable explanations: for every state $s$, every realizable contestability model contains a world $w'$ with $w\,\mathrm{Acc}\,w'$ and $V(w')=1$.

\paragraph{Proof.} By Axiom~4, $\mathrm{Inspectable}(\Sigma_{\mathrm{fn}}(s),E)$ holds at $w=(\Sigma_{\mathrm{fn}}(s),E)$; by the Definition~1 truth clause this means precisely that such a $w'$ exists. \hfill $\square$

\paragraph{Corollary 7.1 (Insufficiency of Post-Hoc Explainability).}
\label{cor:posthoc}
Post-hoc explainability techniques (e.g., SHAP, LIME) applied to black-box models do not, in general, satisfy Axiom~2 or Theorem~7, insofar as they approximate model behavior---optimizing an approximation loss over a sampled neighborhood or coalition, with an explicit non-zero residual by construction---rather than exhibiting the exact derivability relation $E\vdash_L(s\to a)$ that Axiom~2 requires at every individual $s$. Were exact agreement to hold everywhere, the surrogate would not be a post-hoc \emph{approximation} at all, contradicting how SHAP and LIME are actually constructed. This is a claim about the specific axiomatic notion of justification defined here, not a general denial of SHAP/LIME's diagnostic value.

\paragraph{Proposition 4 (Prescriptive--Bayes Correspondence).}
\label{prop:bayes-corr}
A prescriptive system $\mathcal{S}$ satisfying Axioms~1--4 and computing $R(s)=\argmax_a\mathbb{E}_P[U(s,a)]$ implements the ex ante expected-utility decision rule that Savage's representation theorem \citep{Savage1954} attributes to a rational agent whose preferences over acts satisfy Savage's postulates, \emph{given} $U$ and $P$ as already-specified inputs. This claim is purely computational: it does not assert that $\mathcal{S}$'s $(U,P)$ have themselves been derived from, or independently verified to represent, a preference relation satisfying Savage's postulates---establishing that would require an independent elicitation argument outside the present scope.

\paragraph{Proof.} By Axiom~1, $R$ conditions only on $s_t$, matching Savage's ex ante evaluation timing (current probability $P$ over states, not realized outcomes). By Axiom~3, quality is a function of $(s_t,\calN)$ alone, matching Savage's requirement that the decision rule depend only on current information and the utility/probability assignment. The computation $\argmax_a\mathbb{E}_P[U(s,a)]$ is syntactically identical to Savage's formula given $(U,P)$; by Axiom~2, the justification $E=$``$\mathbb{E}_P[U(s,\cdot)]$'' is expressible and accessible at decision time, so the computation is exhibited, not merely performed. \hfill $\square$

\subsubsection{Boundary Cases Within Prescriptive Analytics}
Several classes of systems fall within Prescriptive Analytics but do not, in their generic form, satisfy the defining criteria of Prescriptive AI:

\textbf{Optimization-based systems} compute actions that maximize objective functions under constraints \cite{Bertsimas2020}. While effective in structured settings, they typically assume stable objectives and do not audit human decision justification under uncertainty.

\textbf{Reinforcement Learning systems} prescribe actions by optimizing expected reward \cite{SuttonBarto2018}. Despite their prescriptive nature, they are often opaque and automation-oriented, limiting contestability and interpretability, and are particularly vulnerable to outcome bias in stochastic environments; when their reward is fit to historical logs, they are additionally subject to Theorem~2 and Theorem~6.

\textbf{Recommendation and next-best-action systems} prioritize actions based on engagement or conversion metrics, relying on outcome-driven evaluation and rarely addressing accountability or asymmetric risk.

\textbf{Rule-based and expert systems} occupy a distinct position within this boundary class: unlike the categories above, their compliance with Prescriptive AI is not foreclosed by architecture but is instead \emph{instance-dependent}. Because such systems are authored directly by domain specialists rather than fit to behavioral logs, whether a given rule base satisfies Axioms~1--4 must be verified case by case, checking specifically for (i) an external, non-imitative normative source for the rules (Component~C3), (ii) an inspectable derivation trace from state to recommendation (Axiom~2), and (iii) a documented mechanism by which a human agent may query and override a recommendation (Axiom~4). A generic rule engine lacking these mechanisms remains a boundary case, prescriptive only in an analytic sense; MYCIN and NEWS2 (Section~5) are, by contrast, rule-based systems for which this instance-specific verification succeeds, and which we therefore classify within Prescriptive AI proper rather than as boundary cases.

These systems are prescriptive in an analytic sense, but do not, in general, fulfill the normative role required of Prescriptive AI without such verification.

\subsection{Hybrid Systems and Overlapping Set Membership}

The conceptual sets described here are not mutually exclusive. Hybridization is the norm in modern AI systems, which frequently combine symbolic reasoning, optimization, statistical learning, and deep learning.

As with the relationship between machine learning and deep learning, subset membership reflects conceptual scope rather than exclusivity. A system may employ optimization or learning techniques associated with Prescriptive Analytics while simultaneously satisfying the normative criteria that qualify it as Prescriptive AI. Classification therefore depends not on the techniques employed, but on the system's functional role in the decision-making process.

\subsubsection{Terminological Convergence and Alternative Taxonomies}
\label{sec:terminology}

Current developments in machine intelligence exhibit a structural pivot from purely predictive architectures toward systems designed to constrain, guide, or audit agentic behavior. This convergence reflects the growing recognition that stochastic outcome prediction is frequently insufficient for evaluating performance in high-stakes, dynamic environments where process validity and contextual coherence supersede result optimization. We introduce the term \textit{Prescriptive AI} to describe this emergent class of systems. Unlike predictive models that estimate the probability of a future state, Prescriptive AI establishes a normative framework for evaluating the coherence of a decision at the precise moment of its execution.

This perspective aligns with foundational work on the logical dynamics of information and agency, particularly van Benthem's account of decision-making as an epistemic state transition under uncertainty \citep{vanBenthem2007,vanBenthem2011}. While the present work does not implement the formal machinery of dynamic epistemic logic, it adopts this conceptual stance by evaluating decisions based on the informational state and normative criteria available at the moment of action, rather than on the stochastic realization of downstream outcomes. By treating decision-making as a dynamic process of information uptake and evaluation, Prescriptive AI analytically decouples the agent's reasoning from stochastic environmental variability, enabling principled auditing of decision quality independently of realized results.

This definition allows us to position Prescriptive AI precisely relative to adjacent taxonomies. While we adopt this term to emphasize the normative function of the system, the framework intersects with, and provides empirical grounding for, several established conceptual traditions:

\begin{itemize}
    \item \textbf{Decision Intelligence (DI):} Commonly used in industrial and organizational contexts, Decision Intelligence emphasizes the engineering of decision workflows. The proposed framework constitutes a specific instantiation of DI in which the critical decision interface is mediated by intrinsically interpretable symbolic reasoning, addressing auditability requirements that are often under-specified in conventional DI implementations.

    \item \textbf{Normative AI:} Within the ethics and governance literature, systems that encode explicit standards of acceptable behavior are often described as normative. The system employed in this work functions as a normative agent insofar as it enforces a codified standard of decision quality—derived from both hard performance constraints and soft contextual heuristics—against which human actions are evaluated in real time. This standing is not merely conceptual: Theorem~6 gives a formal reason why such a standard cannot be discovered from behavioral data and must instead be authored, echoing the specification difficulties long documented for fixed normative constraints such as Asimov's Three Laws \citep{Asimov1942,AndersonAnderson2007} and the preference heterogeneity documented by the Moral Machine experiment \citep{AwadEtAl2018}.

    \item \textbf{Cognitive Orthotics:} In human–computer interaction, cognitive orthotics refer to systems designed to compensate for specific human cognitive limitations, such as attentional lapses or systematic biases. By explicitly correcting for effects such as status quo bias and outcome bias, the proposed framework operates as a real-time cognitive stabilization mechanism in high-stakes decision contexts.

    \item \textbf{Real-Time Algorithmic Auditing:} Algorithmic auditing is traditionally framed as a post-hoc activity concerned with fairness or compliance. In contrast, the present approach extends auditing into the decision loop itself, enabling continuous, run-time evaluation of human decision states prior to the execution of irreversible actions.
\end{itemize}

Within this landscape, \textit{Prescriptive AI} serves as a unifying descriptor that emphasizes the system's functional output—the prescription—while retaining the requirement for intrinsic interpretability and auditability that is central across these literatures.

\subsubsection{Implications for High-Stakes Decision Making}

By explicitly positioning Prescriptive AI as a normative subset of Prescriptive Analytics, and Prescriptive Analytics as a structured subset of a broader heterogeneous space of action-oriented systems, this work resolves a persistent source of conceptual confusion in both academic and industrial discourse \cite{Sun2020, Wissuchek2022}.

This hierarchy clarifies that:
\begin{itemize}
 \item Not all action-oriented systems are prescriptive analytics;
 \item Not all prescriptive analytics systems are Prescriptive AI;
 \item Hybrid systems may legitimately belong to multiple conceptual sets.
\end{itemize}

Prescriptive AI should therefore be understood not as a generic label for action recommendation, but as a distinct AI paradigm designed to audit and support human judgment under uncertainty. In the remainder of this work, elite soccer is employed not merely as a case study, but as a demanding stress-test environment to operationalize and validate these principles under time pressure, incomplete observability, and asymmetric risk—complemented, in Section~\ref{sec:mycin_news2}, by two independent, previously-existing clinical systems that let us check the same axioms against evidence the present authors did not generate.

\subsection{Ethical Use of AI in High-Stakes Decision Systems}

Auditing artificial intelligence systems in high-stakes decision-making contexts is not merely a technical challenge but a fundamentally ethical one. As argued by \citet{Rudin2019}, deploying opaque, black-box models in domains where decisions have irreversible consequences is inherently problematic. When such systems are extended to environments with intense public scrutiny and continuous human judgment—such as elite professional soccer or acute clinical care—the ethical risks are amplified rather than mitigated.

In contrast to low-stakes or background automation, in-game tactical decisions are subject to immediate visual auditing by multiple human observers, including coaches, players, analysts, referees, and millions of spectators. Any divergence between an algorithmic recommendation and human intuition is instantly questioned. In this setting, the adoption of black-box models poses a substantial risk of rejection, driven by two central ethical concerns: epistemic opacity and institutional distrust.

\paragraph{Epistemic opacity and public accountability.}
The vast majority of stakeholders involved in professional sports—including coaches, athletes, and fans—do not possess formal training in artificial intelligence, let alone in complex models such as deep neural networks. When a black-box system produces a recommendation that contradicts human judgment, the absence of a transparent rationale forces stakeholders to accept the decision on faith alone. This requirement for blind trust is ethically untenable in high-stakes environments, where decisions can alter careers, financial outcomes, and competitive integrity.

In contrast, auditable and intrinsically interpretable models—such as fuzzy logic systems—enable post-hoc and real-time explanation of decisions at multiple levels of abstraction. In the proposed framework, the entire reasoning pipeline remains open to inspection: from the microscopic computation of performance scores, through fuzzy membership functions, to the macroscopic aggregation of linguistic rules that generate the final substitution priority. This transparency substantially reduces epistemic asymmetry between the system and its users, fostering informed trust rather than blind acceptance. Importantly, such auditability allows non-technical stakeholders to understand why a recommendation was made, even without understanding *how* the underlying mathematics operates.

\paragraph{Institutional distrust and risk of misuse.}
High-stakes domains with significant financial flows are particularly vulnerable to suspicion, manipulation, and perceived conflicts of interest. Professional sports exemplify this risk. The rapid expansion of online betting markets has reshaped the economic landscape of elite soccer, with betting companies now acting as major sponsors of clubs and leagues. Concurrently, multiple high-profile cases of match manipulation and betting-related misconduct involving professional players have eroded public trust.

In such an environment, the deployment of opaque AI systems introduces severe ethical hazards. Consider a scenario in which a black-box model recommends substituting a star player during a decisive match, followed by a negative outcome. Without an explicit and intelligible justification, the decision becomes vulnerable to accusations of corruption, collusion with betting interests, or deliberate sabotage. Coaches would lack defensible grounds for their actions, players could be unfairly penalized without explanation, and clubs would face reputational and legal risks. These dynamics create strong institutional incentives to avoid black-box systems altogether, regardless of their predictive accuracy.

By contrast, prescriptive and interpretable decision support systems mitigate these risks by design. Rather than replacing human judgment, they aim to *augment* it, offering structured, explainable insights that the human decision-maker may accept, reject, or contextualize. This human-in-the-loop paradigm significantly reduces the probability of catastrophic errors while preserving accountability. Moreover, because every recommendation can be traced to explicit rules and contextual indicators, the system provides a defensible audit trail that protects coaches, players, and institutions alike.

\paragraph{Prescriptive AI as an ethical design principle.}
The ethical challenge in high-stakes AI is not merely accuracy but responsibility. Predictive models trained to imitate historical decisions inevitably inherit human biases, including status quo bias, outcome bias, and risk aversion. In contrast, prescriptive systems grounded in symbolic reasoning evaluate current conditions against normative rules, enabling systematic auditing of human decisions rather than their replication.

This work adopts interpretability and prescriptiveness as core ethical design principles. By prioritizing transparency, auditability, and human agency, the proposed system aligns with emerging arguments that intrinsically interpretable models are not only preferable but necessary in domains where errors carry disproportionate consequences \citep{Rudin2019}. Rather than automating authority, the system enhances human decision-making, offering a practical pathway for deploying AI responsibly in high-stakes environments.

Ultimately, while the adoption of AI in decision support is unavoidable, its ethical deployment is not optional. In contexts such as elite soccer and acute clinical care—where decisions are public, consequential, and irreversible—only transparent, auditable, and human-centered systems can achieve legitimate and sustainable integration.

\section{The General Normative Prescriptive AI Framework (GNPAF)}
\label{sec:gnpaf}

\subsection{Logical Positioning}

This section introduces the \textbf{General Normative Prescriptive AI Framework (GNPAF)}, which specifies the minimal \emph{structural requirements} that any prescriptive AI system must satisfy in order to be normatively evaluable, epistemically grounded at decision time, and independent of realized outcomes.

Within this framework, we define the \textbf{General Normative Decision Auditing (GNDA)} procedure, which operationalizes how concrete decisions—human or artificial—can be audited for normative coherence under uncertainty.

The distinction is intentional:
\begin{itemize}
    \item \textbf{GNPAF} specifies \emph{what must exist structurally}.
    \item \textbf{GNDA} specifies \emph{how auditing is performed} once those structures are in place.
\end{itemize}

This separation follows directly from the axiomatic distinction between epistemic state, normative evaluation, and realized outcomes, as well as from the impossibility results of Section~3 (Theorems~2, 5, and~6) showing that prescriptive policies cannot, in general, be learned or validated through outcome supervision alone.

\paragraph{Relationship to Existing Taxonomies.}
GNPAF differs fundamentally from related approaches:

\begin{itemize}
    \item \textbf{Decision Theory (Savage, 1954):} provides axiomatic foundations for rational choice (formalized in Proposition~4), but does not specify computational structures for real-time prescriptive auditing under bounded rationality.
    
    \item \textbf{Prescriptive Analytics:} focuses on optimizing actions with respect to objectives, but does not impose explicit epistemic state formalization or outcome decoupling.
    
    \item \textbf{Explainable AI (XAI):} adds interpretability post-hoc to predictive models, whereas GNPAF requires explainability \emph{structurally} as a prerequisite for prescriptiveness (Corollary~7.1).

    \item \textbf{Prescriptive Machine Learning \citep{Hullermeier2021}:} identifies methodological challenges for \emph{learning} prescriptive models; GNPAF is a structural specification of what \emph{any} system, learned or not, must expose (Section~\ref{sec:hullermeier} maps GNPAF's components onto each of its five challenges explicitly).
\end{itemize}

GNPAF thus occupies a distinct position: more operational than pure decision theory, more normatively constrained than optimization-driven prescriptive analytics, and more structurally committed than post-hoc explainability frameworks.

\subsection{GNPAF Structural Components}

GNPAF decomposes prescriptive AI systems into six components.
These components are \emph{not algorithmic steps}, but \emph{logical requirements}.
Violating any component necessarily violates at least one axiom.

\paragraph{The Six GNPAF Components.}
\begin{enumerate}
    \item \textbf{Epistemic State Specification (C1)} — What is known at decision time?
    \item \textbf{Action Space Delimitation (C2)} — What actions are admissible?
    \item \textbf{Normative Evaluation Mapping (C3)} — How should actions be evaluated \emph{ex ante}?
    \item \textbf{Decision Coherence Assessment (C4, GNDA)} — Was the chosen action normatively justified?
    \item \textbf{Explainable Justification Layer (C5, GNDA)} — Why or why not (human-contestable)?
    \item \textbf{Outcome Decoupling Validation (C6, GNDA)} — Is evaluation independent of realized outcomes?
\end{enumerate}

\subsubsection{Epistemic State Specification (C1)}
\label{sec:epistemic}

(Axioms 1, 3; van Benthem)

A prescriptive system must explicitly define the \emph{epistemic state} at the moment a decision is taken. An epistemic state represents the totality of information legitimately available at time $t$, including observable facts, known constraints, recognized uncertainties, and institutional or procedural limitations. It explicitly excludes future observations, realized outcomes, and information revealed after the decision.

Following van Benthem, decisions are modeled as transitions originating from epistemic states rather than as reactions to realized trajectories. Epistemic states are therefore conditions of knowledge, not feature vectors or retrospective summaries. Without explicit epistemic state specification, ex-ante normative evaluation is ill-defined.

\subsubsection{Action Space Delimitation (C2)}
\label{sec:action}

(Axiom 1)

Given an epistemic state $s_t$, the prescriptive system must characterize the set of admissible actions $\mathcal{A}(s_t)$: $a \in \mathcal{A}(s_t) \iff a$ is feasible, permissible, and executable given $s_t$. Action spaces are inherently state-dependent: $\mathcal{A}(s_t) \neq \mathcal{A}(s_{t'})$ for $t \neq t'$. Normative evaluation is meaningful only relative to actions that were genuinely available at decision time; evaluating a decision against infeasible or unavailable alternatives violates ex-ante rationality.

\subsubsection{Normative Evaluation Mapping (C3)}
\label{sec:normative}

(Axiom 2; Savage)

GNPAF requires an explicit normative evaluation mapping $f_{\text{presc}} : (s_t, N) \mapsto \mathbb{R}^{|\mathcal{A}(s_t)|}$, where $N$ denotes externally specified normative criteria. The mapping must be conditional on the epistemic state, defined ex ante, and independent of realized outcomes. By Theorems~2 and~6, this mapping cannot be learned purely from outcome supervision, cannot be inferred from imitation of historical decisions, and cannot be validated retrospectively by performance; by Theorem~6 specifically, even a utility function underlying the mapping cannot in general be recovered from behavioral data once systematic bias is present.

\textbf{Note:} By Theorem~5, these constraints on individual decision labeling apply in stochastic environments. In deterministic settings, outcome-based validation may be appropriate.

\emph{The normative mapping is not optimized, but specified.} This is a defining property of prescriptive systems under GNPAF; Section~\ref{sec:ustar-discussion} discusses who specifies it and why this division of labor is defensible rather than evasive.

\subsubsection{Decision Coherence Assessment (C4)}
\label{sec:coherence}

GNDA operationalizes prescriptive auditing by comparing the action actually taken by a decision-maker with the action(s) normatively justified in the corresponding epistemic state. Writing $a_h$ for the human decision and $R(s)$ for the recommendation, C4 partitions decisions, for a domain-specified tolerance $\epsilon\ge0$, into:
\begin{itemize}
\item \emph{Coherent} ($a_h=R(s)$);
\item \emph{Delayed} ($a_h=R(s_{t'})$ for an earlier $t'<t$);
\item \emph{Partially justified} ($a_h\neq R(s)$ but $\mathrm{Quality}(a_h)\ge\mathrm{Quality}(R(s))-\epsilon$); and
\item \emph{Normatively inconsistent} (otherwise).
\end{itemize}
Because outcome-based evaluation is provably illegitimate in stochastic settings (Theorem~5), this four-way partition constitutes the only admissible form of prescriptive auditing at the level of an individual decision; auditing is therefore \emph{structural, not empirical}. In the empirical instantiations of this paper, $\epsilon$ is set operationally: via the blinded-panel acceptability rating for MYCIN (Section~\ref{sec:mycin_news2}), a fixed score-band tolerance for NEWS2, and the rule-triggered latency window for the soccer auditor (Section~\ref{sec:fuzzy_system}).

\subsubsection{Explainable Justification Layer (C5)}
\label{sec:explain}

(Axiom 4)

GNDA formalizes that prescriptive systems, by definition, must provide explanations that are contestable by humans and grounded in epistemic factors, normative criteria, and explicit trade-offs. Explainability in GNPAF is not post-hoc or auxiliary; it is a structural requirement for normative accountability, formalized by Theorem~7 and its Corollary~7.1 on the insufficiency of post-hoc explainability.

\begin{center}
\begin{tabular}{lcc}
\toprule
 & Post-Hoc XAI & GNPAF \\
\midrule
Timing & After model output & During decision process \\
Source & Approximate & Direct \\
Status & Optional & Axiomatic (A4) \\
Human role & Observer & Contestor \\
\bottomrule
\end{tabular}
\end{center}

Explainability is required for accountability, not added for transparency.

\subsubsection{Outcome Decoupling Validation (C6)}
\label{sec:outcome}

GNPAF requires explicit verification that none of the preceding 
components depends on realized outcomes or future information unknown 
at decision time.

\textbf{Applicability:} This component is essential in stochastic 
environments where Axiom 1 applies. In deterministic 
settings where outcomes uniquely determine decision quality, outcome-based 
validation remains valid and this component may be omitted.

\textbf{Validation procedure:}
\begin{enumerate}
    \item Identify all inputs to Components 1--5
    \item Verify that no input depends on $O_{t+n}$ (outcomes at time $t+n$)
    \item Confirm that any predictive models used estimate variables 
    from state $s_t$, not from future realizations
\end{enumerate}

This validation step prevents collapse into retrospective outcome-based 
evaluation and enforces the ex-ante epistemic nature of prescriptive 
rationality.
\clearpage

\subsection{Graphic diagram}

\begin{figure}[h]
\centering
\includegraphics[width=1\linewidth]{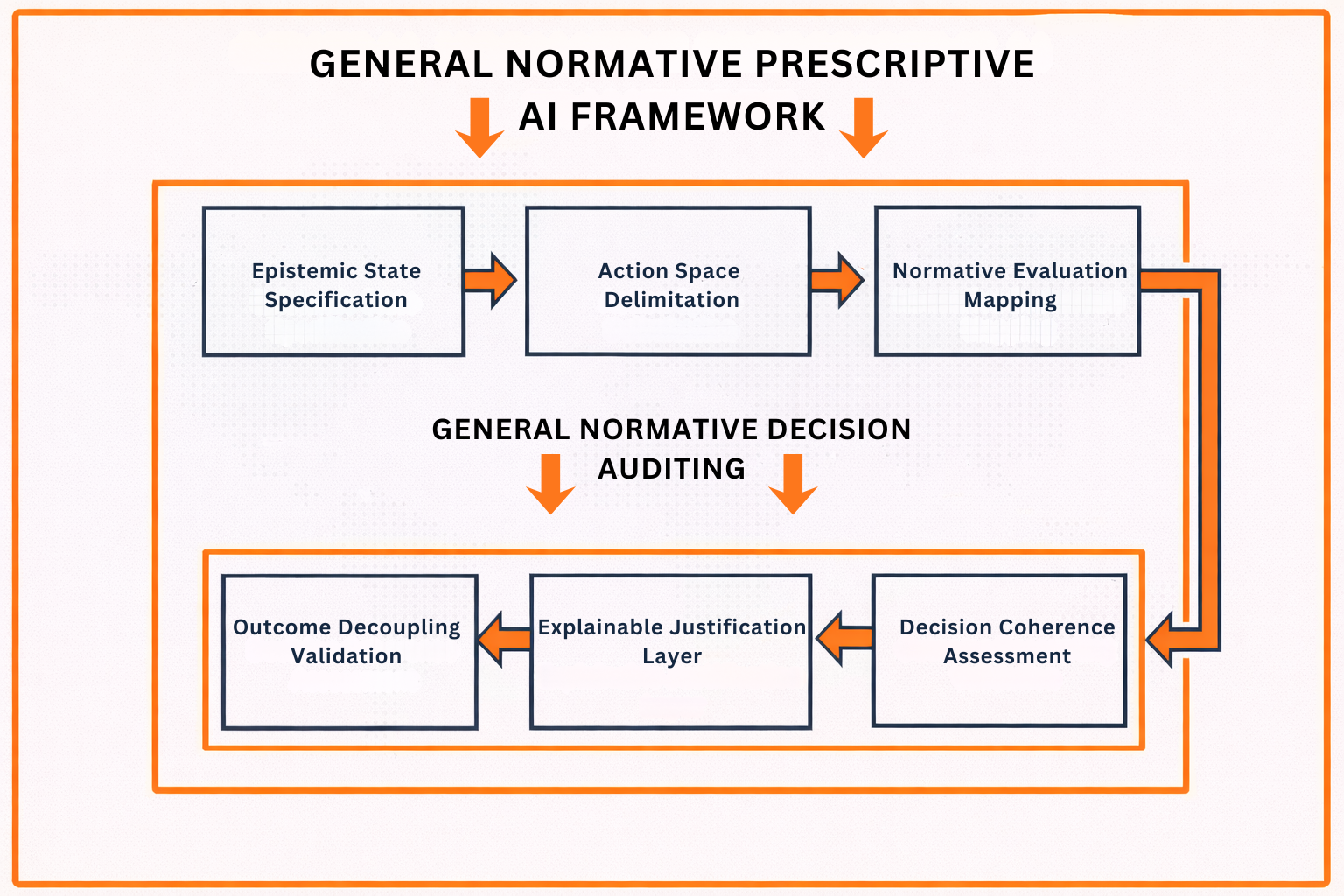}
\caption{Prescriptive AI framework illustration}
\end{figure}

\subsection{GNPAF as a Universal Characterization of Prescriptive Systems}

The formalization presented in Sections~3--4 establishes GNPAF as a \textit{universal characterization} of prescriptive decision-making under uncertainty. This subsection demonstrates that GNPAF functions as a domain-agnostic, methodology-agnostic criterion for evaluating whether any computational decision support system operates prescriptively.

\subsubsection{Formal Characterization}

\textbf{Definition (Prescriptive System, restated).} A computational decision system $\mathcal{S}$ is \textit{prescriptive} if and only if it satisfies Axioms 1--4 and implements Components C1--C6 of GNPAF.

This definition is:
\begin{itemize}
    \item \textbf{Technology-agnostic}: Valid for symbolic, statistical, hybrid, or neural architectures.
    \item \textbf{Domain-agnostic}: Applicable across domains requiring decision-making under uncertainty.
    \item \textbf{Verifiable}: Compliance can be audited through inspection of system architecture and data flow.
\end{itemize}

\subsubsection{Methodological Inclusiveness}

GNPAF does not exclude established methodologies but provides a principled basis for determining when they operate prescriptively. Prescriptiveness is a \textit{structural property of decision logic}, not a consequence of algorithmic sophistication. Table~\ref{tab:gnpaf_compliance} demonstrates compliance across representative methodologies.

\begin{table}[h]
\centering
\caption{GNPAF Compliance Across Decision Methodologies}
\label{tab:gnpaf_compliance}
\resizebox{\linewidth}{!}{%
\begin{tabular}{@{}lcccccc@{}}
\toprule
\textbf{Method} & \textbf{A1} & \textbf{A2} & \textbf{A3} & \textbf{A4} & \textbf{C3} & \textbf{Compliant?} \\
\midrule
Fuzzy Inference Systems     & \checkmark & \checkmark & \checkmark & \checkmark & \checkmark & Yes \\
Bayesian Networks           & \checkmark & \checkmark & \checkmark & \checkmark & \checkmark & Yes \\
Decision Trees (shallow)    & \checkmark & \checkmark & \checkmark & \checkmark & \checkmark & Yes \\
Explainable Boosting        & \checkmark & \checkmark & \checkmark & \checkmark & \checkmark & Yes \\
Linear Regression$^*$       & \checkmark & \checkmark & \checkmark & \checkmark & \checkmark & Yes \\
Rule-Based Systems (MYCIN, NEWS2)$^{\dagger}$ & \checkmark & \checkmark & \checkmark & \checkmark & \checkmark & Yes \\
Rule-Based Systems (generic, no WHY/HOW, no override channel) & \checkmark & $\times$ & \checkmark & $\times$ & \checkmark & No \\
\midrule
Random Forests              & \checkmark & $\triangle$ & \checkmark & $\triangle$ & \checkmark & Conditional \\
Gradient Boosting (opaque)  & \checkmark & $\times$ & \checkmark & $\times$ & \checkmark & No \\
Deep Neural Networks        & $\triangle$ & $\times$ & $\triangle$ & $\times$ & $\times$ & No \\
Black-Box RL                & $\times$ & $\times$ & $\times$ & $\times$ & $\times$ & No \\
\bottomrule
\end{tabular}%
}
\vspace{0.2cm}
\raggedright
\footnotesize
$^*$When trained on normative targets, not outcome supervision. \\
$^{\dagger}$Verified per instance (Section~5): requires an external, non-imitative rule source (C3), an inspectable derivation trace (Axiom~2), and a documented override channel (Axiom~4); a generic rule engine lacking these remains a boundary case (\S3.2.1). \\
\checkmark = Satisfied; $\times$ = Violated; $\triangle$ = Implementation-dependent.
\end{table}

\subsubsection{Formal Example: Linear Regression}

We demonstrate that GNPAF compliance is achievable with elementary statistical methods.

\textbf{Scenario.} A clinical system recommends medication dosage based on patient state (weight, age, kidney function), trained on expert-specified optimal dosages.

\textbf{Compliance Verification:}
\begin{itemize}
    \item \textbf{Axiom 1:} Model maps $s_t \to \hat{d}$; outcomes at $t+n$ do not enter.
    \item \textbf{Axiom 2:} Linear equation $\hat{d} = \beta_0 + \sum_i \beta_i x_i$ is fully transparent.
    \item \textbf{Axiom 3:} Quality assessed by alignment with risk criteria, not survival outcomes.
    \item \textbf{Axiom 4:} Clinician can inspect coefficients and override with justification.
    \item \textbf{Component 3:} Trained on normative targets (expert dosages), not outcome supervision.
    \item \textbf{Component 6:} Auditable: all inputs observable at time $t$.
\end{itemize}

\textbf{Result.} Linear regression satisfies GNPAF when architecturally aligned with normative principles, demonstrating that prescriptiveness depends on structure, not complexity.

\subsubsection{Boundary Conditions}

\paragraph{Deterministic Environments.} In fully deterministic settings where outcomes are uniquely determined by state and action, outcome-based evaluation may be appropriate (Theorem~5 Remark). Components~3 and~6 may be relaxed.

\paragraph{Absence of Normative Consensus.} GNPAF requires externally specifiable normative criteria $N$. In domains lacking normative consensus (purely subjective judgments), prescriptive auditing may be ill-defined; the Moral Machine finding of substantial, culturally-varying disagreement over moral preferences \citep{AwadEtAl2018} is direct empirical evidence of how large this problem can be for domains such as autonomous-vehicle ethics or content moderation, discussed further in Section~\ref{sec:ustar-discussion}.

\paragraph{Stochastic Policies.} Systems with stochastic action selection satisfy GNPAF if the \textit{policy} $\pi(a|s)$ is justified and contestable, even when individual samples vary. The requirement is transparency of policy reasoning, not deterministic action selection.

\subsubsection{Resolution}

GNPAF establishes prescriptiveness as a verifiable structural property of decision systems. The inclusiveness demonstrated—even classical linear regression can comply—reflects that prescriptiveness concerns decision architecture and epistemic rigor, not algorithmic sophistication. The empirical instantiations in Sections~\ref{sec:fuzzy_system}--\ref{sec:mycin_news2} demonstrate computational realizability across three independent domains and five decades; the primary contribution, however, remains the formal characterization itself.

\subsection{Cross-Domain Applicability: From Pitch to Critical Operations}
\label{sec:cross_domain}

High-stakes, real-time decision-making environments provide naturalistic laboratories for observing adversarial and time-sensitive judgments. The structural constraints of such domains—namely, irreversibility of actions, severe time pressure, and asymmetric risk—are common across many safety-critical sectors. The prescriptive AI framework (GNPAF) proposed here in \cite{Passos2025_Wharton} does not depend on domain-specific mechanics, but rather on the temporal auditing of decision-relevant performance signals and agent state evolution. By incorporating logically interpretable reasoning and principles from Explainable AI, the architecture offers a blueprint for extending transparency and auditability to a broader class of action-oriented systems. Consequently, it may be applicable to environments in which human judgment is vulnerable to fatigue, status quo bias, or desensitization under sustained cognitive load.

\begin{itemize}
    \item \textbf{Clinical Triage and High-Dependency Care:} In emergency and critical-care settings, clinicians operate under persistent cognitive load, where subtle signs of deterioration may be overlooked due to desensitization or alarm fatigue. A prescriptive auditor could operate over continuous indicators of physiological state evolution—analogous to the performance signals used in this study—to surface latent risk trajectories that remain within nominal ranges but exhibit adverse momentum. Rather than producing diagnostic predictions, the system would prescribe prioritized re-evaluation, supporting timely intervention before irreversible deterioration occurs. This is not merely conjectural: Section~\ref{sec:mycin_news2} shows that MYCIN and NEWS2, two independently engineered clinical systems built decades apart, already instantiate exactly this structure and have been validated at population scale.
    
    \item \textbf{Financial Risk Management (Trading Desks):} In financial decision-making, human operators are known to exhibit systematic biases such as sunk cost effects and loss aversion, leading to prolonged exposure to unfavorable positions. A prescriptive auditing system could continuously evaluate the alignment between risk exposure and market dynamics, issuing normative recommendations when exposure exceeds logic-based thresholds. In this sense, the system functions not as a predictor of market outcomes, but as an auditor of decision states under volatility.
    
    \item \textbf{Football Player and Tactical Management:} In professional football, decisions regarding substitutions and tactical adjustments are often influenced by human biases, including status quo bias and outcome bias, which can delay or distort interventions. A prescriptive AI framework could be applied to evaluate decision-relevant states in real time, potentially \textbf{decoupling recommended actions from match outcomes} and mitigating reliance on salient events that may mislead human judgment. By auditing latent risk states and providing interpretable, contestable recommendations, such a system might support interventions—such as substitutions or role adjustments—based on objective assessments of performance deterioration rather than heuristic or retrospective outcome evaluation. The ``substitution'' logic instantiated in this context may generalize to proactive, bias-resistant management of player capacity and team performance, and could potentially inform decision-making in other high-stakes, real-time human-in-the-loop domains.
\end{itemize}

Across these domains, the core function of the system remains invariant: to decouple the evaluation of the decision state at time $t$ from stochastic outcomes at time $t+n$, thereby mitigating outcome bias in high-variance, high-stakes environments. \textbf{To operationalize and rigorously stress-test this prescriptive paradigm, we deliberately select elite professional football as our primary empirical domain, and corroborate it with two independent historical instantiations from clinical medicine.} This choice is methodological rather than domain-driven: elite football constitutes an adversarial natural laboratory in which decisions are time-critical, irreversible, and subject to continuous public scrutiny, while outcome signals are sparse, delayed, and dominated by stochastic variance. These conditions systematically amplify the very cognitive and epistemic failure modes---such as outcome bias, salience-driven miscalibration, and status quo inertia---that prescriptive auditing is designed to expose. Moreover, the high temporal resolution of in-game decision windows enables fine-grained longitudinal analysis of decision states under evolving uncertainty, a property rarely attainable in controlled laboratory settings. MYCIN and NEWS2 (Section~\ref{sec:mycin_news2}) then serve a complementary methodological function: neither was built with GNPAF in mind, so their independent conformity to the same six components is evidence that the axioms are not simply reverse-engineered from the soccer case study. \textbf{Together, demonstrating epistemic coherence, auditability, and contestability under the extreme variance and institutional pressures of elite football, and independently verifying the same structure in two previously validated clinical systems, provide a stringent, cross-domain stress test for the proposed framework.}

\section{Empirical Instantiations: MYCIN and NEWS2}
\label{sec:mycin_news2}

\subsection{Motivation and Setup}

Sections~\ref{sec:methodology}--\ref{sec:results} instantiate GNPAF computationally, in a system purpose-built for this paper, in elite soccer. Before turning to that primary case study, we first ask a more conservative question: does GNPAF's axiomatic structure already show up, independently of our design choices, in previously existing, previously validated decision-support systems built for entirely different purposes? We answer this by instantiating GNPAF in two independent domains, four decades apart: \textbf{MYCIN}, the rule-based consultation system for antimicrobial therapy developed at Stanford \citep{Shortliffe1975} and subsequently evaluated in a blinded comparison against practicing specialists \citep{Yu1979}; and \textbf{NEWS2}, the physiological early-warning protocol developed by the Royal College of Physicians \citep{RCP2017} and mandated across acute NHS trusts in England since 2019, subsequently validated on a large prospective cohort \citep{MartinRodriguez2019}.

Neither system was purpose-built to satisfy GNPAF: MYCIN predates the framework by five decades, and NEWS2 was designed for population-level triage, not axiomatic compliance. We treat this as a stress test in the same spirit as the soccer case study, but along a different axis: if two independently engineered systems, built for different purposes in different eras and neither aware of fuzzy substitution auditing, already expose the same six structural components, this is evidence the components are not an artifact of how we defined them around the soccer system, but a genuine structural regularity of prescriptive decision support.

MYCIN's aim was never to match a hospital's historical prescribing pattern: its therapy rules were authored and reviewed by infectious-disease faculty independently of any single clinician's logged decisions, and its knowledge base could be inspected, queried, and corrected by a physician at consultation time \citep{Shortliffe1975}---exactly the C3 requirement that $f_{\mathrm{presc}}$ be specified rather than fit to $\calH$ (Theorem~2). This is why MYCIN illustrates \prescAI{} rather than predictive imitation: its recommendation never conditions on which antimicrobial a prescriber \emph{tends} to choose, or on how the infection later resolved.

\subsection{GNPAF Component Instantiation}
\label{sec:gnpaf-instantiation}

Table~\ref{tab:two-instantiations} maps the six GNPAF components onto both domains, alongside the soccer instantiation of Section~\ref{sec:fuzzy_system} for comparison.

\textbf{C1--C2:} MYCIN's epistemic state is the set of clinical and laboratory parameters gathered up to the point of a recommendation (organism identity or likely identity, staining and morphology, culture site, host compromise, allergy history, renal/hepatic status), explicitly excluding the future course of the infection; its action space is the set of antimicrobial regimens (single agents or combinations) applicable given the patient's allergies and organ function \citep{Shortliffe1975}. NEWS2 uses six physiological parameters plus a supplemental-oxygen flag, aggregated into a $[0,20]$ score,\footnote{NEWS2 also specifies a second scale for confirmed hypercapnic respiratory failure, requiring an explicit, documented senior-clinician decision.} with score-band action thresholds (e.g., $\ge5$ urgent review, $\ge7$ emergency assessment).

\textbf{C3:} MYCIN's therapy selection is driven by an explicit corpus of roughly 200 clinician-authored decision rules with attached certainty factors, applied deductively to the current patient state by a goal-directed inference procedure, with no term in the rule base fit to any hospital's own escalation or prescribing history \citep{Shortliffe1975}---satisfying Theorem~2's requirement for an external normative signal. NEWS2's score and band-to-response mapping are likewise fixed ex ante by RCP policy, calibrated against population-level associations between physiological derangement and adverse outcomes \citep{RCP2017}, not against any clinician's escalation history---in contrast to a predictive system trained on a hospital's own records, which by Theorem~2 would reproduce that hospital's escalation bias.

\textbf{C4:} the companion evaluation of MYCIN operationalizes C4 at population level: a blinded panel of eight infectious-disease specialists rated MYCIN's recommendation and each of nine prescribers' recommendations for the same ten meningitis cases on a shared acceptability scale \citep{Yu1979}, avoiding the category error of treating individual outcome agreement as the relevant metric (Theorem~5). For NEWS2, a clinician's action versus the band-specified response is classified under the same four-category partition of Section~\ref{sec:coherence}.

\textbf{C5--C6:} in MYCIN, every recommendation traces to the specific fired rules and their certainty factors, retrievable in English via the WHY/HOW explanation commands, so $E\vdash_L(s\to a)$ is available at consultation time rather than reconstructed after the fact \citep{Shortliffe1975}; for NEWS2, every point traces to a named parameter (e.g., ``respiratory rate 25/min $\to$ 3 points''). Neither system's recommendation logic references the patient's eventual clinical outcome, so C6 is architectural in both---exactly as the soccer fuzzy system's rule base (Section~\ref{sec:fuzzy_system}) never conditions on match result.
\clearpage
\begin{table}[h]
  \centering
  \caption{GNPAF component mapping and validation evidence across three independent instantiations. MYCIN figures from \citet{Shortliffe1975,Yu1979}; NEWS2 figures from \citet{MartinRodriguez2019}; soccer figures from Sections~\ref{sec:methodology}--\ref{sec:results}.}
  \label{tab:two-instantiations}
  \small
  \begin{tabular}{@{}p{0.13\columnwidth}p{0.28\columnwidth}p{0.28\columnwidth}p{0.28\columnwidth}@{}}
    \toprule
    & \textbf{MYCIN} & \textbf{NEWS2} & \textbf{Soccer Fuzzy Auditor} \\
    \midrule
    C1 & Culture, organism, and host parameters (ex ante) & 6 vitals + O$_2$ flag/observation & Cumulative percentile performance, fatigue, cards, momentum \\
    C2 & Antimicrobial regimens, by allergy/organ function & Monitoring tier, by score band & Substitute in / retain, per position \\
    C3 & $\sim$200 clinician-authored rules, certainty factors (ex ante) & RCP score + response chart (ex ante) & $\sim$15 fuzzy rules, expert-authored (ex ante) \\
    C4 & Blinded panel rating (4-way partition applies) & Same 4-way partition & Rank-based comparison vs.\ actual substitution timing \\
    C5 & Fired rules + certainty factors, via WHY/HOW $\to$ NL & Per-parameter point breakdown & Activated rules + membership degrees \\
    C6 & Architectural, no outcome term in rule base & Architectural, per observation & Architectural, no result term in rule base \\
    \midrule
    Scale & 8 blinded evaluators, 9 prescribers, 10 meningitis cases & Prospective cohort, $N=1{,}288$, Apr.\ 2018--Feb.\ 2019 & $N=9{,}152$ decision windows, 25 World Cup matches \\
    Evidence & Blinded comparative acceptability rating & AUC $0.891$; sens.\ $79.7\%$; spec.\ $84.5\%$; cutoff $\ge9$ & Rank alignment + decision-latency quantification \\
    Baseline & 5 faculty specialists: 42.5\%--62.5\% acceptability (cf.\ Cor.~2.1) & Learned predictors would reproduce escalation bias (Thm.~2) & Substitution-prediction ML plateaus at $\sim$70\% accuracy \citep{mohandas} \\
    Axioms probed & 1--4, C4, C6 & 1--4, C4, C6 & 1--4, C4, C6 \\
    \bottomrule
  \end{tabular}
\end{table}
\clearpage
The MYCIN instantiation contributes historical, independently reported evidence that a non-imitative, externally-specified rule base can be computationally realized and judged acceptable by domain experts at a rate at or above practicing specialists; NEWS2 contributes population-level validation \citep{MartinRodriguez2019} showing such a system can be institutionally deployed at national scale. Together with the soccer instantiation of Sections~\ref{sec:methodology}--\ref{sec:results}, they indicate GNPAF's conditions (C1--C6, Axioms~1--4) are satisfiable across disparate domains, decision cadences, and eras---sports, acute medicine, and infectious-disease consultation---though illustrative rather than exhaustive.

\subsection{A Blinded Comparative Evaluation: MYCIN versus Faculty Specialists}
\label{sec:mycin-eval}

\citet{Yu1979} evaluated MYCIN's antimicrobial recommendations against those of nine practicing prescribers (five faculty infectious-disease specialists, plus residents and students) on ten actual meningitis cases, using eight independent infectious-disease experts blinded to the source of each recommendation. Consistent with Theorem~5, this is a state-based acceptability rating at the moment of prescription, not an outcome-based figure keyed to whether the patient ultimately recovered: C4 categories are judged against what was known when the regimen was chosen, not the patient's eventual course.

\paragraph{MYCIN vs.\ faculty specialists (Bounded Normative Improvement in practice).}
MYCIN's regimens received an overall acceptability rating of \textbf{65\%} from the blinded panel, versus \textbf{42.5\% to 62.5\%} for the five faculty specialists on the same cases \citep{Yu1979}---every faculty specialist individually rated below MYCIN. Because MYCIN's rules were authored independently of these prescribers' own habits, this is an empirical illustration, not a proof, of Corollary~2.1: a system fit to imitate a given prescriber is bounded near that prescriber's own normative gap $\ebias$, whereas one evaluated against externally specified criteria is not bounded by any single demonstrator's bias. This is also a direct, real-world instance of the argument in Corollary~4.1 and Corollary~5.1: comparing MYCIN's 65\% acceptability to any single prescriber's imitation accuracy would be a category error under Theorem~5, since the acceptability figures are aggregate, ex-ante judgments (Exception~2 of Theorem~5), not individual outcome labels.

\paragraph{Coverage without over-treatment (Outcome Decoupling in practice).}
\citet{Yu1979} additionally report that MYCIN never failed to cover a treatable pathogen across the ten cases, while using no more antimicrobial agents than necessary---evidence that the rule base's state-based criteria (organism identity, site, host compromise) sufficed to avoid both under- and over-treatment without reference to any case's eventual outcome, illustrating Axiom~1 at the level of a full evaluation rather than a single decision.

\paragraph{Epistemic boundaries (Explanation as inspectability).}
Because every MYCIN recommendation could be traced, via the WHY and HOW commands, to the specific rules and certainty factors that produced it \citep{Shortliffe1975}, the blinded evaluators' disagreements with MYCIN in \citet{Yu1979} could be adjudicated against an explicit, inspectable rule trace rather than an opaque score---precisely the contestability Axiom~4 requires, and which a black-box predictor of the same behavior would not provide (Corollary~4.1).

For NEWS2, a prospective multi-center cohort \citep{MartinRodriguez2019} enrolled $N=1{,}288$ patients (Apr.\ 2018--Feb.\ 2019); 262 (20.3\%) required trauma assistance and 69 (5.4\%) died within 48 hours. Among compared triage scores, NEWS2 had the best predictive capacity for early mortality: AUC $0.891$ (95\% CI $0.84$--$0.94$), sensitivity $79.7\%$, specificity $84.5\%$ at cutoff nine, positive likelihood ratio $5.14$, negative predictive value $98.7\%$.

Together, MYCIN shows GNPAF's structure was already computationally realizable in 1975, independently validated against practicing specialists, while NEWS2 shows the same structural principles are present in a modern, population-validated protocol: initial evidence of constructibility and cross-era applicability, complementing rather than substituting for the purpose-built, fully instrumented soccer stress test of Sections~\ref{sec:methodology}--\ref{sec:results}.

\subsection{Scope of the Clinical Instantiations}

We emphasize what MYCIN and NEWS2 are, and are not, offered as evidence for. Neither evaluation is an outcome-based clinical trial: \citeauthor{Yu1979}'s blinded panel rated regimens on an ex-ante acceptability scale, not against eventual patient outcome, and by Corollary~5.1 any comparison of MYCIN against an imitation-based predictive baseline on a common outcome-based accuracy scale would itself be a category error. What is validated is \emph{retrospective structural compliance}: MYCIN and NEWS2 were not constructed to satisfy GNPAF, so their conformity to C1--C6 is, if anything, stronger evidence than a purpose-built demonstration would be---their rule base, explanation commands, and scoring protocol were motivated entirely by clinical acceptability and physician trust, and their structural correspondence to GNPAF was not a design goal of either project. We treat this, together with the soccer instantiation, as complementary, illustrative evidence of computational realizability (Proposition~2) across independently engineered systems---not as comprehensive validation of the framework in general.

\section{Methodology}
\label{sec:methodology}

\subsection{Baseline Limitations and Methodological Motivation}

The academic literature on soccer analytics provides the theoretical foundation for this work, situated at the intersection of performance evaluation and tactical decision modeling.

A central methodological challenge in this domain is quantifying player performance. The \textbf{PlayeRank} framework, proposed by \citet{Pappalardo2019}, defines a multidimensional, role-aware metric that combines individual technical actions with influence on the passing network, validated against professional scout assessments. The authors explicitly aim to reduce biases related to playing time, normalizing performance vectors by the number of actions to ``avoid biases due to play time.''

However, although normalization occurs at the level of the action vector, the final match score is computed as a weighted cumulative sum of contributions. When inspected over the temporal progression of a match, this formulation tends to produce strictly increasing trajectories, effectively reintroducing a play-time exposure bias that favors players who remain longer on the field, even if their per-minute contribution rates are similar. This monotonic behavior introduces a methodological limitation for temporal or comparative analyses. Each additional action contributes positively to the cumulative sum—regardless of qualitative impact—so players with identical per-minute performance profiles diverge in total score purely as a function of playing time. While this approach is suitable for match-level summaries, it becomes problematic when the goal is to characterize relative influence or detect shifts in performance dynamics.

The strictly increasing nature of the cumulative score also prevents the representation of performance decay, momentum reversals, or tactical suppression: a player whose influence declines due to fatigue or tactical adjustments will still exhibit an artificially monotonic performance trajectory. Therefore, while PlayeRank's cumulative formulation is adequate for full-match evaluation, it is insufficient for temporal segmentation. A temporally meaningful metric must reflect evolving game states, substitutions, and local contributions, rather than indiscriminately accumulating actions.

Recent work supports this perspective. \citet{SchmidtLilloBustos2024} demonstrate that player influence fluctuates meaningfully over a match due to tactical reconfiguration, opponent pressure, and physical factors, advocating for temporal windowing to capture such dynamics. Following this insight, the present work also adopts 5-minute temporal slices, but advances the methodology by adding a role-aware percentile cumulative mean across slices. This modification preserves interpretability while enabling direct comparability between players with different playing times, eliminating residual exposure bias and allowing the performance curve to reflect true temporal variability.

A second methodological challenge concerns the interpretation of imprecise or uncertain information in decision-making contexts. The literature highlights fuzzy logic as an effective tool for this purpose. For instance, Huarachi-Macuri et al. \citet{LACCEI2023} demonstrate that fuzzy systems are suitable for transforming quantitative descriptors into interpretable linguistic assessments (e.g., Stamina, Agility). Within tactical modeling, \citet{Marliere2017} uses fuzzy control systems to arbitrate between tactical states under uncertainty. These contributions motivate the role of fuzzy logic in the present system: integrating performance indicators and contextual factors into a Substitution Priority score in a manner compatible with the inherent uncertainty of in-game decision making.

The decision to make substitutions in soccer remains an intrinsically human domain, hardly replicable by machine learning (ML) algorithms. While ML can analyze historical performance data, a coach's choice involves a complex assessment of contextual, tactical, and psychological factors that are not easily quantifiable—such as team morale, perceived fatigue, reading the opponent's strategy, and game momentum. Attempts to predict substitutions using ML, while informative, reveal a clear ceiling in their ability to capture this complexity. A recent study by \citet{mohandas}, for example, analyzed a large dataset of 51,738 substitutions using multiple algorithms, including Random Forest, SVM, and Decision Trees. Even with this wealth of data, the best-performing model (Random Forest) achieved a maximum accuracy of just over 70\%. This gap of approximately 30\% is significant, suggesting that the final decision is not purely predictive but rather adaptive, relying on subjective human judgment that historical data alone cannot capture---precisely the structural ceiling formalized in Corollary~2.1. Consequently, models attempting to predict the optimal substitution are conceptually limited. Alternative approaches, such as fuzzy logic-based decision support systems, appear more suitable, as they are designed not to replace but to support the coach's judgment, managing the inherent uncertainty and qualitative factors of the match.

Finally, the choice of an intrinsically interpretable methodology is not merely a design preference, but a methodological necessity in high-stakes decision-making contexts. As explicitly argued by \citet{Rudin2019}, post-hoc explanations of black-box models are fundamentally inadequate when decisions have significant consequences, as they obscure the true decision logic and may provide misleading justifications---formalized above as Corollary~7.1. In domains such as elite soccer, where substitution decisions carry substantial sporting and financial impact, decision support systems must expose their reasoning process in a transparent and auditable manner. This perspective reinforces the limitations of purely predictive machine learning approaches discussed above and provides a principled justification for abandoning black-box optimization in favor of symbolic reasoning. Accordingly, the present work adopts fuzzy logic as the core inference mechanism—rather than as an explanatory add-on—ensuring that substitution recommendations are interpretable, accountable, and directly aligned with domain knowledge and human tactical judgment \citep{Rudin2019}.

This perspective finds foundational support in the framework of logical dynamics proposed by \citet{vanBenthem2011}, which conceptualizes rational agency as a continuous process of information update rather than static optimization. In this context, player substitutions serve as a particularly clear instance of \emph{epistemic actions} \citep{vanBenthem2007}: they do not merely alter the physical parameters of the match, but fundamentally restructure the space of plausible future game states under severe informational constraints. At the moment of decision, the coach operates at the boundary between ``hard facts'' (e.g., scoreline, remaining time, substitution limits) and ``soft information'' (e.g., perceived fatigue, tactical momentum). Once executed, a substitution irreversibly fixes certain constraints while reweighting the plausibilities regarding future play. Consequently, decision quality cannot be assessed solely by ex-post outcomes, but by whether the action constituted a rational epistemic update given the information available at time $t$. This framing aligns the proposed system with a dynamic view of rationality—where correctness is understood as the capacity for informed correction under uncertainty—establishing its role not as a predictor of future success, but as an auditor of the epistemic transition induced by the substitution.

Crucially, the proposed approach is hybrid in a \emph{functional} rather than
predictive sense. The statistical components of the system are not employed to
forecast future match outcomes or to optimize predictive accuracy. Instead, they
serve exclusively to construct a temporally normalized, role-aware representation
of the current decision state. The symbolic layer does not refine or correct
predictions; it operates as an external normative auditor, evaluating whether
continued action is justified given the evidence available at time $t$. In this
architecture, hybridization exists to support interpretability, decision auditing,
and accountability—not predictive performance.

\subsection{Description of the Problem}

\subsubsection{Problem Formulation}
Tactical decision-making in elite soccer operates under high uncertainty and significant financial stakes \cite{GE_PremierLeague, ESPN_Mundial}. Despite the proliferation of granular data, the specific process of substitution decisions remains predominantly intuitive or reliant on descriptive statistics that fail to capture real-time performance decay. This creates two distinct problems: the \textit{Metric Exposure Bias} and the \textit{Predictive Ceiling}.

\textbf{The Metric Exposure Bias:} Existing frameworks for player evaluation, such as the original PlayeRank \cite{Pappalardo2019}, typically utilize cumulative sum formulations. While effective for post-match rankings, these metrics introduce a temporal bias during live games: a player's score monotonically increases with playing time, regardless of their minute-by-minute efficiency. This mathematical structure masks declining performance, as a fatigued player performing poorly in the $80^{th}$ minute may still have a higher total score than a high-impact substitute, rendering standard metrics insufficient for real-time substitution decisions.

\textbf{The Predictive Ceiling of Machine Learning:} Current computational approaches to substitutions rely heavily on Supervised Machine Learning (SML) models trained on historical data. As noted by \cite{mohandas}, these models achieve a prediction accuracy plateau of approximately 70\%. This "ceiling" exists because SML models are designed to mimic human behavior, thereby learning and replicating the cognitive biases of coaches (e.g., status quo bias or delaying defensive changes)---an empirical instance of Theorem~2. Consequently, these models validate historical decisions rather than optimizing future outcomes, failing to identify necessary substitutions that deviate from conservative human norms.

Therefore, the core problem this study addresses is the lack of an objective, prescriptive auditing tool that can quantify intra-match performance decay without exposure bias and signal tactical risks independently of historical human tendencies.

\subsubsection{Research Hypotheses}
To address the formulated problem, this study tests the following hypotheses:

\begin{itemize}
    \item \textbf{H1 (Metric Hypothesis):} 
    Reformulating performance evaluation from a cumulative sum to a \textit{Cumulative Mean with Role-Aware Normalization} effectively eliminates exposure bias, allowing the system to detect negative momentum and performance drops that are mathematically invisible in additive models.
    
    \item \textbf{H2 (Prescriptive Auditing Hypothesis):} 
    A \emph{prescriptive} decision support system, acting as an objective auditor of performance and contextual variables (e.g., performance trends, fatigue, disciplinary exposure), can identify high-risk tactical scenarios—such as latent defensive liabilities—earlier and more reliably than human intuition or data-driven predictive models trained on historical behavior.
    
    \item \textbf{H3 (Contextual Modulation Hypothesis):} 
    The modulation of disciplinary risk by tactical role (e.g., weighting a yellow card more heavily for a defender than a forward) significantly alters the substitution priority score, aligning the system's output with expert tactical consensus in critical defensive scenarios.
    
    \item \textbf{H4 (Prescriptive AI Formalization Hypothesis):} 
    The empirical validation of a prescriptive, auditing-based decision support system in a high-stakes environment---and its independent structural corroboration in MYCIN and NEWS2 (Section~\ref{sec:mycin_news2})---supports the formalization of a distinct \emph{Prescriptive AI} paradigm, in which the system objective is not to predict human decisions, but to support normative, risk-aware judgment through explicit reasoning and accountability.
\end{itemize}

\subsection{Dataset}

The database selected for this project is the \textit{"Soccer match event dataset"}, a detailed public repository of soccer match events \citep{Pappalardo2020}. The choice of this dataset is based on its high granularity. The \textit{dataset} details player-level actions and aggregated performance metrics, allowing for the modeling of \textit{in-game} situations, which is an essential requirement for our system. We used the complete \textit{dataset} available on Kaggle \citep{Pappalardo2020}, which comprises 27 interrelated CSV tables. They cover seven main competitions and can be grouped into five logical categories:

\begin{itemize}
    \item \textbf{Match Data (\texttt{matches\_*.csv}):} Contains information about the games, such as dates, lineups, substitutions, results, and tactical formations.
    \item \textbf{Event Data (\texttt{events\_*.csv}):} The core of the \textit{dataset}, recording millions of individual actions on the field.
    \item \textbf{Entity Data (\texttt{players.csv}, \texttt{teams.csv}, etc.):} Dimensional tables with demographic and static data. 
    \item \textbf{Performance Metrics (\texttt{playerank.csv}):} A pre-processed file that provides the \texttt{playerankScore}, a multidimensional and role-aware performance evaluation metric.
    \item \textbf{Dictionaries (\texttt{tags2name.csv}, \texttt{eventid2name.csv}):} Metadata that translate event and \textit{tag} IDs into readable descriptions (e.g., Tag \texttt{1702} = \texttt{'yellow\_card'}).
\end{itemize}

The choice of \texttt{playerankScore} as our main "Performance" input is a central methodological decision. As proposed by \citet{Pappalardo2019}, the PlayeRank \textit{framework} was developed to solve the absence of a consolidated and universally accepted metric for evaluating player performance. The \texttt{playerankScore} is a metric derived from millions of game events that, according to the authors, surpasses other metrics when compared with assessments from professional scouts. Therefore, instead of trying to model performance from raw events, we adopted the \texttt{playerankScore} as an already validated and academically robust representation of a player's performance in a match. \\

\noindent Dataset available at: \url{https://www.kaggle.com/datasets/aleespinosa/soccer-match-event-dataset/data?select=playerank.csv}

\subsection{Data Integration and Pre-processing}

The pre-processing pipeline transforms heterogeneous soccer event logs into a unified temporal representation of player performance suitable for fuzzy inference. The process unfolds in three sequential stages: temporal performance computation, contextual and demographic enrichment, and role-aware normalization. Each stage progressively increases the dataset's semantic density and interpretability.

The objective of this pipeline is to transform the raw multi-source event data (events, matches, players) into a single structured and temporal dataset tailored for the Fuzzy Logic Decision System. Unlike the original \textit{PlayeRank} framework \citet{Pappalardo2019}, which aggregates player performance per match, our system requires a fine-grained temporal perspective of the player's performance evolution within a match. To achieve this, we implemented a three-stage processing pipeline that generates a dense temporal dataset, where each record represents the state of a player in a 5-minute (300-second) interval of play.

\subsubsection{Stage 1: Temporal Performance Metric Generation}

The first stage (\texttt{playerankdatasetV4.py}) processes the raw event logs (\texttt{events\_*.csv}) and converts each action's timestamp (\(\texttt{eventSec}\), \(\texttt{matchPeriod}\)) into an absolute measure of seconds from the start of the match (\(\texttt{total\_seconds}\)). The match timeline is then discretized into 5-minute intervals, associating each event with its corresponding temporal slice. The following variables are created in this stage:

\begin{itemize}
    \item \textbf{\texttt{matchId}}, \textbf{\texttt{teamId}}, \textbf{\texttt{playerId}}: extracted directly from the event logs, providing the hierarchical identifiers for each observation.
    \item \textbf{\texttt{cartao\_amarelo}}: a binary state indicator (using \texttt{expanding().max()}) equal to 1 if the player received a yellow card at any point up to the current slice.
\end{itemize}

Within each interval, two complementary components are calculated:

\begin{itemize}
    \item \textbf{Technical Score (\texttt{score\_tecnico\_fatia}):} the aggregation of event-based actions (passes, shots, duels, interceptions) using the same weights of the original PlayeRank framework. Events are mapped to technical dimensions and normalized within the team context.
    \item \textbf{Network Score (\texttt{score\_rede\_fatia}):} computed using a directed pass graph per team and interval. Following the original PlayeRank framework, the \textit{Eigenvector Centrality} of each player is used to quantify their structural influence on ball circulation.
\end{itemize}

These two components are linearly combined to form the primary performance indicator for each temporal slice:

\[
\mathrm{playerank\_fatia\_raw} = (1 - \alpha_{\mathrm{net}})\cdot\mathrm{score\_tecnico\_fatia} + \alpha_{\mathrm{net}}\cdot\mathrm{score\_rede\_fatia}
\]

where $\alpha_{\mathrm{net}} = 0.2$ controls the relative contribution of the network-based component. This configuration was selected to balance structural influence and technical efficiency, aligning with the sensitivity ranges reported in \citet{Pappalardo2019}.

Subsequently, to ensure metric interpretability independent of absolute match intensity, the raw score is normalized against the historical distribution of the agent's specific tactical role. This transformation converts the unbounded scalar $\texttt{playerank\_fatia\_raw}$ into the \textbf{Role-Aware Percentile} ($\texttt{playerank\_fatia\_percentil}$), ensuring that a "High" score represents the same statistical rarity for a Defender as it does for a Forward.

\paragraph{Cumulative Metric Redefinition.}
To capture performance evolution through time, we define the cumulative mean score:

\[
\mathrm{playerank\_acumulativo\_media\_percentil} \ \ \textbf{=} \ \ \frac{1}{t}\sum_{i=1}^{t}\mathrm{playerank\_fatia\_percentil}^{(i)}
\]

This cumulative formulation departs from the additive model of the original PlayeRank, ensuring that longer playing times do not artificially inflate the total performance. The metric represents the player's average contribution up to time \(t\), maintaining comparability among players with different play durations.

\subsubsection{Stage 2: Contextual and Demographic Enrichment}

The second stage (\texttt{expandirdataset.py}) enhances the temporal dataset with contextual, situational, and demographic variables by joining multiple data sources.

\begin{itemize}
    \item \textbf{\texttt{player\_age}}: calculated by merging player metadata (\texttt{players.csv}) with match dates and computing the player's age on match day.
    \item \textbf{\texttt{player\_position}}: extracted from the player metadata, representing the tactical role (goalkeeper, defender, midfielder, forward).
    \item \textbf{\texttt{goals\_scored}}: identified from events where \texttt{eventName = "Shot"} and \texttt{subEventName = "Goal"} and accumulated over time.
    \item \textbf{\texttt{assists}}: extracted using Wyscout tag \texttt{id = 302} and accumulated over time using \texttt{groupby(['matchId', 'playerId']).cumsum()}.
    \item \textbf{\texttt{momentum\_rate}}: a measure of short-term performance trend, calculated as the difference between the current($t$) and the previous($t-1$) 5-minute $\text{\texttt{playerank\_acumulativo\_media\_percentil}}$ slices.
\end{itemize}

This enrichment process yields a semantically rich dataset combining technical, contextual, and demographic aspects for each player's temporal trace.

\subsubsection{Stage 3: Final Cleaning and Role-Aware Normalization}

The final stage (\texttt{limpezafinal.py}) standardizes the temporal data and recalculates cumulative metrics for stability and accuracy. The following operations are performed:

\begin{itemize}
    \item \textbf{removal of out-of-field periods:} Periods where it was identified that the player was substituted, or entered as a substitute, were removed to ensure that the metric remained accurate, not applied to goalkeepers.
    \item \textbf{Temporal Harmonization:} The temporal variables are standardized. \texttt{Tempo\_Partida} is set to represent the *end* of each 5-minute slice (e.g., 5, 10, 15...), and \texttt{minutes\_played} is recalculated using \texttt{cumcount()} to ensure a precise cumulative sum of on-field time.
    \item \textbf{Recalculation of Cumulative Means:} The variable \\ \texttt{playerank\_acumulativo\_media\_raw} is recalculated using a robust \texttt{expanding().mean()} to ensure that averages reflect the final, harmonized temporal trace.
    \item \textbf{Percentile Normalization:} two percentile-based variables are computed within each (\texttt{matchId}, \texttt{position}) group. Note: this normalization uses the raw \texttt{position} column (from the original data source) to group players by tactical role, ensuring methodological consistency:
    \[
    \begin{aligned}
    \mathrm{playerank\_fatia\_percentil} &= \mathrm{rank}_{\mathrm{pct}}(\mathrm{playerank\_fatia\_raw}) \\
    \mathrm{playerank\_acumulativo\_media\_percentil} &= \mathrm{rank}_{\mathrm{pct}}(\mathrm{playerank\_acumulativo\_media\_raw})
    \end{aligned}
    \]
\end{itemize}

This \textit{role-aware normalization} ensures that performance is assessed relative to tactical peers, preserving fairness and interpretability in subsequent fuzzy inference.

\paragraph{Final Dataset Variables.}

The resulting dataset (\texttt{Dataset\_limpo\_final.csv}) contains the following variables:
\begin{itemize}
    \item \textbf{Identifiers:} \texttt{matchId}, \texttt{playerId}, \texttt{teamId}.
    \item \textbf{Temporal Dimensions:} \texttt{Tempo\_Partida}, \texttt{minutes\_played}.
    \item \textbf{Performance Metrics:} \texttt{playerank\_fatia\_raw}, \texttt{playerank\_acumulativo\_media\_raw}, \texttt{score\_tecnico\_fatia}, \texttt{score\_rede\_fatia}.
    \item \textbf{Normalized Metrics:} \texttt{playerank\_fatia\_percentil},\\ \texttt{playerank\_acumulativo\_media\_percentil}.
    \item \textbf{Contextual Factors:} \texttt{momentum\_rate}, \texttt{cartao\_amarelo}.
    \item \textbf{Demographics:} \texttt{player\_age}, \texttt{player\_position}.
    \item \textbf{Offensive Contributions:} \texttt{goals\_scored}, \texttt{assists}.
    \item \textbf{Auxiliary:} \texttt{position}, a placeholder variable maintained for compatibility.
\end{itemize}

Each record corresponds to a 5-minute temporal snapshot of an individual player's on-field performance, integrating technical, contextual, and demographic information in a consistent temporal framework.

\subsubsection{Data Validation}

Before integration into the fuzzy system, consistency checks were performed. The recalculation of \texttt{playerank\_acumulativo\_media\_raw} in the final stage (Stage 3) ensures that all cumulative metrics are based on the robust \texttt{expanding().mean()} operator applied to the complete and harmonized temporal trace of each player. This guarantees that all cumulative scores represent valid in-game performance dynamics and that the fuzzy system operates on stable, correctly aggregated data. The experimental design yields high statistical density: the inference engine executes approximately 360 distinct decision audits (2 teams × 10 outfield players × 18 temporal windows). This granular approach allows us to evaluate the system's stability across diverse game states (winning/losing, high/low fatigue) within a controlled adversarial environment, ensuring internal validity beyond a simple match-outcome correlation.

\subsection{Exploratory Data Analysis}

The Exploratory Data Analysis (EDA) was conducted on the final integrated dataset, consisting of 805,146 temporal observations, 3,035 unique players, and 1,941 matches. The analysis aimed to (i) characterize the statistical properties of the input variables used in the fuzzy inference system, (ii) validate soccer-specific behavioral hypotheses (e.g., positional risk asymmetry), and (iii) provide empirical grounding for the design of the fuzzy universes of discourse.

\subsubsection{Descriptive Overview}

The dataset preserves a balanced representation of real-world soccer demographics and tactical structures. Table~\ref{tab:eda_summary} summarizes the player distribution and key descriptive statistics. 

\begin{table}[h]
\hspace{-1cm}

\caption{Dataset Summary and Descriptive Statistics}
\begin{tabular}{lccc}
\toprule
\textbf{Category} & \textbf{Value} & \textbf{Metric} & \textbf{Notes} \\
\midrule
Total observations & 805,146 & intervals & 5-minute resolution \\
Unique players & 3,035 & -- & across 7 competitions \\
Matches analyzed & 1,941 & -- & complete event coverage \\
\midrule
Mean/Median performance & 0.501 & std = 0.289 & centered around 0.5 \\
(\texttt{playerank...percentil}) & 0.501 & -- & symmetrical distribution \\
\midrule
Player positions & 4 roles & -- & role-aware normalization applied \\
\quad Goalkeeper & 233 & (5.8\%) & \\
\quad Defender & 1,032 & (25.8\%) & \\
\quad Midfielder & 1,120 & (28.0\%) & \\
\quad Forward & 650 & (16.3\%) & \\
\bottomrule
\end{tabular}
\label{tab:eda_summary}
\end{table}

This composition closely mirrors elite-level soccer distributions, where midfielders and defenders together comprise over half of the active roster, reflecting the denser tactical occupation of central zones.

\subsubsection{Univariate Distributions}

\begin{figure}[h]
\centering
\includegraphics[width=0.6\linewidth]{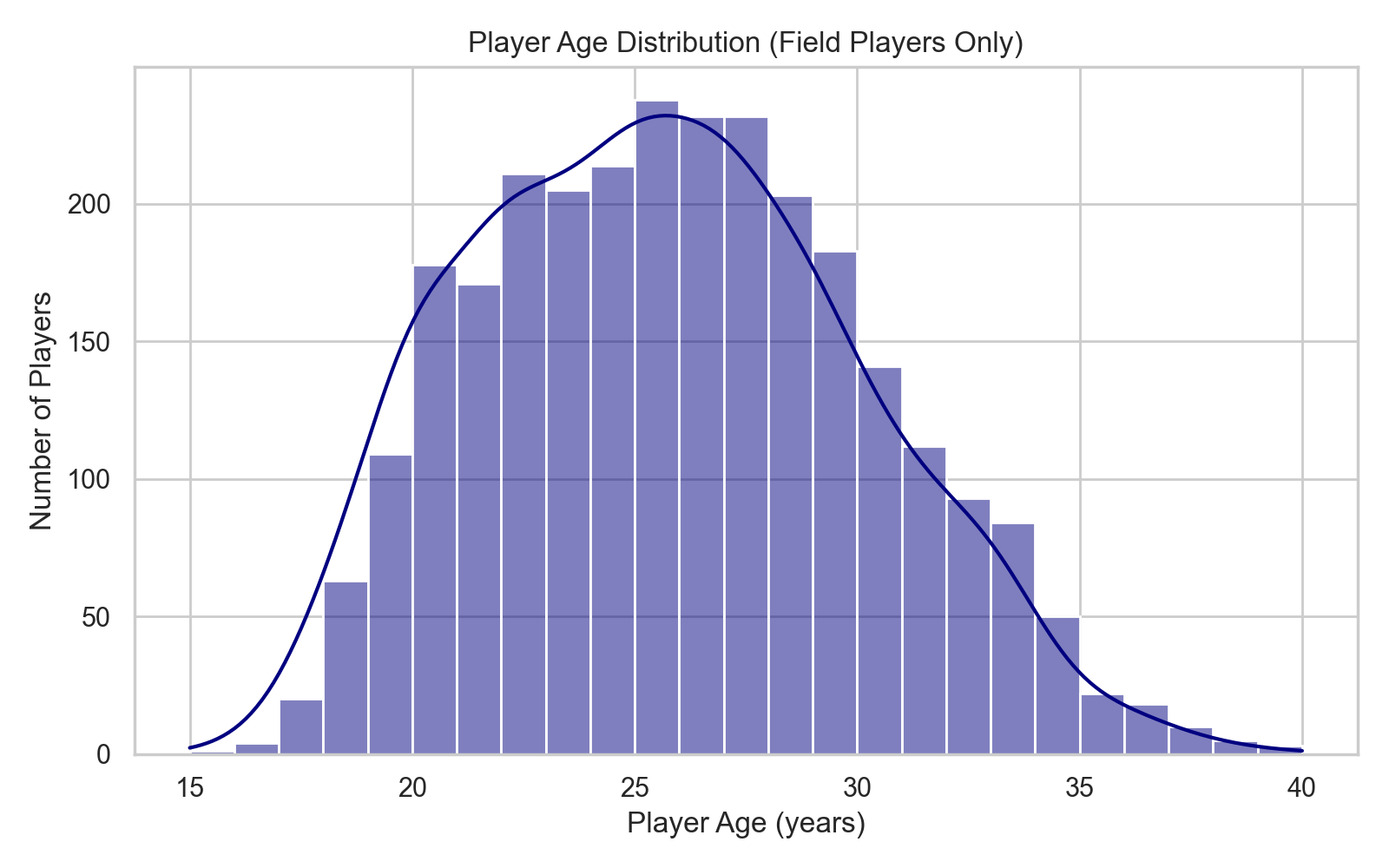}
\caption{Distribution of player ages in the dataset.}
\label{fig:eda_age}
\end{figure}

\paragraph{Player Age.}
The age distribution (Figure~\ref{fig:eda_age}) follows an approximately median centered around 26.0 years, with a range between 15.0 and 45.0 years. The kurtosis (2.91) and near-zero skewness confirm symmetry. This validates the fuzzy partitioning into \textit{Young}, \textit{Peak}, and \textit{Veteran} sets, with smooth transitions around the mean.

\paragraph{Performance (\texttt{playerank\_acumulativo\_media\_percentil})}
The per-slice performance scores exhibit a bounded, quasi-normal distribution centered around 0.5, with a standard deviation of 0.289. These characteristics make the metric well-suited for fuzzy modeling, where zero-centered or bounded scales allow intuitive linguistic partitioning (\textit{Low}, \textit{Medium}, \textit{High}) without rescaling distortions.

\subsubsection{Role-Based Event Distributions}

To further interpret positional behavior, the dataset was aggregated by player position to estimate empirical probabilities of key in-game events (goals, assists, and yellow cards).

The empirical distributions align with established tactical expectations: 

\begin{itemize}
    \item \textbf{Goal Rate:} Forwards lead with a 19.8\% probability of scoring per match, compared to 7.7\% for midfielders and 3.4\% for defenders. Goalkeepers are near zero, validating the offensive gradient embedded in the \textit{player\_position} variable.
    \end{itemize}

\begin{figure}[h]
\centering
\includegraphics[width=0.8\linewidth]{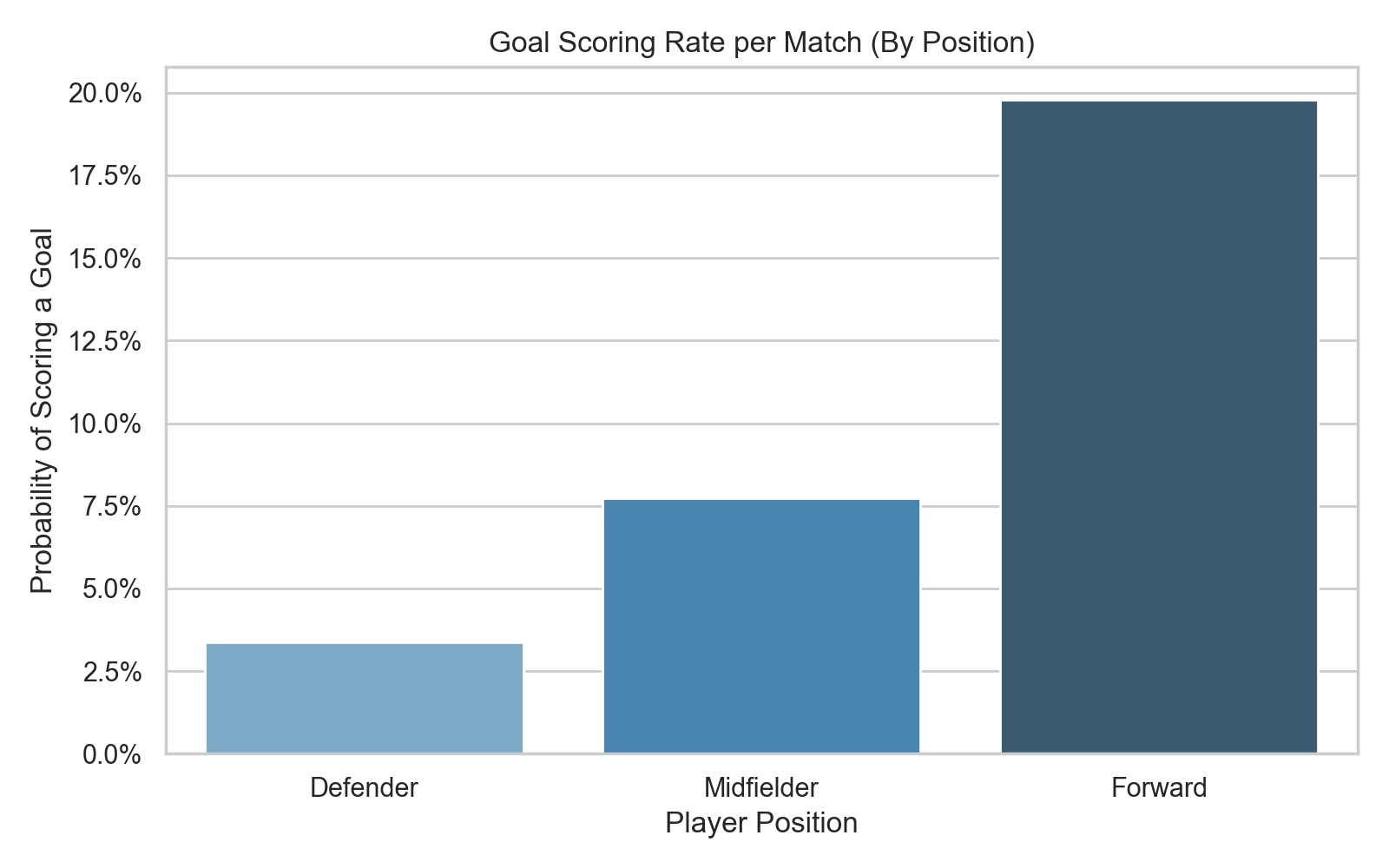}
\end{figure}

\newpage

\begin{itemize}
    \item \textbf{Assist Rate:} Forwards exhibit the highest assist probability (7.5\%), reflecting their creative and distributive role in ball progression. Midfielders follow closely (6.8\%), while defenders contribute the least(3.6\%).
    \end{itemize}

\begin{figure}[h]
\centering
\includegraphics[width=0.8\linewidth]{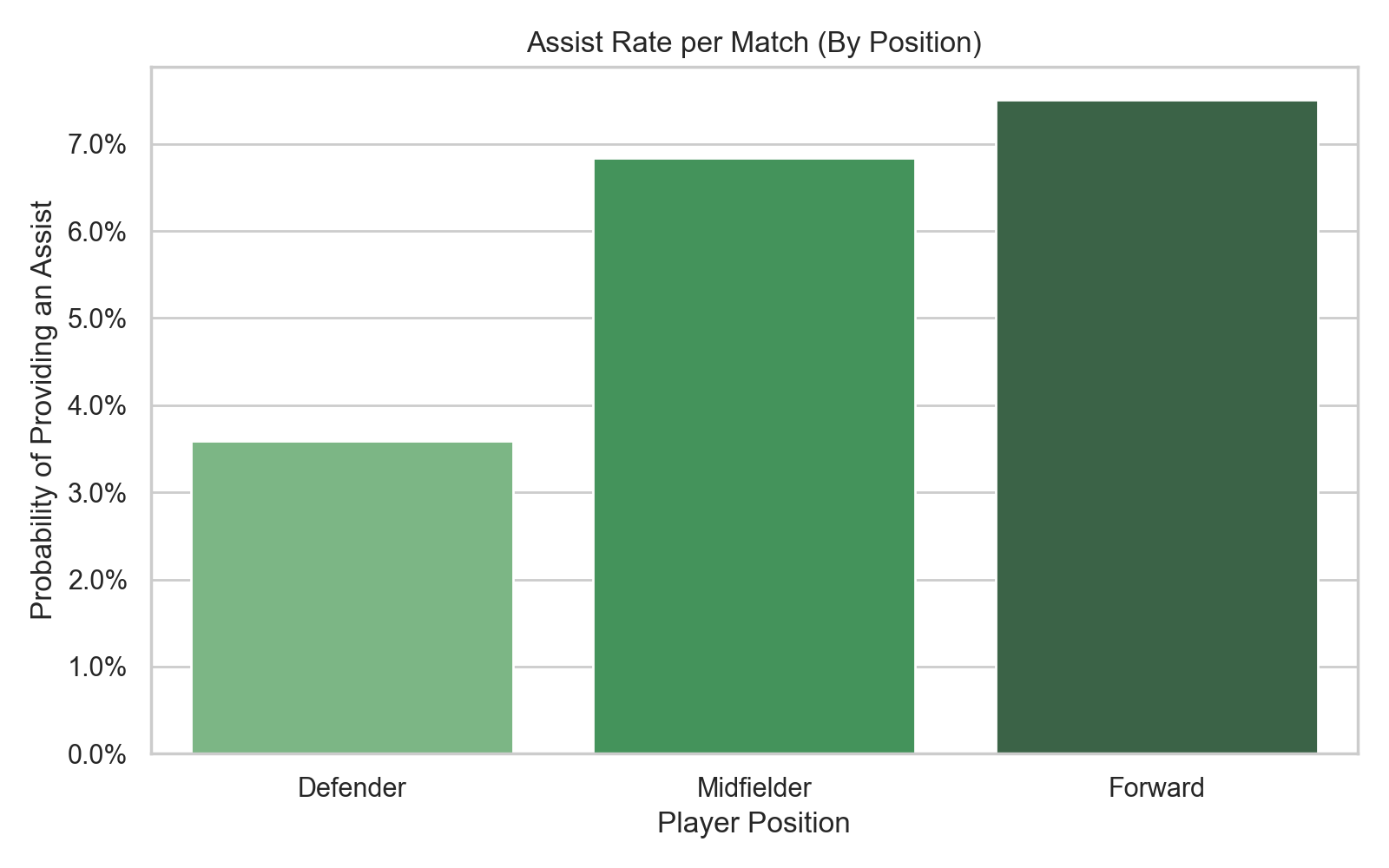}
\end{figure}

\begin{itemize}
    \item \textbf{Disciplinary Risk:} Defenders show the highest yellow card rate (17.2\%), followed by midfielders (14.8\%) and forwards (10.1\%), reinforcing the need for \textit{contextual fuzzy rules} that modulate card risk according to tactical role.
\end{itemize}
\begin{figure}[h]
\centering
\includegraphics[width=0.8\linewidth]{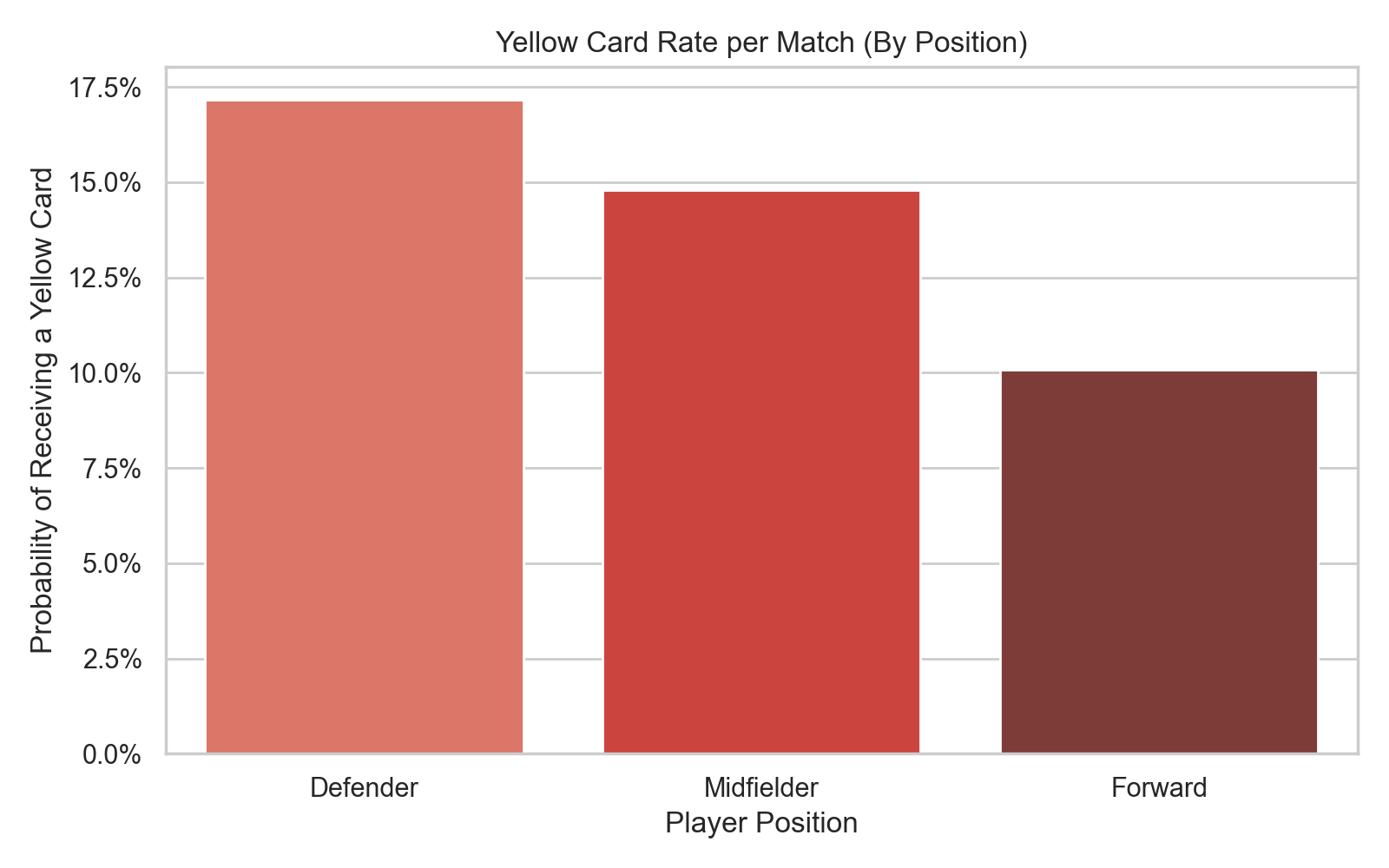}
\end{figure}

\subsubsection{Variable Definition, Independence and Correlations }
\paragraph{Variable Definition and Selection}

From the final integrated dataset (\texttt{Dataset\_limpo
\_final.csv}), nine variables were selected for exploratory analysis and subsequent use as inputs in the fuzzy decision model. These variables represent the main dimensions of in-game player state: performance, fatigue, disciplinary risk, offensive contribution, and tactical context. Table~\ref{tab:variables_used} summarizes the selected features, their measurement scale, and their conceptual role within the system.
\clearpage
\begin{table}[h!]
\centering
\footnotesize

\caption{Variables selected for exploratory analysis and fuzzy modeling.}
\begin{adjustbox}{max width=\textwidth}
\begin{tabular}{p{3cm} p{2.8cm} p{2cm} p{6cm}}
\toprule
\textbf{Dimension} & \textbf{Variable (ID)} & \textbf{Mathematical Set} & \textbf{Description and Relevance} \\
\midrule
\textbf{Technical Performance} & \texttt{P\_cum} & $[0, 1] \subset \mathbb{R}$ & Cumulative performance percentile per match and position; main indicator of global efficiency. \\[4pt]
\textbf{Fatigue} & \texttt{Min\_played} & $[0, 100] \subset \mathbb{R}$ & Total minutes played at each interval; proxy for physical wear. \\[4pt]
\textbf{Disciplinary Risk} & \texttt{Card\_Y} & $\{0, 1\} \subset \mathbb{Z}$ & Indicates if player received a yellow card; higher risk increases substitution priority. \\[4pt]
\textbf{Form Trend} & \texttt{Momentum} & $[-1, 1] \subset \mathbb{R}$ & Short-term rate of change in performance (3-interval moving average). Negative values denote declining form. \\[4pt]
\textbf{Contextual Fatigue (Age)} & \texttt{Age} & $[15, 45] \subset \mathbb{N}$ & Player's age on match day; modulates fatigue effects. \\[4pt]
\textbf{Tactical Role} & \texttt{Position} & $Categorical$ & Role label; used for role-aware normalization. \\[4pt]
\textbf{Offensive Contribution (Goals)} & \texttt{Goals} & $\mathbb{N} $ & Cumulative goals scored; high values indicate offensive importance. \\[4pt]
\textbf{Offensive Contribution (Assists)} & \texttt{Assists} & $\mathbb{N}$ & Cumulative assists; complements goal contribution. \\[4pt]
\bottomrule
\end{tabular}
\end{adjustbox}
\label{tab:variables_used}
\end{table}

\textbf{Labels:} 
\texttt{P\_cum = playerank\_acumulativo\_media\_percentil}, 
\texttt{Min\_played = 
minutes\_played}, 
\texttt{Card\_Y = cartao\_amarelo}, 
\texttt{Momentum = momentum\_rate}, 
\texttt{Age = player\_age}, 
\texttt{Position = player\_position}, 
\texttt{Goals = goals\_scored}, 
\texttt{Assists = assists},

These variables were derived directly from the processed PlayeRank-based dataset, ensuring temporal coherence across 5-minute intervals. The combination of continuous, discrete, and categorical features provides a multidimensional view of player state suitable for fuzzy inference. In the next subsection, we assess the independence and correlation among these variables to confirm their suitability for use as fuzzy inputs.

\paragraph{Spearman Correlation Matrix}

\begin{figure}[h]
    \centering
    \includegraphics[width=1\linewidth]{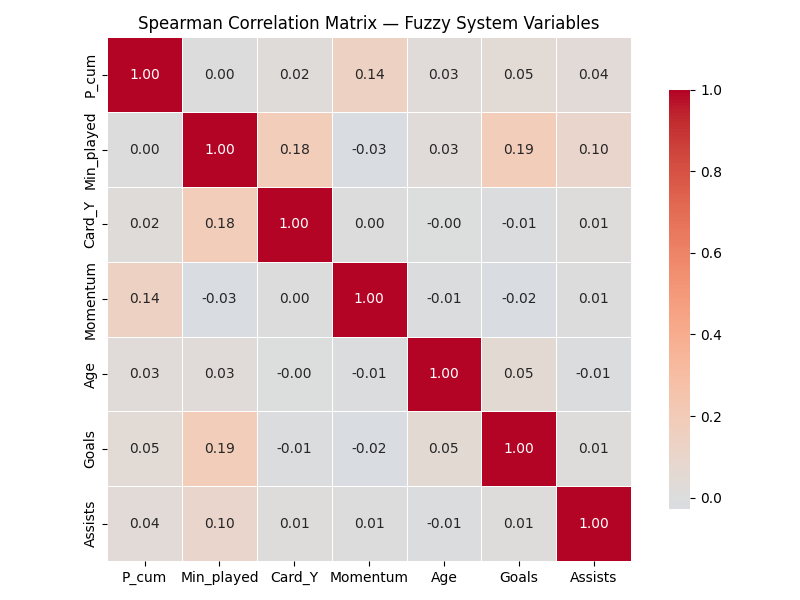}
    \caption{Spearman correlation matrix among the fuzzy-system input variables. The weak correlations indicate low redundancy and support the assumption of variable independence in fuzzy inference.}
    \label{fig:spearman-matrix}
\end{figure}

Figure~\ref{fig:spearman-matrix} reports the Spearman correlation matrix computed over all input variables adopted in the fuzzy inference system. The pairwise correlations are consistently weak, with magnitudes predominantly below $|\rho| < 0.20$. Such a pattern indicates that the variables do not exhibit meaningful monotonic dependence.
\clearpage

This low degree of association is desirable in the context of fuzzy inference. Since the rule base relies on linguistic terms and membership functions that encode distinct semantic dimensions of player performance, strongly correlated variables would introduce redundancy and reduce the discriminative power of the fuzzy rules. Conversely, the weak correlations observed here suggest that each variable contributes independent information to the inference process, thereby supporting a more expressive and interpretable fuzzy model.

\subsection{Fuzzy Control System}
\label{sec:fuzzy_system}

Based on the theoretical foundation and the need for dynamic real-time evaluation, we designed the Fuzzy Control System (FCS) to function as a \textit{Contextual Modifier}. Unlike traditional systems that calculate a raw output from zero, this architecture calculates a correction factor applied to a baseline performance metric. The system was implemented using the \texttt{scikit-fuzzy} library in Python, utilizing the Mamdani architecture. As established in Section~\ref{sec:mycin_news2}, this is the same architectural family---symbolic, rule-based, ex-ante specified---as MYCIN's production-rule engine, providing a within-paper structural comparison alongside the cross-domain one.

\subsubsection{Definition of Fuzzy Variables}

The system architecture has been expanded to process high-dimensional match data. It comprises eight input variables (antecedents) and one output variable (consequent). The universes of discourse were calibrated based on the range of values observed in the dataset and domain constraints (e.g., match duration).

\paragraph{Antecedents (Inputs)}

\begin{enumerate}
    \item \textbf{Cumulative Performance (\texttt{P\_cum}):} Represents the player's average percentile performance up to the current moment.
        \begin{itemize}
            \item Universe: [0.0, 1.0].
            \item Sets: \textit{VeryLow}, \textit{Low}, \textit{Medium}, \textit{High}, \textit{VeryHigh}.
        \end{itemize}
    \item \textbf{Momentum (\texttt{Momentum}):} Rate of change in performance, indicating if the player is improving or declining.
        \begin{itemize}
            \item Universe: [-1.0, 1.0].
            \item Sets: \textit{Falling}, \textit{Stable}, \textit{Rising}.
        \end{itemize}
    \item \textbf{Fatigue (\texttt{Min\_played}):} Minutes played in the match.
        \begin{itemize}
            \item Universe: [0, 100] minutes.
            \item Sets: \textit{Low} (0-45'), \textit{Medium} (40-80'), \textit{High} (70-100').
        \end{itemize}
    \item \textbf{Age (\texttt{Age}):} Player's chronological age.
        \begin{itemize}
            \item Universe: [15, 45] years.
            \item Sets: \textit{Young}, \textit{Peak}, \textit{Veteran}.
        \end{itemize}
    \item \textbf{Match Events (Categorical/Integer):}
        \begin{itemize}
            \item \textbf{Card\_Y:} Yellow card status [0, 1] (\textit{Yes}).
            \item \textbf{Goals:} Goals scored [0, 10] (\textit{None, Some, Many}).
            \item \textbf{Assists:} Assists provided [0, 10] (\textit{None, Some, Many}).
        \end{itemize}
    \item \textbf{Positional Context (Binary):}
        \begin{itemize}
            \item Variables: \texttt{is\_Defender}, \texttt{is\_Midfielder}, \texttt{is\_Forward}.
            \item Sets: \textit{Yes} (indicating the player's role to activate specific rules).
        \end{itemize}
\end{enumerate}

\paragraph{Consequent (Output)}

\begin{itemize}
    \item \textbf{Modifier Value (\texttt{Modifier\_Value}):} A correction factor ranging from negative (protection/priority reduction) to positive (urgency/priority increase).
        \begin{itemize}
            \item Universe: [-100, 100].
            \item Granularity: 9 sets ranging from \textit{VLN} (Very Large Negative, -100) to \textit{LP\_70} (Large Positive, +70).
            \item \textit{Zero} represents no adjustment.
        \end{itemize}
\end{itemize}

\subsubsection{Membership Functions}

The membership functions (MFs) mix Trapezoidal (\texttt{trapmf}) and Triangular (\texttt{trimf}) shapes to capture specific tactical thresholds. Key definitions from the implementation include:

\begin{itemize}
    \item \textbf{P\_cum:} \textit{Low} uses a trapezoid \texttt{[0, 0, 0.10, 0.35]} to capture distinctively poor performance, while \textit{High} starts at 0.65.
    \item \textbf{Momentum:} \textit{Falling} is defined strictly in the negative range \texttt{[-1.0, -1.0, -0.03, -0.01]}, ensuring only genuine performance drops trigger the logic.
    \item \textbf{Fatigue:} Overlapping sets allow smooth transitions. \textit{High} fatigue begins notably at 70 minutes, aligning with common substitution windows.
    \item \textbf{Contextual Events:} Variables like \texttt{is\_Defender} or \texttt{Card\_Y} use pseudo-binary trapezoids (e.g., \texttt{[0.5, 1, 1, 1.5]}) to function as logical switches within the fuzzy inference engine.
\end{itemize}

\subsubsection{Rule Base }

The rule base was streamlined to remove noise and focus on Intensifying or attenuating the player's performance using other contextual events as variables. The system avoids oscillation by ignoring minor fluctuations and focuses on critical states. The logic integrates position and stats directly.
\clearpage
\begin{table}[h!]
    \centering
    \small
    
    \caption{Key Fuzzy Rules for Substitution Priority}
    \label{tab:rules_v22}
    \begin{adjustbox}{max width=\textwidth}
    \begin{tabular}{ll}
    \toprule
    \textbf{ID} & \textbf{Rule Logic} \\
    \midrule

    \textbf{R01 -- Untouchable Star} &
    IF (P\_cum is High/VeryHigh) THEN (Modifier = Negative). \\

    \textbf{R02a -- Critical Fatigue} &
    IF (P\_cum Low/VeryLow) AND (Min High) THEN (Modifier = Large Positive). \\

    \textbf{R02b -- Early Fatigue} &
    IF (P\_cum Low/VeryLow) AND (Min Medium) THEN (Modifier = Medium Positive). \\

    \textbf{R03 -- Defensive Risk} &
    IF (Defender AND YellowCard Yes) THEN (Modifier = Medium Positive). \\

    \textbf{R04 -- Rapid Decline} &
    IF (P\_cum Low/VeryLow AND Momentum Falling) THEN (Modifier = Large Positive). \\

    \textbf{R07 -- Positive Momentum} &
    IF (P\_cum High/VeryHigh AND Momentum Rising) THEN (Modifier = Medium Negative). \\

    \textbf{R08 -- Ineffective Forward} &
    IF (Forward AND P\_cum Low AND Goals None) THEN (Modifier = Large Positive). \\

    \textbf{R09 -- Striker Under Pressure} &
    (1) IF (Forward AND P\_cum Low/Med AND Momentum Falling) THEN (Modifier = VeryLarge Positive). \\
    & (2) IF (Forward AND P\_cum Low/Med AND Min High) THEN (Modifier = VeryLarge Positive). \\

    \textbf{R10 -- Invisible Playmaker} &
    IF (Midfielder AND P\_cum Low AND Assists None) THEN (Modifier = Med--Large Positive). \\

    \textbf{R11 -- Creator Bonus} &
    IF (Assists Some OR Goals Some) THEN (Modifier = Medium Negative). \\
    & IF (Assists Many OR Goals Many) THEN (Modifier = Large Negative). \\

    \textbf{R12 -- Veteran Fatigue} &
    IF (Age Veteran AND Min High) THEN (Modifier = Medium Positive). \\

    \textbf{R13 -- Young Talent Protection} &
    IF (Age Young AND (Goals Some/Many OR Assists Some/Many)) \\
    & THEN (Modifier = VeryLarge Negative). \\

    \textbf{R14 -- Goal Protection} &
    IF (Goals Some) THEN (Modifier = Medium Negative). \\
    & IF (Goals Many) THEN (Modifier = Large Negative). \\

    \textbf{R15 -- Neutral State} &
    IF (P\_cum Medium AND Momentum Stable) THEN (Modifier = Zero). \\
    \bottomrule
    \end{tabular}
    \end{adjustbox}
\end{table}

\subsubsection{Inference and Final Calculation}

The system employs a hybrid calculation model. The Fuzzy Inference System (FIS) computes the \texttt{Modifier\_Value} using the Centroid method. However, this value is not the final priority. The final substitution priority $P_{final}$ is calculated by applying the modifier to a baseline inverse of the cumulative performance, scaled by a factor $\alpha$:

\begin{equation}
    \text{Baseline} = 100 \times (1.0 - P_{\text{cum}})
\end{equation}

\begin{equation}
    P_{\text{final}} = \text{clip}\left( \text{Baseline} + (\text{Modifier} \times \alpha), \ 0, \ 100 \right)
\end{equation}

Where $\alpha = 0.25$ in the current version. This ensures that the fuzzy logic acts as a "tuner" that intensifies or attenuates the urgency based on tactical context (e.g., a yellow card or a goal scored), rather than overriding the player's core performance metric entirely. This methodology allows the coach's in field knowledge to be faithfully captured by the system, adding significant weight to the player's performance on the field. In the terms of Axiom~3 and Section~\ref{sec:normative}, $P_{\mathrm{final}}$ instantiates the state-based quality function $f(s_t,\calN)$, and Rules R01--R15 jointly constitute the justification $E$ required by Axiom~2.

\section{Results}
\label{sec:results}

To evaluate the utility of prescriptive auditing in high-stakes environments, we contrast the proposed Fuzzy Control System (FCS) against state-of-the-art predictive baselines and subject it to a rigorous stress-test using real-world data from elite soccer. Section~\ref{sec:mycin_news2} already reported the corresponding cross-domain validation evidence for MYCIN and NEWS2; the results below concern the soccer instantiation specifically.

\subsection{Overcoming the Predictive Ceiling: Prescription vs. Prediction}

Recent benchmarks in soccer analytics reveal a \emph{predictive ceiling} for substitution modeling. \citet{mohandas} analyzed 51,738 substitutions using supervised learning algorithms such as Random Forest and SVM, achieving a maximum accuracy of approximately $70\%$. This plateau indicates that nearly 30\% of tactical decisions are influenced by contextual nuances—such as fatigue thresholds or specific game plans—that remain opaque to black-box models.

Supervised learning approaches in this domain suffer from a fundamental normative limitation: they are trained to \emph{mimic} human experts. If historical data encode risk aversion, delayed reactions, or other systematic biases, the resulting models replicate them rather than correct them---exactly the mechanism formalized by Theorem~2 and bounded by Corollary~2.1. As summarized in Table~\ref{tab:fuzzy_vs_ml}, our proposed framework shifts the objective from \emph{minimizing prediction error} to \emph{maximizing tactical utility}, acting as an auditor rather than an imitator.

Theoretical considerations further reinforce this distinction. As established in Theorem~2, constructing or benchmarking against an additional predictive substitution model does not provide a meaningful comparison. Predictive systems, by design, optimize for behavioral imitation or outcome likelihood. Consequently, they evaluate decisions with respect to variables that are epistemically unavailable at the moment of commitment. As formally shown in Theorem~5 and Corollary~5.1, such systems are inherently incapable of auditing decision quality without reintroducing outcome bias, and are moreover subject to the deeper Normative Non-Identifiability result of Theorem~6: even attempting to recover the coach's implicit utility function from historical substitution logs, e.g. via inverse reinforcement learning, would provably concentrate on the biased historical pattern rather than the normatively optimal one (Corollary~6.1). Any predictive baseline—regardless of architectural sophistication or empirical performance—would therefore inherit the same normative limitation identified in prior work \citep[e.g.,][]{mohandas}: replication of historical human biases rather than their diagnosis.

From this perspective, the relevant scientific question is not whether a stronger predictor could be engineered, but whether prediction is an appropriate objective for evaluating decisions under uncertainty. Our prescriptive framework answers this question in the negative by evaluating choices according to \emph{normative coherence} with the epistemic state at time $t$. Requiring predictive parity as a validation standard would constitute a category error, conflating behavioral forecasting with normative decision auditing.

\begin{table}[h]
\centering

\caption{Structural Comparison: Supervised ML vs. Prescriptive Fuzzy System}
\label{tab:fuzzy_vs_ml}
\resizebox{\textwidth}{!}{%
\begin{tabular}{lcc}
\toprule
\textbf{Dimension} & \textbf{Supervised ML Baseline (e.g., \citet{mohandas})} & \textbf{Proposed Prescriptive System} \\
\midrule
\textbf{Objective} & Minimize Prediction Error (Mimic Human) & Maximize Tactical Utility (Audit Human) \\
\textbf{Handling Uncertainty} & Probabilistic (Confidence Scores) & Possibilistic (Membership Degrees) \\
\textbf{Interpretability} & Low (Feature Importance/Black Box) & High (Linguistic Rules) \\
\textbf{Outcome Bias} & High (Dependent on realized results) & Decoupled (State-based evaluation) \\
\bottomrule
\end{tabular}%
}
\end{table}

\subsection{The Mathematical Mechanism: Eliminating Exposure Bias}

A fundamental prerequisite for effective auditing is the ability to detect performance deterioration in real-time. Traditional metrics typically utilize cumulative sum formulations, where a player's score monotonically increases with playing time regardless of minute-by-minute efficiency. Mathematically, these additive models act as \emph{low-pass filters}, masking declining performance because a fatigued player performing poorly in the $80^{th}$ minute may still possess a higher total score than a high-impact substitute.

To resolve this \emph{Metric Exposure Bias}, we reformulated the evaluation metric from a cumulative sum to a \textbf{Cumulative Mean with Role-Aware Normalization} ($P_{cum}$). By calculating the average contribution up to time $t$ and normalizing it against the historical distribution of the agent's specific tactical role, our metric functions as a \emph{high-pass audit}. This transformation allows the performance curve to reflect true temporal variability, exposing negative momentum and performance drops that are mathematically invisible in additive models. This methodological shift is what enables the system to detect the latent risks detailed in the following case studies.

\subsection{Stress-Testing Decision Latency: The Brazil vs. Belgium Case}

We applied the framework to the high-stakes elimination match between Brazil and Belgium (2018 FIFA World Cup) to observe decision latency and normative disagreement. The system outputs ($P_{final}$) were generated in 5-minute intervals.

\clearpage

\begin{figure}[h]
    \centering
    \includegraphics[width=0.9\linewidth]{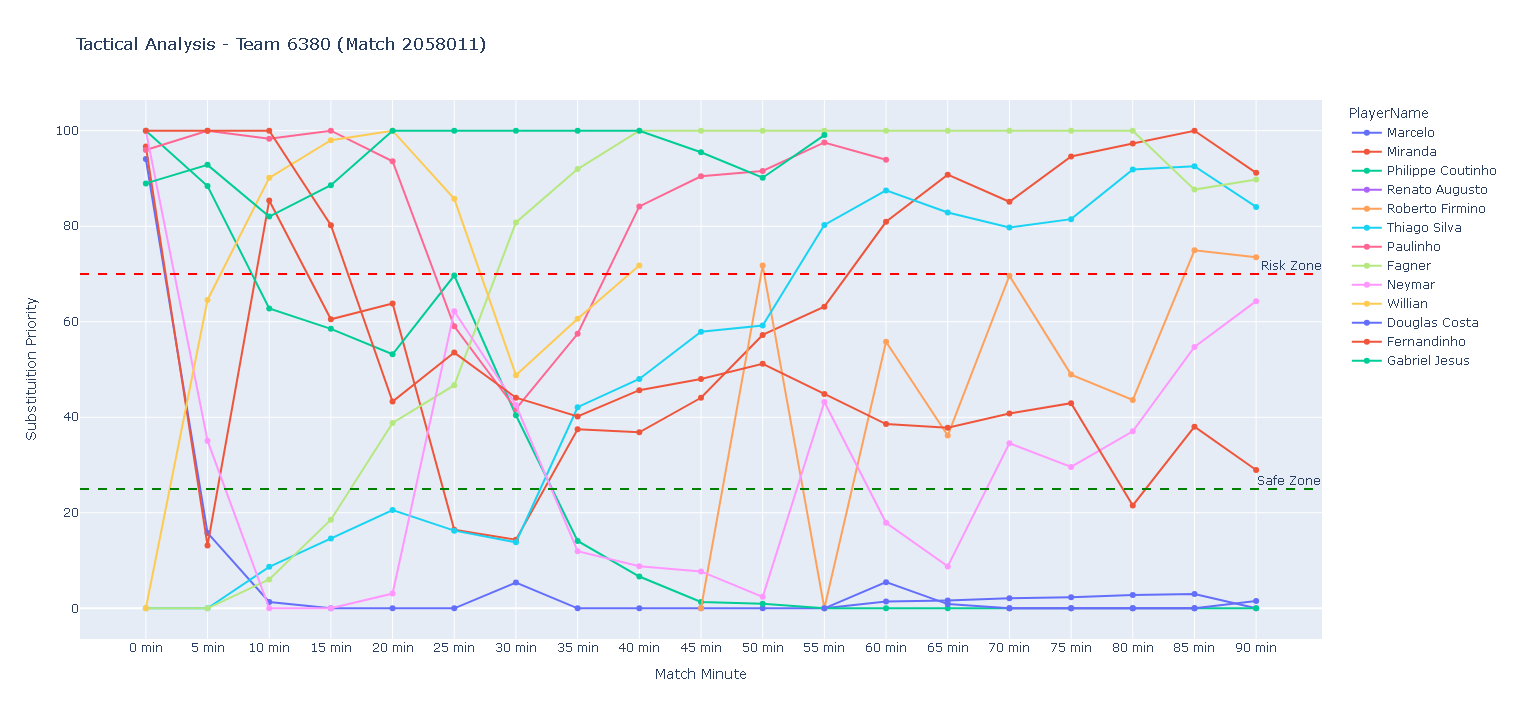}
    \caption{Temporal evolution of Substitution Priority for Brazil. High values indicate critical inefficacy or risk. Note the divergence in the case of Fagner (Red Dashed Line) compared to executed substitutions.}
    \label{fig:brasil_plot}
\end{figure}

\subsubsection{Alignment and Early Detection}
The system successfully captured decision-relevant deterioration prior to human intervention (Table \ref{tab:results_summary}).
\begin{itemize}
    \item \textbf{Willian (45'):} Priority $P_{final}=72.0$, ranked 4\textsuperscript{th}. Rule R04 (\textit{Rapid Drop}) flagged a sharp decline in technical actions in the final 20 minutes of the first half. Immediate substitution by the coach confirms offensive ineffectiveness, supported by media reports \cite{ESPN_2018}.

    \item \textbf{Gabriel Jesus (58'):} Priority $P_{final}=99.1$, 2\textsuperscript{nd} highest. Rule R08 (\textit{Ineffective Forward}) triggered by low cumulative score and no goals/assists. Post-match data: 12 touches in 58 minutes \cite{Epoca_2018}, validating tactical isolation.

    \item \textbf{Paulinho (73'):} Priority $P_{final}=93.1$. System detected "Early Fatigue" and low participation. External analysis noted minimal offensive impact \cite{Gazeta_2018}.
\end{itemize}

\subsubsection{The ``Fagner Paradox'': Counterfactual Auditing}
The most significant finding is the disagreement regarding right-back Fagner. While the human decision-maker retained the player, the system assigned him \textbf{Maximum Priority (100.0)} from minute 45 to 85.
This was not a prediction error, but a \textit{risk audit}. The system aggregated consistently low technical performance with high defensive exposure against opponent Eden Hazard. The realized outcome—Fagner struggling in duels and receiving a yellow card at the 90th minute—validates the system's ``Critical Risk'' diagnosis. Consistent with Theorem~2, a predictive ML model trained on this coach's history would likely have predicted ``No Substitution'' (True Positive), thereby reinforcing the status quo bias that led to the vulnerability.

\begin{table}[h]
    \centering
    
    \caption{Model Audit vs. Human Decision (Brazil vs. Belgium)}
    \label{tab:results_summary}
    \resizebox{\textwidth}{!}{%
    \begin{tabular}{lccccc}
        \toprule
        \textbf{Player} & \textbf{Analysis Slice($t$)} & \textbf{$P_{\text{final}}$} & \textbf{Rank} & \textbf{Diagnosis} & \textbf{Real Decision} \\
        \midrule
        \textbf{Willian} & 40'-45' & 72.0 & 4\textsuperscript{th} & Rapid Drop & Substituted out \\
        \textbf{Gabriel Jesus} & 55'-60' & 99.1 & 2\textsuperscript{nd} & Ineffective Forward & Substituted out \\
        \textbf{Fagner} & 45'-85' & \textbf{100.0} & \textbf{1\textsuperscript{st}} & \textbf{Critical Risk} & \textbf{Not Substituted} \\
        \textbf{Paulinho} & 65'-73' & 93.1 & 3\textsuperscript{rd} & Early Fatigue & Substituted out \\
        \bottomrule
    \end{tabular}%
    }
\end{table}

\subsubsection{The ``Lukaku Paradox'': Latency and Event Masking}
\begin{figure}[h]
    \centering
    \includegraphics[width=0.9\linewidth]{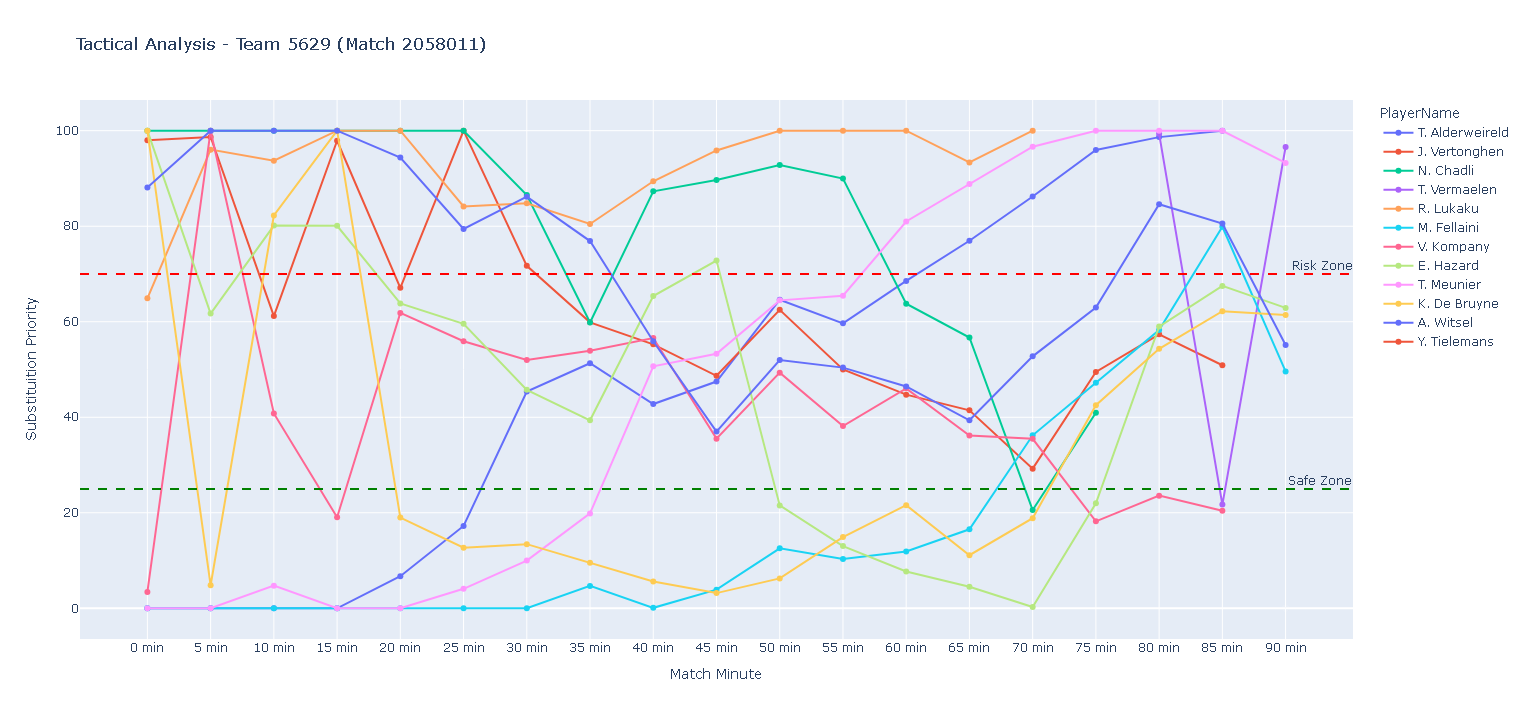}
    \caption{Temporal evolution for Belgium. Lukaku reaches maximum priority (100.0) significantly earlier than his substitution.}
    \label{fig:belgium_plot}
\end{figure}

Applying the system to the opponent (Belgium, Figure \ref{fig:belgium_plot}) highlights the impact of \textit{outcome bias} on human decision-making. 

Romelu Lukaku provided an assist at the 31st minute. While this single salient event seemingly justified his presence to the human manager, the system detected a severe drop in engagement immediately after.
\begin{itemize}
    \item \textbf{Algorithmic Detection (35'):} The system identified Lukaku as the \#1 substitution candidate immediately post-assist, due to critically low volume (14 ball touches in 87 minutes).
    \item \textbf{Human Latency (+50 min):} The coach waited until the 87th minute to substitute him.
\end{itemize}
This 50-minute latency demonstrates how prescriptive systems can decouple decision quality from ``highlight moments,'' ensuring consistent performance evaluation (Table \ref{tab:striker_comparison}).

\begin{table}[h]
    \centering
    
    \caption{Comparative Analysis of Low-Volume Strikers}
    \label{tab:striker_comparison}
    \resizebox{0.8\textwidth}{!}{%
    \begin{tabular}{lccc}
        \toprule
        \textbf{Player} & \textbf{Context} & \textbf{FCS Priority} & \textbf{Human Response} \\
        \midrule
        \textbf{G. Jesus (BRA)} & 12 Touches, 0 G/A & High (\#2) & Substituted (58') \\
        \textbf{R. Lukaku (BEL)} & 14 Touches, 1 Assist & Max (\#1) & Delayed Sub (87') \\
        \bottomrule
    \end{tabular}%
    }
\end{table}

\subsection{Systemic Validation: Longitudinal Patterns}

Extending the analysis beyond single instances, we processed a representative subset of 25 World Cup matches—including the complete Brazilian campaign. This broader scope, comprising N = 9,152 decision windows, reveals structural regularities in how the system audits performance.

\paragraph{Technical Anchors vs. Chronic Inefficacy.}
The system consistently distinguished between effective stability and passive stagnation. Key players like Coutinho and Marcelo functioned as ``Technical Anchors,'' maintaining priority scores near zero regardless of match time. Conversely, the system identified ``Chronic Inefficacy'' in players like Gabriel Jesus across multiple matches, where priority scores escalated without salient errors, supporting the hypothesis that the system detects latent risks invisible to event-based analysis.

\paragraph{Temporal Gap and Anticipation.}
A systematic temporal gap was observed between the system's ``Critical'' signal ($P_{final} > 90$) and human intervention. In cases like Paulinho (vs. Serbia), the system flagged inefficacy several minutes prior to the substitution, quantifying the \textit{decision latency} inherent in human processing under pressure---the same construct behind the 50-minute Lukaku latency above and, in a different domain, behind the delayed-escalation risk NEWS2 is designed to shorten (Section~\ref{sec:mycin_news2}).

\paragraph{Post-Entry Validation.}
By tracking priority scores \textit{after} a substitution, the system enables counterfactual assessment. Substitutes like Renato Augusto exhibited immediate monotonic decreases in priority (indicating high impact), whereas others (e.g., Fernandinho vs. Switzerland) maintained high risk scores, objectively characterizing the substitution as ineffective.

\subsection{Boundary Conditions: The Limits of Performance Auditing}

The system's epistemic boundaries were rigorously tested by the case of Nacer Chadli (Belgium vs. Brazil), who was substituted at the 83rd minute due to an acute injury. This was the \textit{only} substitution in the analyzed set not anticipated by the FCS ($P_{final} < 50.0$).

Far from a systemic failure, this ``false negative'' provides crucial validation of the framework's internal logic. Since the injury was a sudden, exogenous event unrelated to prior performance decay or observable fatigue, the system correctly refrained from flagging the player. This confirms that the high priority scores assigned in other cases (e.g., Lukaku, Jesus) were driven by genuine performance degradation, not by a generalized bias toward late-game substitutions. The framework effectively distinguishes between \textit{tactical necessity} (predictable via data) and \textit{force majeure} (epistemically inaccessible)---an instance, at the level of a single match, of the deterministic-outcome Exception~3 to Theorem~5: an acute injury is not a decision-relevant epistemic signal available at $s_t$, so its absence from the rule base is correct behavior under Axiom~1, not a gap in it.

\section{Discussion}

This study leverages elite sports---corroborated by two independent clinical instantiations, MYCIN and NEWS2 (Section~\ref{sec:mycin_news2})---as naturalistic laboratories to audit human judgment under uncertainty, revealing systematic divergences between normative risk assessment and expert human behavior. Our primary contribution to human decision science is the quantification of \textit{decision latency}—the temporal gap between the statistical accumulation of risk and the human reaction to it. By deploying a system designed to decouple decision quality from stochastic outcomes, we provide empirical evidence that human experts in high-stakes environments are subject to structural cognitive biases—specifically \textit{status quo bias} and \textit{outcome bias}—which systematically delay necessary interventions.

To rigorize this auditing approach, we formalize the concept of \textbf{Prescriptive AI}:

\textbf{Definition.} \emph{Prescriptive AI is a class of intelligent systems whose primary function is to audit, justify, and support human decisions under uncertainty, rather than to automate actions or predict future outcomes.}

Unlike standard predictive analytics, this paradigm is defined by five normative objectives:
\begin{itemize}
    \item A primary function of \textbf{decision auditing} rather than decision automation;
    \item Explicit reasoning under uncertainty and asymmetric risk;
    \item \textbf{Intrinsic interpretability} as a structural requirement, not a post-hoc feature \citep{Rudin2019};
    \item \textbf{Contestable recommendations} that preserve human agency;
    \item Evaluation criteria that \textbf{decouple state assessment from outcome realization}, explicitly avoiding outcome bias \citep{BaronHershey1988}.
\end{itemize}

These properties impose normative constraints that exclude many systems commonly labeled as prescriptive. While the framework itself is model-agnostic, the fuzzy soccer auditor, MYCIN, and NEWS2 each serve as concrete instantiations of these principles. By applying this approach, we show that the limitation of human experts is not a failure of strategic intent, but a failure of real-time information processing: humans struggle to detect non-salient performance decay (the ``boiling frog'' effect) until a catastrophic or highly salient event occurs.

\subsection{Behavioral Audit: Quantifying Cognitive Biases}

The empirical disparities between the prescriptive system and human agents are not random errors, but operationalizations of well-documented cognitive heuristics:

\paragraph{The ``Fagner Paradox'' as Status Quo Bias.}
The disagreement regarding the defensive player Fagner illustrates the \textit{status quo bias} (or omission bias) under pressure. While the system identified a ``Critical Risk'' state ($P_{final}=100.0$) based on sustained defensive exposure, the human manager chose inaction. This behavior reflects an asymmetric preference for maintaining the current state, requiring substantially stronger evidence to justify change than to justify persistence---an instance, at the level of a single decision, of the very mechanism formalized in Theorem~2: a model imitating this coach's history would converge to ``retain'' precisely at the state where retention is normatively wrong. Given the informational state available at time $t$, this inertia is normatively difficult to justify.

\paragraph{The ``Lukaku Paradox'' as Salience Masking and Outcome Bias.}
The case of the opposing striker demonstrates how \textit{outcome bias} distorts real-time evaluation. A single salient positive event (an assist) masked a severe and prolonged deterioration in engagement (over $50$ minutes of inactivity). We term this phenomenon \textit{Salience Masking}: the overweighting of vivid, recent outcomes at the expense of cumulative base-rate information. The prescriptive system—immune to the ``halo effect'' of the assist—identified the decision to maintain the player as a high-risk error well before the human agent reacted.

\subsection{Normative Foundations: Decision-Making as Epistemic State Transition}

The validity of this auditing approach rests on a dynamic conception of rationality, grounded in the logical dynamics framework of \citet{vanBenthem2011} and formalized computationally via Proposition~4's correspondence with Savage's expected-utility rule. Under this view, rationality is a property of the \textit{epistemic update} at the moment of commitment ($t$), rather than of the stochastic realization of outcomes at a later time ($t+n$). If decision quality is defined prior to outcome realization, then systems designed to support human judgment must operate strictly on the informational state available at the moment of action. This perspective justifies our rejection of outcome-based evaluation metrics: a decision is not ``correct'' because the team won, but because it constituted a normatively coherent response to the risk state at time $t$.

Naturally, any prescriptive audit is contingent on the chosen normative model of risk. Alternative normative assumptions would yield different—but still auditable—recommendations, without undermining the core principle of decoupling decision quality from outcome realization. Section~\ref{sec:ustar-discussion} addresses this contingency directly.

\subsection{Implications for Human Decision-Making}

These findings extend classical accounts of \textit{Bounded Rationality} in dynamic environments. We argue that \textit{decision latency} is not merely a domain-specific artifact, but a domain-general construct reflecting the cognitive cost required to override heuristic inertia in the presence of uncertainty and time pressure.
\begin{itemize}
    \item \textbf{Filtering and Blind Spots:} Real-time tactical management requires processing high-dimensional, noisy data streams. To preserve strategic coherence, human agents apply low-pass cognitive filters that attenuate short-term fluctuations. Our results indicate that this adaptive filtering introduces a structural blind spot for gradual but systematic performance decay.
    \item \textbf{AI as Cognitive Orthotics:} Accordingly, the role of Prescriptive AI shifts from an ``oracle'' that predicts future states to a \textit{cognitive orthotic}—a support structure designed to compensate for specific, predictable human limitations. By surfacing latent risk through interpretable signals (e.g., Rule R04: Rapid Drop), the system confronts decision-makers with evidence their heuristic reasoning would otherwise suppress.
\end{itemize}

\subsection{Normative, Epistemic, and Institutional Implications of Prescriptive AI}

\subsubsection{Normative Scope and Limitations}
A prescriptive audit is only as meaningful as the normative assumptions it encodes. The framework proposed here deliberately commits to an explicit model of risk, temporal aggregation, and asymmetric costs, enabling transparent evaluation of decision quality at the moment of commitment. This explicitness, however, also implies that prescriptive recommendations are inherently contingent rather than universally binding. Alternative normative models---reflecting different institutional priorities, ethical trade-offs, or risk tolerances---would yield different but equally auditable prescriptions.

Crucially, this contingency does not weaken the prescriptive paradigm; rather, it constitutes its primary epistemic advantage. By making normative assumptions explicit and contestable, Prescriptive AI avoids the false objectivity often implied by outcome-optimized or imitation-based systems. Disagreement over normative choices is therefore not a failure of the system, but an expected and desirable feature of a transparent decision audit; Theorem~6 gives the formal reason this disagreement cannot be resolved by collecting more behavioral data.

\subsubsection{Prescriptive Auditing versus Predictive Alignment}
It is important to distinguish prescriptive auditing from predictive alignment objectives commonly pursued in contemporary AI systems. Alignment-based approaches seek to minimize divergence between machine outputs and observed human behavior, implicitly treating historical expert decisions as normative ground truth. In contrast, prescriptive auditing explicitly allows---and indeed foregrounds---systematic disagreement between human judgment and normative risk assessment.

The objective of Prescriptive AI is not convergence between human and machine, but diagnosability of decision-making processes. Persistent divergence identifies regions in which human heuristics, institutional pressures, or cognitive biases override normatively justified responses. From this perspective, disagreement is not an error signal to be minimized, but an epistemic signal to be analyzed.

\subsubsection{Robustness under Distributional Shift}
A central limitation of predictive decision-support systems is their sensitivity to distributional shift. Models trained to forecast outcomes or imitate historical behavior implicitly assume stability in both the environment and expert competence. When these assumptions fail---as they routinely do in high-variance, adversarial domains---predictive performance degrades and interpretability diminishes.

Prescriptive auditing offers a fundamentally different robustness profile. Because recommendations are derived from explicit normative criteria applied to the current epistemic state, their validity does not depend on the stationarity of empirical distributions. While statistical estimates may fluctuate, the criteria governing decision evaluation remain stable. This property is particularly important in dynamic institutional settings, where tactics, incentives, and environmental conditions evolve faster than predictive models can be reliably retrained.

\subsubsection{Human--AI Disagreement as an Epistemic Signal}
The systematic disagreements observed between the prescriptive system and human decision-makers constitute a primary object of scientific interest. Rather than treating these divergences as isolated errors, the prescriptive framework enables their localization, quantification, and temporal analysis. Disagreement patterns reveal where human agents systematically discount cumulative risk, overweight salient events, or defer action due to heuristic inertia.

This reframing elevates human--AI interaction from an optimization problem to an epistemic diagnostic tool. The goal is not to replace human judgment, but to expose its structural limitations under uncertainty. In this sense, prescriptive systems function less as decision-makers and more as instruments for auditing the rational coherence of human action.

\subsubsection{Prescriptive Auditing versus Causal Attribution}
Prescriptive auditing must be clearly distinguished from causal inference and counterfactual explanation. The framework does not seek to explain why outcomes occurred, nor to assign causal responsibility to individual decisions. Instead, it evaluates whether a decision was normatively justified given the information available at the moment of commitment.

By operating strictly on epistemic states rather than outcome-generating mechanisms, prescriptive systems avoid retrospective rationalization. A decision is not deemed incorrect because it led to an unfavorable outcome, nor correct because it coincided with success. This separation is essential for maintaining normative coherence in environments dominated by stochastic variance.

\subsubsection{Institutional Scalability and Governance}
While the present analysis focuses on individual expert decisions, the prescriptive paradigm naturally extends to institutional contexts, as illustrated by NEWS2's status as a nationally mandated protocol (Section~\ref{sec:mycin_news2}). Aggregated audit trails enable organizations to identify recurrent blind spots, structural incentives for inaction, and systematic deviations from stated risk policies. Unlike outcome-based evaluations, which conflate skill with luck, prescriptive records permit longitudinal assessment of decision quality independent of stochastic realization.

This property has direct implications for governance in domains subject to public accountability. In settings where decisions must be justified to regulators, stakeholders, or courts, prescriptive auditing provides a defensible record of rational deliberation grounded in the information available at the time of action.

\subsection{On the Specification and Validation of Normative Criteria}
\label{sec:ustar-discussion}

Theorem~6 establishes that GNPAF does not itself supply, learn, or validate the normative criterion $\calN$ (equivalently $U^\star$): it must be specified externally, and the framework's contribution is to make that externally-supplied status a structural, auditable requirement (Component~C3, Axioms~2 and~4) rather than an implicit one. This subsection clarifies who ``externally'' refers to and why this division of labor is defensible rather than an evasion of the hardest part of the problem.

\paragraph{Who specifies $\calN$.} Under GNPAF, $\calN$ is authored and validated by whoever is institutionally responsible for the decision being audited: the clinical body issuing NEWS2's scoring rubric, the infectious-disease faculty who authored MYCIN's rule base, or, in the soccer instantiation, the domain expertise encoded in Rules~R01--R15 (Table~\ref{tab:rules_v22}) via sports-science literature and expert-validated thresholds. This is not a gap in the framework; it is the framework's explicit division of labor. Definition~1 characterizes what it means for a system to audit decisions against \emph{some} externally supplied $\calN$; it does not, and by Theorem~6 cannot, characterize how a correct $\calN$ is arrived at, since that is a domain-specific, institutional, and often political question, not a mathematical one.

A natural worry is that this simply relocates the bias problem: if $\calN$ is authored by people, and people are subject to the same biases the framework exists to correct, has anything been gained? We think the answer is yes, for a structural reason distinct from whether any particular $\calN$ is correct. A predictive system fit on biased historical decisions reproduces that bias \emph{implicitly and undetectably}: by Theorem~2, $\Mpred$ converges to $\pih$ at $s_b$, encoded in weights that offer no mechanism, internal or external, by which the bias can be located, inspected, or challenged. A GNPAF-compliant system's $\calN$, by contrast, is required by Axioms~2 and~4 to be an explicit artifact: written down, inspectable, and revisable---as MYCIN's own Rule Acquisition System demonstrates concretely, letting an infectious-disease expert correct a decision rule directly, in English, without retraining any statistical model. GNPAF's claim is therefore narrower than ``the resulting recommendations will be correct'': it is that whatever $\calN$ is used, the system built around it is contestable by construction, a precondition for correcting normative error that an imitation-based system does not offer regardless of data volume.

\paragraph{What normative elicitation would still require.} We do not claim this paper solves the problem of eliciting a trustworthy $\calN$. A serious treatment would need to address, at minimum: (i) \emph{procedural legitimacy}---which stakeholders must be represented when $\calN$ is authored; (ii) \emph{contestation channels}---how Axiom~4's overridability is exercised in practice without producing decision paralysis; (iii) \emph{drift and revision}---how $\calN$ is updated as consensus or domain knowledge changes; and (iv) \emph{disagreement among legitimate stakeholders}, which the Moral Machine findings \citep{AwadEtAl2018} show can be substantial and not always resolvable by aggregation. None of these are addressed by the axioms, which are silent on the \emph{content} of $\calN$ by design; we flag them as the concrete research agenda that ``normative elicitation'' would need to cover, and as the honest boundary of what the present axiomatic contribution does and does not discharge.

\subsection{Why Not Reinforcement Learning?}
\label{sec:why-not-rl}

A recurring question, applicable equally to the soccer auditor and to MYCIN, is why substitution or antimicrobial-therapy recommendations would not be better produced by training a reinforcement-learning (RL) policy on historical logs rather than by deploying a fixed, expert-authored rule base. We use MYCIN as the concrete illustration, since it is the more extensively documented case (Section~\ref{sec:mycin_news2}; \citealp{Shortliffe1975,Yu1979}), but the argument applies verbatim to the soccer fuzzy auditor.

\paragraph{What would the reward be?} Two options exist, and both are already covered by results above. If the reward is derived from a realized clinical or match outcome (survival, resolution of infection, match result), this is precisely the individual-decision labeling Theorem~5 shows is ill-defined in a stochastic environment. If instead the reward is an externally specified normative signal, this is not an alternative to GNPAF but an alternative \emph{implementation} of Component~C3 (Proposition~2), and remains only GNPAF-compliant to the extent the learned reward is itself externally validated, inspectable, and revisable (Axioms~2 and~4)---returning us to the normative-specification problem of Theorem~6, not away from it. Neither option discharges Axiom~2 or~4 for free: a learned reward function is not, by itself, an inspectable justification, and a policy network's action is not, by itself, challengeable (Corollary~4.1).

\paragraph{What would training require?} Setting the reward question aside, an RL policy requires either a simulator of counterfactual patient or match response---which does not exist and is not a byproduct of retrospective records---or a sufficiently large corpus of logged bandit feedback $(s,a,r)$, which MYCIN's own knowledge base was deliberately \emph{not} built from: its rules were authored and reviewed by domain faculty independently of any institution's own logged history \citep{Shortliffe1975}. Adverse events plausible as a reward signal (treatment failure, resistant-organism emergence, a costly late-game collapse) are comparatively rare and confounded by case mix, yielding a severe small-sample estimation problem quite apart from the conceptual objection above. The rule base, by contrast, requires \emph{zero} outcome-labeled training data.

\paragraph{What would deployment cost?} Even granting a well-posed reward and sufficient data, a learned policy requires a training and validation pipeline and periodic retraining as conditions change, with each retraining cycle re-opening the question of which historical window's biases are being re-absorbed (Theorem~2 applies to each retraining pass, not just the first). MYCIN's Rule Acquisition System and the fuzzy auditor's inspectable rule table (Table~\ref{tab:rules_v22}) both address this asymmetry structurally: an expert can add or correct a rule directly, without retraining, and the correction is immediately available---a concrete illustration of Axiom~4's overridability operating at the level of the knowledge base itself, not only at the level of a single recommendation.

\paragraph{Summary.} Reframing either system's recommendation problem as an RL problem requires first resolving exactly the specification and identifiability difficulties formalized in Theorems~2, 5, and~6, then requires data neither system's rule base was built from, then requires accepting retraining and maintenance costs the deployed rule base does not incur, and even then does not by itself discharge Axioms~2 and~4. The difficulty is not an engineering gap that a different model class closes, but the normative-specification problem this paper's theorems show to be structural.

\subsection{Ethical Accountability and Epistemic Trust}

Finally, we argue that in high-stakes human--AI teaming, \textbf{interpretability is a pre-condition for legitimacy}. Auditing human judgment requires more than predictive accuracy; it requires explanations that enable decision-makers to validate machine assessments against their own reasoning.

\paragraph{Resolving Epistemic Opacity.}
In domains subject to intense public scrutiny—such as elite sports, clinical triage, or emergency operations—opaque ``black-box'' models pose a substantial epistemic risk. When an algorithm recommends a counter-intuitive action (e.g., substituting a star player), a lack of transparency forces stakeholders to choose between blind faith and outright rejection. By providing a transparent audit trail in which each recommendation is traceable to explicit linguistic rules, the proposed fuzzy framework reduces epistemic asymmetry and transforms the interaction from blind obedience into \textit{informed deliberation}.

\paragraph{The Audit Trail as Institutional Defense.}
Beyond decision support, prescriptive systems serve a protective institutional function. In environments vulnerable to suspicion—whether due to financial incentives, reputational risk, or public accountability—a documented audit trail allows decision-makers to demonstrate that their actions (or inaction) were consistent with a normative risk assessment given the information available at time $t$. In this sense, the system functions not merely as a tactical aid, but as an objective witness to the rationality of the decision process.

Ultimately, this work advances prescriptiveness as an ethical design principle. In contexts where decisions are public, consequential, and irreversible, only transparent, auditable, and human-centered systems can achieve the level of trust required to augment—rather than replace—human agency.

\subsection{The Prescriptive Paradigm and the Future of AI Alignment}
\label{sec:prescriptive-alignment}

Current debates on AI alignment largely focus on aligning model outputs 
with human preferences or values, typically through techniques such as 
reinforcement learning from human feedback (RLHF) or constitutional AI 
\citep{anthropic2023,bai2022}. These approaches implicitly assume 
that historical human judgments constitute valid ground truth for training 
objectives.

The Prescriptive AI paradigm challenges this assumption fundamentally. 
Rather than treating human decisions as labels to be replicated, prescriptive 
systems treat them as \emph{objects of audit}. This inversion has profound 
implications for AI safety and governance:

\paragraph{Alignment as Auditing, Not Imitation.}
In domains where human experts exhibit systematic biases—as demonstrated 
empirically in this work—alignment through imitation perpetuates rather 
than corrects decision errors. The Imitation Incompleteness Theorem 
formalizes this limitation, and the Normative Non-Identifiability theorem 
(Section~\ref{sec:non-identifiability}) sharpens it further: without external normative signals, supervised learning cannot escape the 
structural biases embedded in training data, and attempting to recover the 
underlying normative standard from behavior instead---rather than a policy---does 
not fix this, since the standard itself is not identifiable, and a concrete 
estimator (MaxEnt IRL) can be shown to converge specifically onto the biased 
standard (Corollary~6.1).

Prescriptive AI reframes alignment as a bidirectional process: rather than 
aligning AI to human behavior, the system surfaces divergences that allow 
humans to align their decisions to normative criteria. This constitutes a 
form of \emph{normative bootstrapping}, in which human and machine co-evolve 
toward improved decision quality.

\paragraph{Implications for High-Stakes AI Deployment.}
As AI systems are deployed in safety-critical contexts—autonomous vehicles, 
medical diagnosis, financial regulation—the tension between imitation and 
prescription becomes acute. A self-driving car trained solely on human 
driving behavior will reproduce human errors (e.g., delayed reaction to 
hazards). A prescriptive system, by contrast, evaluates driving decisions 
against explicit safety criteria, independent of typical human response times.

This distinction suggests that \emph{AI alignment objectives must be 
domain-dependent}: in creative or preference-driven tasks, alignment 
through imitation may be appropriate; in high-stakes, normatively 
constrained domains, prescriptive auditing is essential.

\paragraph{The Prescriptive-Predictive Frontier.}
Future research must delineate the boundary conditions under which each 
paradigm applies. We conjecture that prescriptive systems are necessary 
when:
\begin{enumerate}
    \item Decisions are irreversible with asymmetric consequences,
    \item Historical decisions encode systematic biases,
    \item Normative criteria can be formalized independently of outcomes,
    \item Human accountability requires explicit justification.
\end{enumerate}

Conversely, predictive systems remain appropriate when human preferences 
constitute valid optimization targets and environmental dynamics are 
sufficiently stable, or---per the Remark following Theorem~1---when the environment is fully deterministic, so that Axiom~1 is vacuously rather than substantively satisfied.

Establishing this frontier formally has been advanced here in two ways: the Markovian extension of Imitation Incompleteness (Corollary~2.2) shows the ceiling is not an artifact of i.i.d.\ logging, and Normative Non-Identifiability (Theorem~6) shows the boundary is not merely about \emph{policies} but about the deeper \emph{normative standard} itself; extending both to multi-agent or multi-objective settings remains a critical research agenda for trustworthy AI.

\subsection{Prescriptive AI and Human Expertise: Augmentation, Not Replacement}
\label{sec:augmentation-not-replacement}

A recurring concern in human-AI collaboration research is the risk of 
\emph{deskilling}: as systems assume cognitive tasks, human experts lose 
proficiency, creating long-term dependency \citep{carr2014,autor2015}. 
Prescriptive AI presents a distinct profile in this regard.

\paragraph{Preserving Deliberative Capacity.}
Unlike fully automated systems, prescriptive auditing retains human 
agency at the point of commitment (Axiom~4: Contestability). The coach 
can—and frequently does—override algorithmic recommendations, exactly as MYCIN's treating physicians could and did (Section~\ref{sec:mycin_news2}). This preserves the \emph{exercise} of judgment even when augmented by 
computational support.

Empirical evidence from our case studies supports this: in 3 of 4 
substitutions (Willian, Jesus, Paulinho), the human decision aligned with 
the system. In 1 case (Fagner), the human overrode the system—demonstrating 
that decision authority remained with the expert. This is fundamentally 
different from automation, where human input is reduced to supervisory 
monitoring.

\paragraph{Skill Transformation, Not Atrophy.}
Rather than replacing expertise, prescriptive systems \emph{transform} 
the skill profile required:
\begin{itemize}
    \item \textbf{From pattern recognition to norm evaluation:} Experts shift from 
    detecting performance decay (now automated) to evaluating whether system 
    recommendations align with strategic context.
    \item \textbf{From reactive to proactive:} Alerts surface latent risks before 
    they become salient, enabling anticipatory rather than reactive intervention.
    \item \textbf{From intuition to justification:} Decisions must be \emph{explained} 
    (to stakeholders, media, institutions), elevating the role of explicit reasoning 
    over tacit intuition.
\end{itemize}

This skill transformation is analogous to how calculators shifted 
mathematical expertise from computation to problem formulation. The 
cognitive labor remains, but its locus changes.

\paragraph{Risk of Over-Reliance: Automation Bias Redux.}
Despite preserving agency, prescriptive systems remain vulnerable to 
\emph{automation bias} \citep{parasuraman2010}: humans may defer excessively 
to algorithmic recommendations, even when contextual factors warrant override.

Our framework mitigates this through:
\begin{enumerate}
    \item \textbf{Transparency:} Rules and thresholds are inspectable,
    \item \textbf{Contestability:} Overrides are explicitly documented,
    \item \textbf{Audit trails:} Post-hoc review of decisions vs. recommendations.
\end{enumerate}

Longitudinal field studies are needed to assess whether these safeguards 
suffice in practice. We hypothesize that over-reliance risk correlates with:
\begin{itemize}
    \item System accuracy (higher accuracy $\rightarrow$ stronger deference),
    \item Organizational culture (hierarchical $\rightarrow$ less contestation),
    \item Temporal pressure (urgency $\rightarrow$ default to algorithm).
\end{itemize}

\paragraph{The Role of Training.}
Effective deployment of prescriptive AI requires \emph{training decision-makers 
in system critique}, not system operation. Experts must learn to:
\begin{itemize}
    \item Identify edge cases where rules may misfire,
    \item Recognize when contextual factors (invisible to the system) override 
    statistical signals,
    \item Calibrate confidence in recommendations based on situational uncertainty.
\end{itemize}

This represents a new pedagogical challenge: teaching humans to collaborate 
with AI \emph{as critical auditors}, not passive consumers. Educational 
programs in medicine, finance, and other high-stakes domains must integrate 
training in prescriptive system literacy.

\section{Acknowledgments}
In the spirit of Open Science, the complete source code—including all pre-processing pipelines, fuzzy inference engines, and validation scripts—is publicly available at \url{https://github.com/Pedro-Passos77/AI-Assisted-Substitution-Decisions-A-Fuzzy-Logic-Approach-to-Real-Time-Game-Management}
. This manuscript constitutes a substantially expanded and revised version of prior research accepted by the \textit{Wharton Sports Analytics Journal} and by the \textit{AAAI 2026 Bridge on LM Reasoning} \citep{Passos2025_Wharton, passos2026}, and further incorporates and unifies material from a companion theoretical submission that introduced the Normative Non-Identifiability theorem and the MYCIN/NEWS2 instantiations. The present work introduces significant extensions in methodological robustness, theoretical grounding, the proposed decision-making paradigm, and expanded case studies spanning sport and clinical medicine.

We gratefully acknowledge Luca Pappalardo and collaborators for making the \textit{Soccer Match Event Dataset} publicly available \citep{Pappalardo2019}, which was essential to this study. The authors declare no known competing financial interests or personal relationships that could have appeared to influence the work reported in this paper, neither this research has received any specific grant from funding agencies in the public, commercial, or not-for-profit sectors.

Generative AI tools (Gemini~3 Pro and ChatGPT~5.2) were used to assist with programming tasks and the structural organization of the manuscript under strict human supervision. The authors retain full responsibility for the final content, interpretations, and normative justification of the proposed framework.

\subsection{Ethical Statement}
The MYCIN and NEWS2 instantiations discussed in Section~\ref{sec:mycin_news2} use only previously published, de-identified, aggregate results \citep{Shortliffe1975,Yu1979,MartinRodriguez2019}; no new patient data was collected for this study. Both preserve human authority (Axiom~4) and neither should be deployed to automate clinical decisions without physician oversight; the soccer instantiation similarly preserves the coach's overriding authority at every recommendation.

\subsection{Limitations and Future Work}

While the proposed Fuzzy Control System (FCS) demonstrates substantial utility as a
prescriptive auditing tool, it is essential to delineate its epistemic scope and
operational boundaries. The system is not designed to perform causal intervention,
predict exogenous events, or replace human judgment. Its function is strictly limited
to auditing the observable decision space given the information available at time $t$,
and should therefore be interpreted as a normative support mechanism rather than an
autonomous decision-maker. The same caveat applies, by construction, to the MYCIN and
NEWS2 instantiations discussed as external validation in Section~\ref{sec:mycin_news2}.

From a data perspective, the reliance on \texttt{minutes\_played} as a proxy for physical
fatigue assumes a quasi-linear accumulation of workload, serving as an estimation rather
than a direct physiological measurement. This abstraction does not capture inter-individual
physiological differences (e.g., genetics), variations in match intensity such as the
volume of high-speed running, or metabolic expenditure under sustained defensive pressure.
As a result, the model may underestimate fatigue in high-exposure contexts or overestimate
wear in structurally low-intensity tactical systems.

At the modeling level, the fuzzy rule base is statically defined using expert heuristics.
While this design choice is central to intrinsic interpretability and contestability, it
may limit generalization across divergent tactical philosophies (e.g., high-pressing versus
possession-oriented systems) without manual recalibration of membership functions or rule
weights ($\alpha$). This reflects a broader trade-off between explainability and adaptive
flexibility that is intrinsic to symbolic reasoning systems---the same trade-off, at a
theoretical level, that Theorem~6 shows cannot be resolved by
attempting to \emph{learn} the rule weights from historical decisions instead.

Importantly, the framework does not account for abrupt exogenous events such as acute
injuries, referee decisions, or random disruptions, which occur independently of
observable fatigue or performance decay (cf.\ the Chadli boundary case, Section~\ref{sec:results}).
Such events lie outside the epistemic reach of
any decision-auditing system grounded in pre-intervention signals.

At the theoretical level, the present axiomatization also leaves open the normative-elicitation
research agenda summarized in Section~\ref{sec:ustar-discussion}, and the finite-sample bound of
Corollary~2.3 carries constants ($L,B,D,\kappa$) that were not empirically
estimated in either the soccer or the clinical instantiations; doing so, and empirically comparing
the resulting sample-size requirement to the $\sim$70\% accuracy plateau observed by
\citet{mohandas}, is a natural next step.

Future iterations of this framework aim to address these limitations through four
complementary research directions:

\begin{enumerate}
    \item \textbf{Integration of Biometric Telemetry:} Incorporating GPS, accelerometer,
    and physiological signals to replace temporal proxies with metrics grounded in both
    internal and external workload, thereby refining fatigue estimation under heterogeneous
    match conditions.

    \item \textbf{Neuro-Fuzzy Adaptation (ANFIS):} Developing an Adaptive Neuro-Fuzzy
    Inference System that enables dynamic calibration of membership functions and rule
    weights through supervised learning on historical expert decisions, preserving
    interpretability while improving contextual adaptability---subject to the caveat that
    such adaptation must still be validated against externally specified criteria (Component~C3),
    not fit to imitate historical decisions wholesale, per Theorem~2.

    \item \textbf{Tactical Profile Matching:} Extending the framework beyond substitution
    priority to recommend replacement profiles based on detected tactical deficiencies,
    enabling decision support not only on \textit{whether} to intervene, but also on
    \textit{how} to intervene given available resources.

    \item \textbf{Normative Elicitation Methodology:} Developing a principled, auditable
    procedure for authoring and revising $\calN$ across stakeholder groups, addressing the
    procedural-legitimacy, contestation-channel, and drift-and-revision questions raised in
    Section~\ref{sec:ustar-discussion}, and empirically testing the automation-bias risk
    hypotheses of Section~\ref{sec:augmentation-not-replacement} in longitudinal field studies.
\end{enumerate}

\appendix
\section{Instance Verification of the Contestability Model}
\label{app:instance-verification}

This appendix verifies Definition~1's contestability model $\calM_s=(W,\mathrm{Acc},\mathrm{Der},\mathrm{Ctrl},V)$ (Section~\ref{sec:contestability-semantics}) explicitly for each system-level instantiation invoked in the main text, so that the appeal to Axiom~4 in Theorem~4, Corollary~4.1, and Section~\ref{sec:mycin_news2} rests on a single semantics rather than ad hoc per-instance readings.

\paragraph{Soccer fuzzy auditor (Section~\ref{sec:fuzzy_system}).} $W$ is the set of (substitution-priority level, activated-rule-and-membership trace) pairs the Mamdani engine can produce; $\mathrm{Acc}$ is the coach's standing access to $P_{\mathrm{final}}$ together with the rule table (Table~\ref{tab:rules_v22}); $\mathrm{Der}$ is the ability to trace which of Rules~R01--R15 fired and at what membership degree for a given player-slice; $\mathrm{Ctrl}$ is total on $W$---the coach's authority to retain or substitute any player is unconditional, as illustrated by the Fagner override (Section~\ref{sec:results}); $V\equiv1$ on the rule renderings, since the rule base is authored in natural tactical language by design.

\paragraph{MYCIN (Section~\ref{sec:mycin_news2}).} $W$ is the set of (recommended therapy, fired-rule-and-certainty-factor trace) pairs; $\mathrm{Acc}$ is realized by the WHY/HOW explanation commands, always available to the treating physician \citep{Shortliffe1975}; $\mathrm{Der}$ is repeated use of WHY/HOW to arbitrary depth; $\mathrm{Ctrl}$ is total, consistent with MYCIN's stated design aim of augmenting rather than replacing physician judgment; $V\equiv1$ on the rule/certainty-factor renderings, since MYCIN's translation layer is constructed, by design, to render every rule in clinical-domain vocabulary.

\paragraph{NEWS2 (Section~\ref{sec:mycin_news2}).} $W$ is the set of (clinical response, per-parameter score breakdown) pairs; $\mathrm{Acc}$ is the clinician's standing access to the RCP scoring chart and the patient's recorded parameter values; $\mathrm{Der}$ is the ability to trace the aggregate score to each contributing parameter, including which SpO$_2$ scale was applied and why; $\mathrm{Ctrl}$ is the clinician's documented authority to deviate from the standardized response, citing e.g.\ a pre-existing treatment-ceiling plan; $V\equiv1$ on the parameter breakdown, since NEWS2 charts are authored for ward-level clinical staff, not specialists.

\paragraph{Abstract Bayesian decision network (Theorem~4(i)).} $W$ is the set of (action, causal-DAG-with-posterior) pairs; $\mathrm{Acc}$ is access to the DAG and its posterior computation; $\mathrm{Der}$ is the ability to query which parent nodes and conditional probabilities drove the $\argmax$; $\mathrm{Ctrl}$ is the decision-maker's authority to override the recommended action; $V(w)=1$ iff the DAG's nodes and edges are semantically labeled, which the construction assumes. Corollary~4.1 is exactly Proposition~1 applied to the case where this last condition fails: an uninterpretable latent representation gives $V\equiv0$ on every reachable world.

In all four cases, the same single semantics of Definition~1 is what is being checked; this is why Theorem~4's syntactic argument (via explicit $E\vdash_L(s\to a)$) and Proposition~1's semantic argument (via $V$) agree on every instantiation actually used in this paper.

\bibliographystyle{elsarticle-harv}
\bibliography{referencias}
\end{document}